\documentclass{article}

\usepackage[preprint]{neurips_2026}

\usepackage[utf8]{inputenc} 
\usepackage[T1]{fontenc}    
\usepackage{hyperref}      
\usepackage{url}           
\usepackage{booktabs}      
\usepackage{amsfonts}      
\usepackage{nicefrac}       
\usepackage{microtype}      
\usepackage{xcolor}         
\usepackage{graphicx}
\usepackage{subcaption}
\usepackage{amsmath}
\usepackage{makecell}
\usepackage{colortbl}
\usepackage{float}
\usepackage{calc}
\usepackage{multirow}
\usepackage{adjustbox}
\usepackage{color,xcolor,colortbl}

\definecolor{lightgray}{gray}{0.95}
\definecolor{color3}{gray}{0.95}
\definecolor{rouse}{rgb}{0.981,0.961,0.941}
\definecolor{light-yellow}{rgb}{1,1,0.93}
\definecolor{light-green}{rgb}{0.95,1,0.95}

\title{VDFP: Video Deflickering with Flicker-banding Priors}

\author{
	Zhiyi Zhou$^{1}$\thanks{Equal contribution.},\enspace
    Libo Zhu$^{1}$\footnotemark[1],\enspace
    Zihan Zhou$^{1}$,\enspace
    Yulun Zhang$^{1}$\thanks{Corresponding author: Yulun Zhang, yulun100@gmail.com},\enspace
    Xiaokang Yang$^{1}$  \\
	$^{1}$Shanghai Jiao Tong University
}

\begin{document}

\maketitle
\vspace{-35pt}
\begin{figure}[htbp]      
\centering
\includegraphics[width=\textwidth]{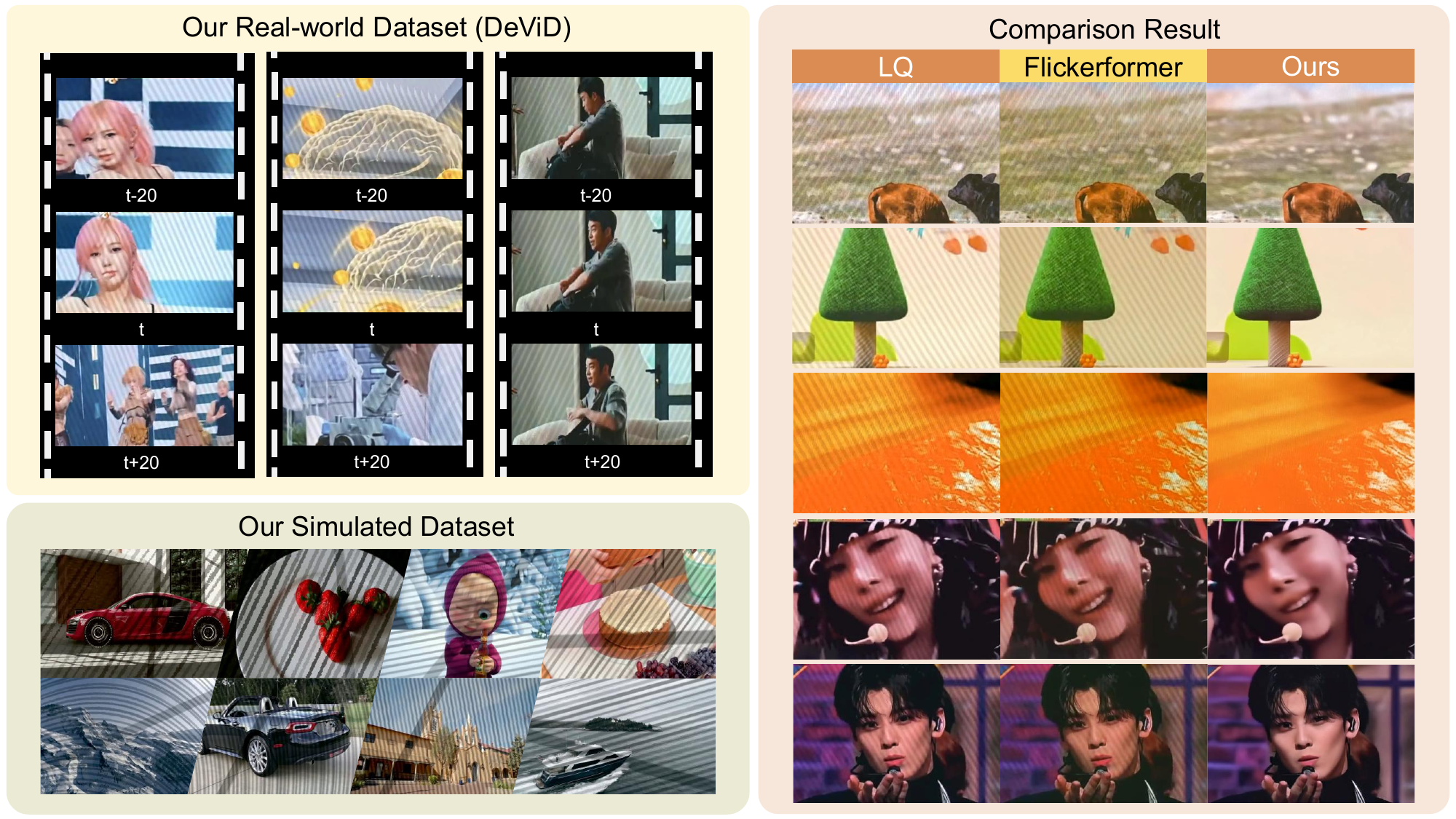}
\vspace{-12.5pt}
\caption{Overview of our datasets and comparisons. The upper-left part shows our real-world dataset (DeViD) and the lower-left part shows our simulated dataset. The right part presents low-quality (LQ) video frames and deflickering results of Flickerformer~\cite{flickerformer} and our method VDFP.}
\label{fig:main_picture}
\end{figure}

\begin{abstract}
Capturing digital screens with smartphones frequently induces severe banding due to hardware synchronization mismatches. Existing video restoration methods struggle with these structured, periodic luminance fluctuations, often resulting in residual artifacts or over-smoothed textures. We firstly construct DeViD, a real-world dataset in various scenes to deal with the lack of available datasets.Then we propose VDFP (\textbf{V}ideo \textbf{D}eflickering with \textbf{F}licker-banding \textbf{P}riors), a novel perception-guided generation framework. First, we introduce a Degradation Field Modeling Based on Rolling Shutter Mechanism (DFM) capable of synthesizing complex multi-banding scenarios. Second, we present a spatial-temporal continuous prior perception (CPP). Unlike traditional binary segmentation, this module is optimized via a Flicker-Aware Mean Squared Error (FA-MSE) to capture the luminance transitions. By zero-initializing an augmented input layer, our model preserves pre-trained generative priors as well as spatial-temporal prior perception. Extensive experiments demonstrate that VDFP significantly outperforms other methods, eliminating complex banding with high-fidelity spatial details and temporal consistency. Our dataset and code will be released at~\url{https://github.com/ZhiyiZZhou/VDFP}.
\end{abstract}
\begin{figure*}[t!]
    \centering
    \captionsetup[subfigure]{skip=-0.1pt}
    \begin{subfigure}[b]{0.19\linewidth}
        \includegraphics[width=\linewidth]{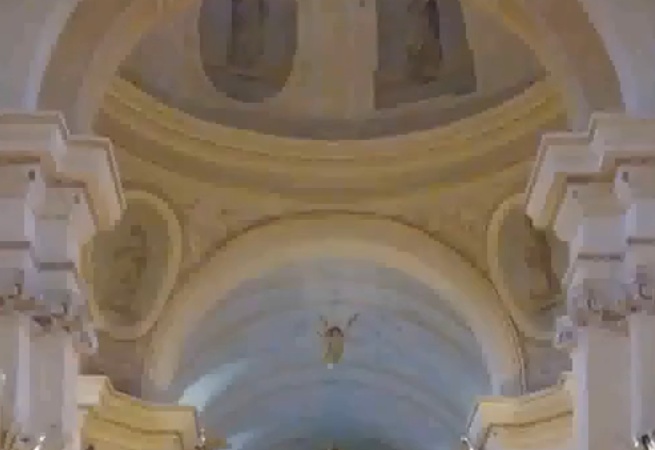}
        \subcaption*{GT}
    \end{subfigure}\hfill
    \captionsetup[subfigure]{skip=-0.1pt}
    \begin{subfigure}[b]{0.19\linewidth}
        \includegraphics[width=\linewidth]{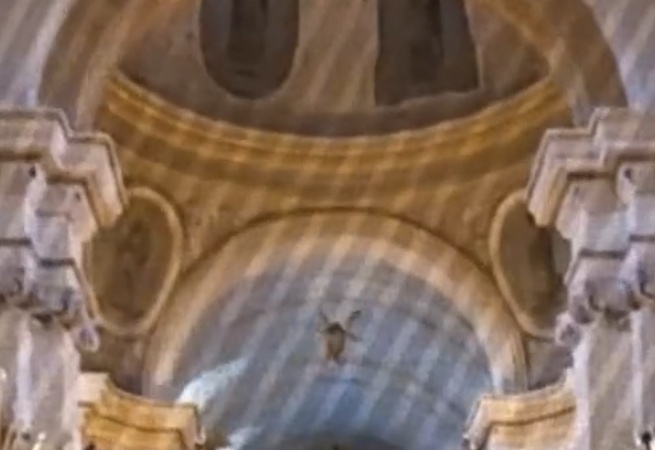}
        \subcaption*{LQ}
    \end{subfigure}\hfill
    \captionsetup[subfigure]{skip=-0.1pt}
    \begin{subfigure}[b]{0.19\linewidth}
        \includegraphics[width=\linewidth]{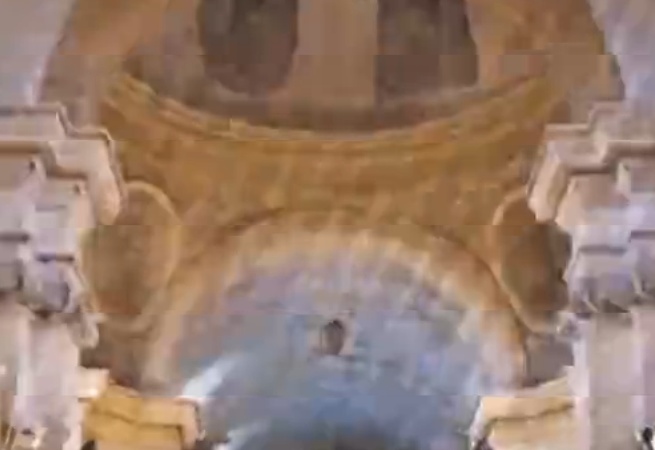}
        \subcaption*{STBN~\cite{STBN}}
    \end{subfigure}\hfill
    \captionsetup[subfigure]{skip=-0.1pt}
    \begin{subfigure}[b]{0.19\linewidth}
        \includegraphics[width=\linewidth]{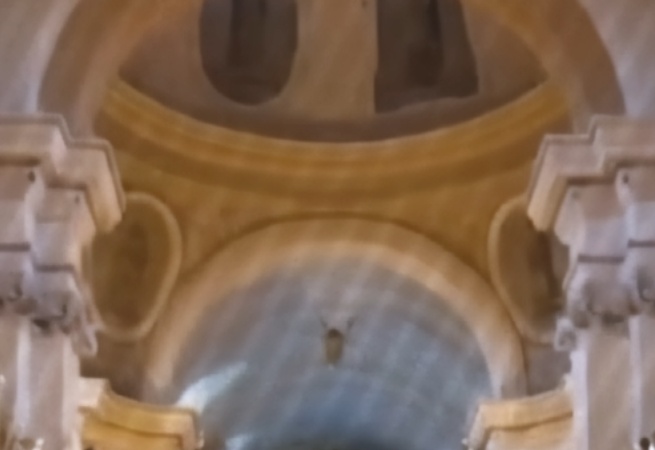} 
        \subcaption*{STAR~\cite{star}}
    \end{subfigure}\hfill
    \captionsetup[subfigure]{skip=-0.1pt}
    \begin{subfigure}[b]{0.19\linewidth}
        \includegraphics[width=\linewidth]{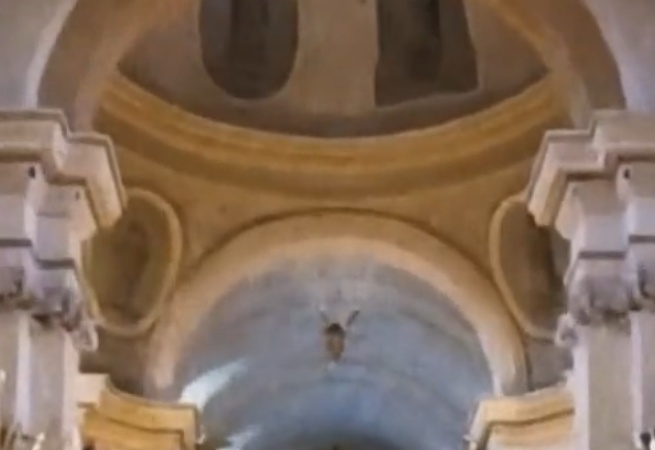}
        \subcaption*{DLoRAL~\cite{DLoRAL}}
    \end{subfigure}
    
    \vspace{-1pt}

    \captionsetup[subfigure]{skip=-0.1pt}
    \begin{subfigure}[b]{0.19\linewidth}
        \includegraphics[width=\linewidth]{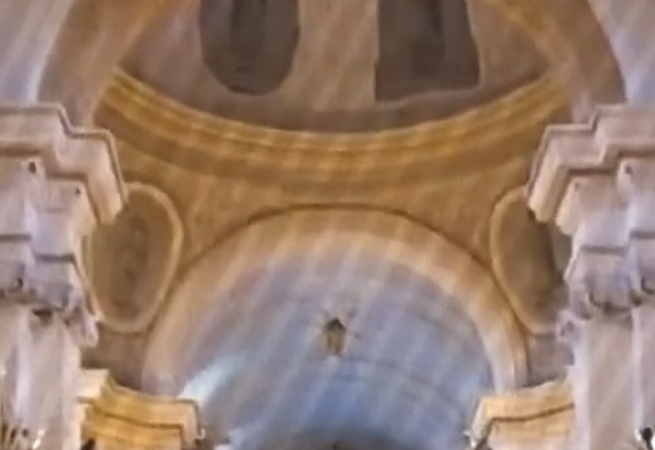}
        \subcaption*{FPANet~\cite{oh2025fpanet}}
    \end{subfigure}\hfill
    \captionsetup[subfigure]{skip=-0.1pt}
    \begin{subfigure}[b]{0.19\linewidth}
        \includegraphics[width=\linewidth]{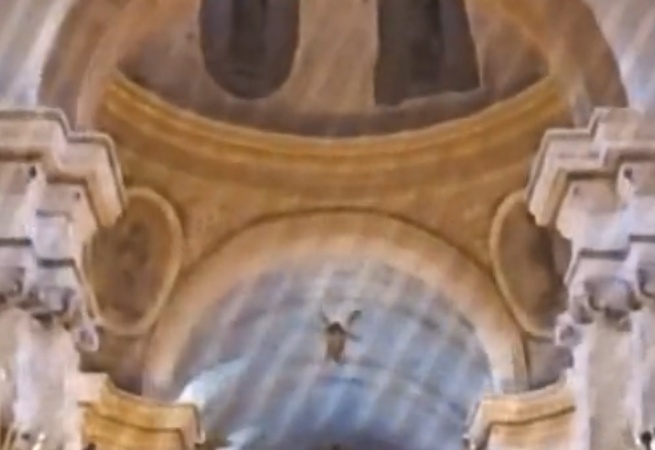}
        \subcaption*{CompEvent~\cite{zhong2025compevent}}
    \end{subfigure}\hfill
    \captionsetup[subfigure]{skip=-0.1pt}
    \begin{subfigure}[b]{0.19\linewidth}
        \includegraphics[width=\linewidth]{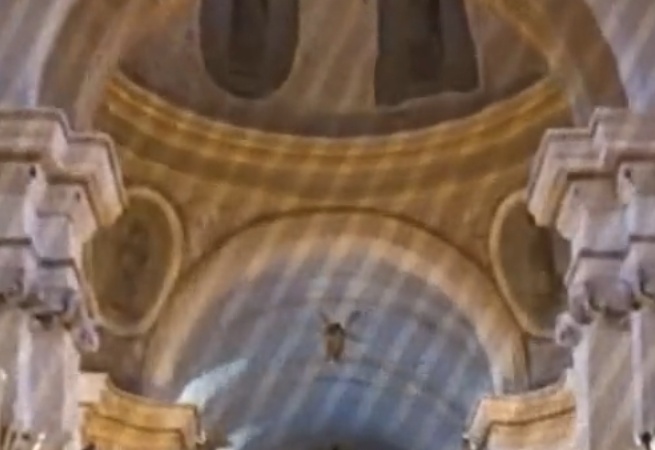}
        \subcaption*{Flickerformer*~\cite{flickerformer}}
    \end{subfigure}\hfill
    \captionsetup[subfigure]{skip=-0.1pt}
    \begin{subfigure}[b]{0.19\linewidth}
        \includegraphics[width=\linewidth]{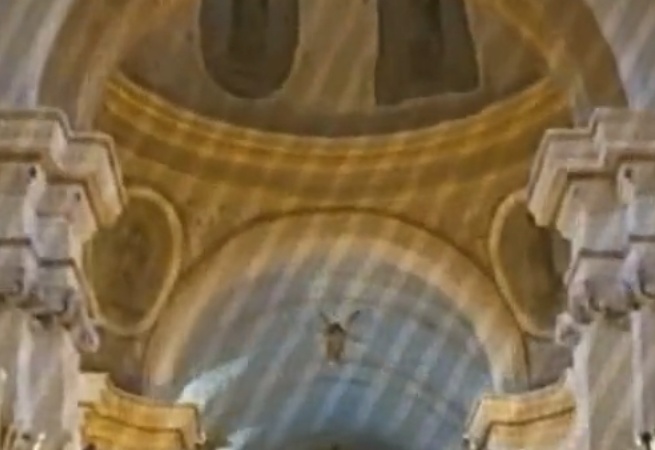}
        \subcaption*{Flickerformer~\cite{flickerformer}}
    \end{subfigure}\hfill
    \captionsetup[subfigure]{skip=-0.1pt}
    \begin{subfigure}[b]{0.19\linewidth}
        \includegraphics[width=\linewidth]{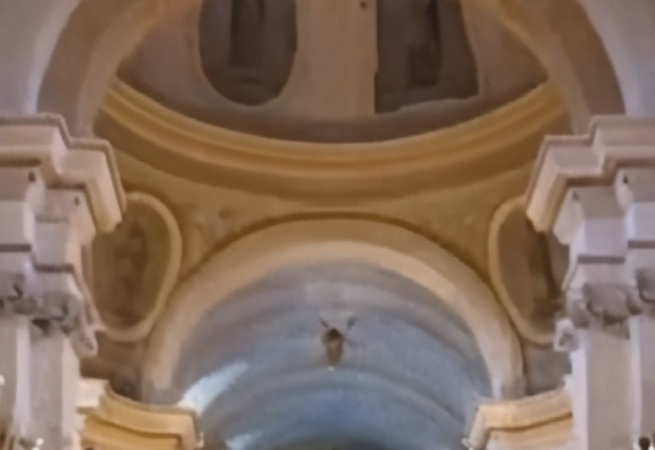}
        \subcaption*{VDFP (ours)}
    \end{subfigure}
    
    \vspace{-3mm}
    \caption{Visual comparison between VDFP and other methods trained on our real-world dataset DeViD. Specifically, Flickerformer* represents using the official checkpoint to test.}
    \label{fig:visual_comparison_one}
    \vspace{-6mm}
\end{figure*}

\setlength{\abovedisplayskip}{2pt}   
\setlength{\belowdisplayskip}{2pt}

\section{Introduction}
\vspace{-4mm}
It's commonly observed in daily life that when people use smartphones to capture display matrices, periodic luminance stripes often occur in videos, reducing quality of videos. Imagine common scenes in life, such as: (i) Fans recording videos of their idol’s face on a concert screen are annoyed by distracting banding. (ii) Children filming cartoon scenes on a television are disappointed at the banding. (iii) When a football fan want to catch his favorite player's marvelous performance rebroadcast on the screens live but with black banding blocking the player.

Fundamentally, this banding artifact arises from the synchronization mismatch between temporal emission pattern of a display and capturing mechanism of a camera~\cite{RIFLE}. Despite the prevalence of these screen-induced banding, models specifically targeting this phenomenon remain scarce. Current research primarily focuses on either image-level banding removal~\cite{RIFLE} or ambient light burst flicker removal~\cite{flickerformer}. Moreover, directly applying other video restoration networks, such as video super-resolution~\cite{DLoRAL}, video deblurring~\cite{zhong2025compevent}, video denoising~\cite{STBN}, and video demoiréing~\cite{oh2025fpanet}, fails to resolve the issue effectively, either leaving residual banding or resulting in over-smoothed spatial details.

There are two primary obstacles in video deflickering. First, there is a lack of publicly available video datasets for screen-induced deflickering, making it difficult to train and evaluate targeted models. Second, there is no existing model exclusively designed for video deflickering. Existing methods fail primarily due to a mismatch in degradation complexity because real-world captures frequently contain two different kinds of flicker-banding artifacts occurring simultaneously.

To address these challenges, we first tackle the data scarcity and degradation mismatch. We establish a real-world testing dataset in a wide range of common scenes in daily life, the \textbf{De}flickering \textbf{Vi}deo \textbf{D}ataset (DeViD). Furthermore, we introduce a physics-informed Degradation Field Modeling (DFM) based on the rolling shutter mechanism. Unlike previous uniform degradations, our DFM module synthesizes the complex multi-banding scenarios, just like in the real life.

Building upon this foundation, we propose VDFP (\textbf{V}ideo \textbf{D}eflickering with \textbf{F}licker-banding \textbf{P}riors), a novel two-stage framework. Our first stage introduces a Spatial-Temporal Continuous Prior Perception (CPP) module. Instead of binary segmentation, this module is optimized via a Flicker-Aware Mean Squared Error (FA-MSE) to predict a continuous confidence map. In the second stage, we integrate this spatial-temporal prior into a video diffusion model. By zero-initializing an augmented input layer to accept the prior, our model preserves pre-trained generative capabilities while adaptively guiding the restoration process. Figure~\ref{fig:main_picture} shows our datasets and visual comparisons on real-world data.

Although VDFP demonstrates exceptional effectiveness, a limitation lies in the hardware diversity of the DeViD dataset, which exclusively contains banding captured from LED matrices. Consequently, banding induced by other displays, such as LCD and OLED, remains for future investigation.

In summary, our main contributions are as follows:
\vspace{-6pt}
\begin{itemize}
    \vspace{-1mm}
    \item We construct a real-world video deflickering dataset (DeViD) containing various scenes.
    \vspace{-1mm}
    \item We propose a degradation field modeling pipeline (DFM) based on rolling shutter mechanism, synthesizing multi-banding scenarios in real life, to address lack of training datasets.
    \vspace{-1mm}
    \item We introduce a spatial-temporal continuous prior perception (CPP) module with FA-MSE and a zero-initialized condition injection, successfully removing nearly all banding.
    \vspace{-1mm}
    \item Extensive experiments demonstrate that our model (VDFP) significantly outperforms other current methods (see Fig.~\ref{fig:visual_comparison_one}), which deal with other video restoration tasks.
\end{itemize}
\begin{figure}[t!]       
    \centering
    \includegraphics[width=\textwidth]{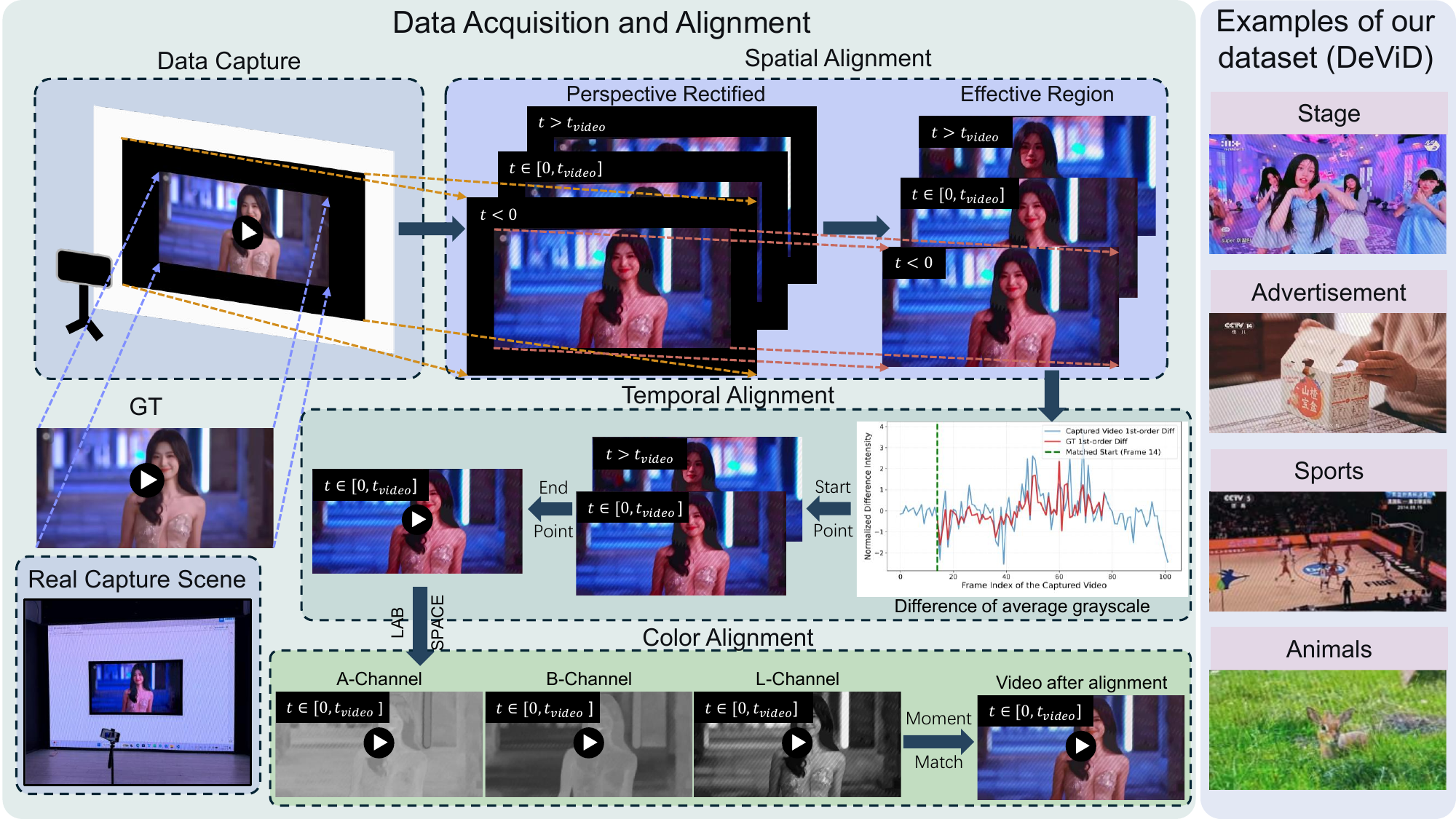}
    \vspace{-5mm}
    \caption{Overview of the data acquisition and alignment of DeViD (left) and examples of some scenes contained in DeViD, such as animals, sports, advertisement, and stages etc (right). The left details the capture process, followed by the spatial (border cropping and stretching) alignment, temporal (grayscale difference) alignment, and color (LAB space) alignment.}
    \label{fig:top}
    \vspace{-5mm}
\end{figure}

\vspace{-5mm}
\section{Related Work}
\vspace{-4mm}
\subsection{Diffusion Model}
\vspace{-3mm}
Diffusion model has gradually become the mainstream method in restoration and generation tasks, substituting GANs~\cite{GANs,GANs_2,GANs_3,GANs_4} and transformers~\cite{transformer,transformer2,transformer3}~\cite{Survey_On_Diffusion}. The early-stage video diffusion model turns the basic U-Net~\cite{Unet} to the 3D U-Net, such as VDM~\cite{VDM}. However, processing videos directly in the pixel space may cause great computational burden. To address this, researchers have introduced Latent Diffusion Models~\cite{LDM}, which significantly enhance generation efficiency by conducting spatiotemporal modeling of features within a low-dimensional latent space. With the increase in the demand of long video generation and temporary consistency, the underlying model of video diffusion models have gradually shifted from U-Net to Transformer, such as Sora~\cite{Sora}. Due to effectiveness of diffusion models, they are also applied to video restoration.

\vspace{-4mm}
\subsection{Video Demoiréing}
\vspace{-3mm}
In digital photographing, moiréing occurs when the grid of a display screen aligns imperfectly with the sensor array of a camera~\cite{digital_demoire}. Early studies establish the paradigm of processing in spatial and temporal areas by introducing temporal consistency loss~\cite{2022ECCVdemoire}. However, because moiréing may influence explicit alignment computations such as optical flow, researchers begin to find multi-dimensional feature alignment strategy, such as direction-aware~\cite{DTNet} and frame-level post-alignment modules~\cite{oh2025fpanet}. To further alleviate computational burdens, network architectures shift toward alignment-free designs~\cite{sung2025mocha}, or adopt Mamba~\cite{xu2025alignmentfreerawvideodemoireing}. Others trace back to physical mechanisms. They leverage the linear characteristics of RAW domain data~\cite{yue2023recaptured}, combine RAW and YCbCr dual color spaces~\cite{YCb}, or introduce physical optical priors through a focused-defocused dual-camera system~\cite{DuDemoire2025}.

\vspace{-4mm}
\subsection{Video Deflickering}
\vspace{-3mm}
Flicker-banding manifests as periodic luminance stripes in captured videos, primarily caused by temporal mismatches between camera acquisition and light source modulation~\cite{RIFLE}. More details are in the Appendix~\ref{FB}. Video deflickering is less explored compared to other restoration tasks, such as video demoiréing. The current research related to deflickering are RIFLE~\cite{RIFLE},the one for image deflickering, and BurstDeflicker~\cite{BurstDeflicker_lishenqu} and Flickerformer~\cite{flickerformer}, the ones targeted for ambient light deflickering. All these studies are much easier than video deflickering.

\vspace{-4mm}
\section{Preliminary}
\vspace{-3mm}
Flicker-banding is a common visual artifact when digital cameras capture electronic screens, typically as spatial distortions such as alternating bright and dark striations, grid or curved patterns. Intensity and geometry of bands are highly sensitive to camera exposure settings. 
This phenomenon results from a severe temporal desynchronization between acquisition system and display medium~\cite{RIFLE}. 

This asynchronous interaction stems from inherent mismatch between two devices. On the capture end, the modern CMOS sensors employ a rolling shutter mechanism, which exposes and reads the scene sequentially row by row rather than instantaneously~\cite{durini2019high}. This introduces a progressive temporal delay across spatial dimensions. Conversely, on the emission end, modern electronic displays (including LCDs, OLEDs utilizing Pulse-Width Modulation~\cite{geffroy2006organic}, and multiplexed LED matrices) do not provide constant illumination. Instead, they rapidly modulate their light output at specific high frequencies to control perceived human-eye brightness.

\begin{figure}[t!]       
    \centering
    \includegraphics[width=\textwidth]{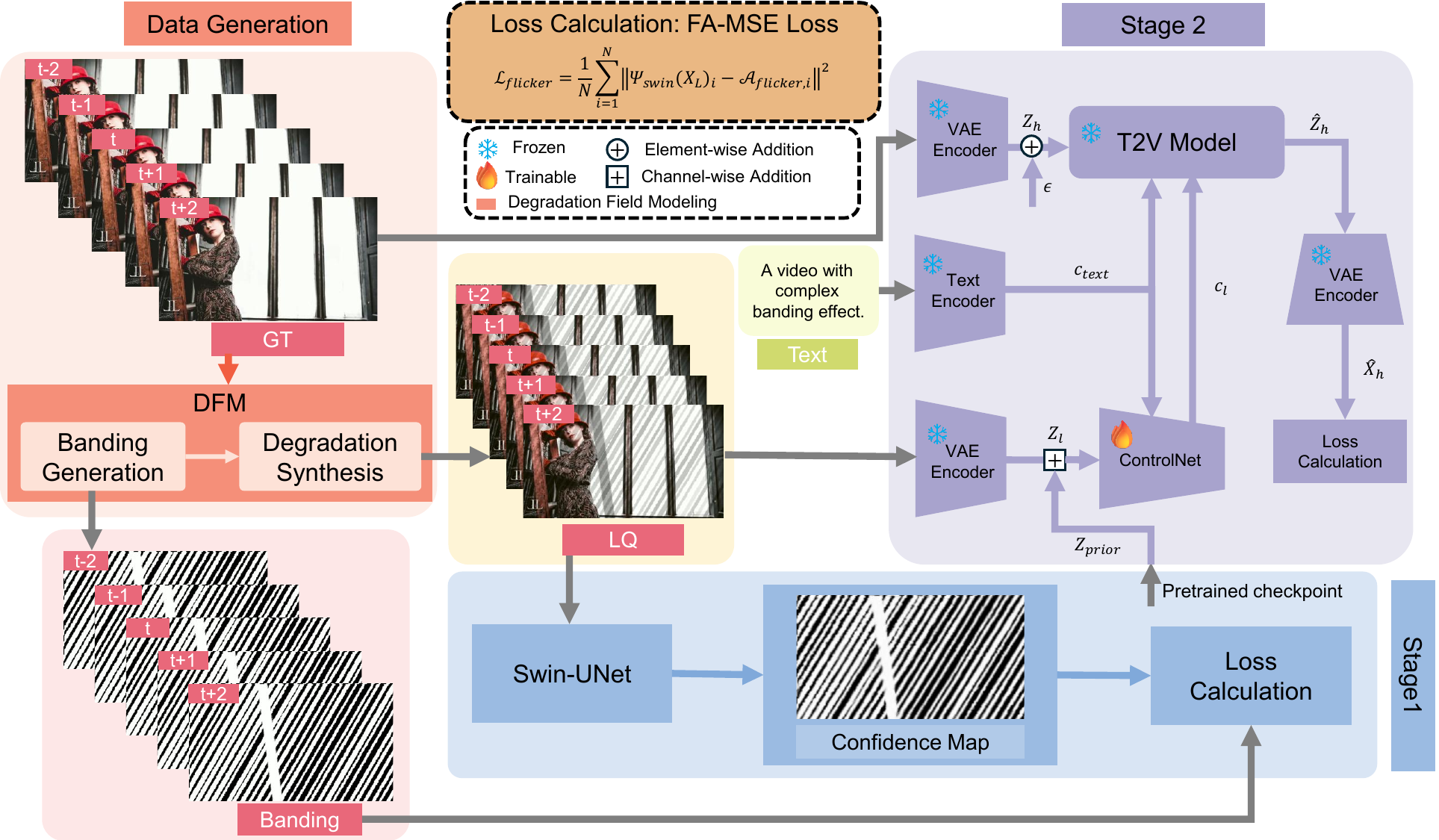}
    \vspace{-4mm}
    \caption{Overview of our proposed model (VDFP). The first stage is the training process of the banding mask prediction module based on the Swin-Unet model. The second stage is the deflickering model based on the STAR model.The first stage of pretrained checkpoint is added to the second stage.}
    \label{fig:main_process}
    \vspace{-6mm}
\end{figure}

\vspace{-3mm}
\section{Dataset: DeViD}
\vspace{-4mm}
Since there're no video datasets for flicker-banding caused by mismatch between screens and cameras, we create a dataset called \textbf{De}flickering \textbf{Vi}deo \textbf{D}ataset (DeViD). The overview of our dataset (DeViD), data acquisition, and data alignment are shown in Fig.~\ref{fig:top}. More examples are in the Appendix~\ref{section:DeViD}.

\vspace{-10pt}
\subsection{Data Collection}
\vspace{-8pt}
In order to guarantee the universality of our dataset, we collect videos in a wide range of scenes. We totally collect 108 videos, 60 of which are videos directly collected from the video super-resolution datasets and 48 of which are newly collected in the common scenes of daily life by ourselves.

Firstly, we collect videos from the official video super-resolution datasets, such as SPMCS~\cite{tao2017spmc}, YouHQ~\cite{zhou2024upscaleavideo} and UDM10~\cite{PFNL}. These datasets are very important because they provide insights about what kind of elements need considering in video process, such as color contrast, scene brightness and darkness, and video length and so on. But only with these datasets are not enough because our task, video deflickering, is a very new task and we need to collect more videos in more various scenes.

Next, to better contain common scenarios in daily life, we collect videos from various websites. Considering common usage of the LED matrix displays in our daily life, we decide to contain more scenarios including stages, red carpets, cartoons, movies, TV series, sports competitions, advertisement, and video games. Rich diversity in scene content ensures that models trained on our dataset can robustly generalize to complex, real-world deflickering tasks.

\vspace{-10pt}
\subsection{Capture Procedure}
\vspace{-8pt}
To capture banding in our real life, we firstly design a website with the background in white and surrounded by thick black frames. We then project the website onto the LED display and fix the smartphone on a mobile phone stand, stimulated by the data capture procedure of the UHDM~\cite{UHDM}. With all these settings, we can control the camera and videos at the same time. The position of the mobile phone stand is changed every several shots to capture more kinds of banding artifacts.

Since the captured videos include a larger field of view than the original content displayed on the LED screen (e.g., the black borders and white background surrounding the videos) and has a longer duration due to the manual start/stop of recording before and after the video playback, the accurate spatial, temporal alignment, and color alignment is equipped for further comparison.

\vspace{-4.5mm}
\section{Method: VDFP}
\vspace{-4mm}
Established on our baseline, STAR~\cite{star}, we introduce degradation field modeling based on rolling shutter mechanism (DFM) and spatial-temporal continuous prior perception (CPP) modules to better address the banding removal task in videos. The main process of our model can be seen in Fig.~\ref{fig:main_process}.
\vspace{-12pt}
\subsection{Degradation Field Modeling (DFM)}
\vspace{-8pt}

\begin{figure}[t!]       
    \centering
    \includegraphics[width=\textwidth]{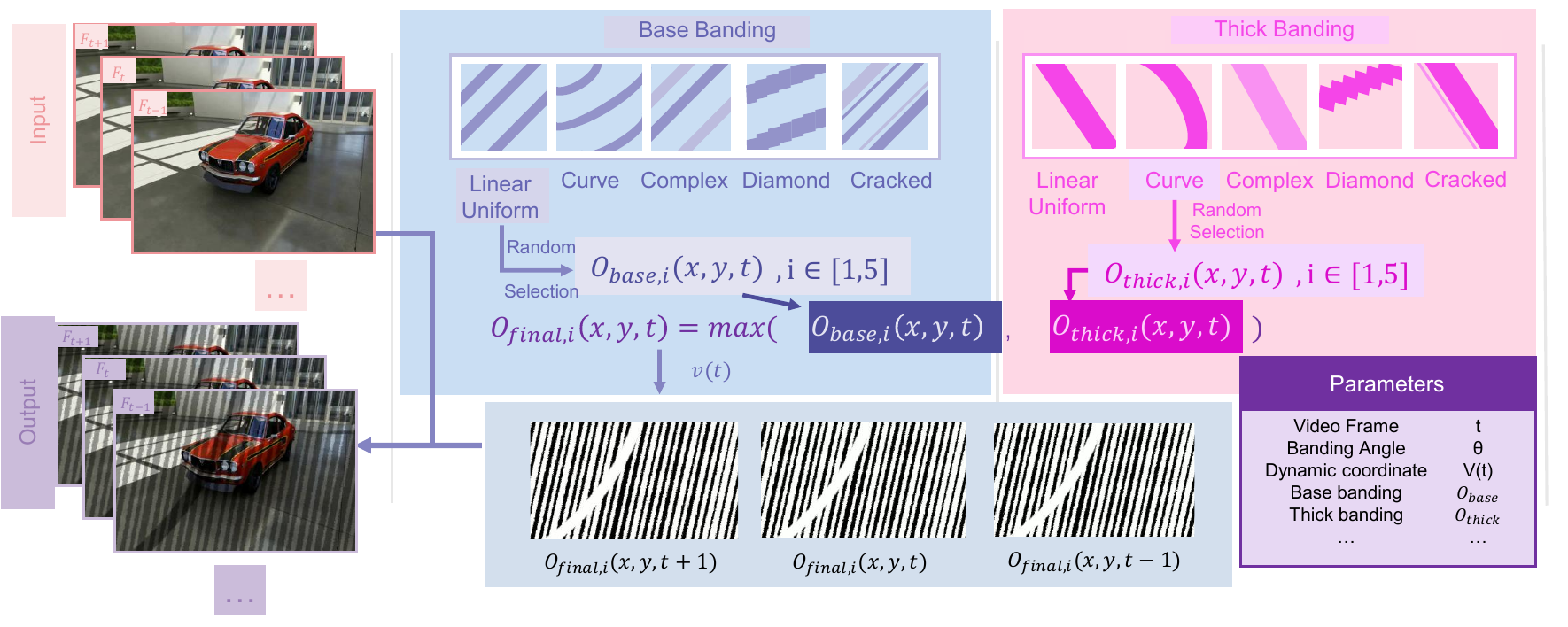}
    \vspace{-6mm}
    \caption{Overview of our data simulation pipeline.}
    \label{fig:simulation_pipeline}
    \vspace{-6mm}
\end{figure}

Since intensity and geometry of flicker-banding is sensitive to  exposure of cameras and the capture perspective, some types of banding patterns may be hard to capture in daily life. So we propose a degradation field modeling based on the one processing images in RIFLE~\cite{RIFLE} mainly in the two following perspective. The process of our simulation pipeline can be seen in Fig.~\ref{fig:simulation_pipeline}.

\textbf{Kinematic Spatiotemporal Modeling}. To accurately simulate the real-world situation, we generate a parameterized spatiotemporal degradation field $O_l(x, y, t)$ (where $l \in \{base, thick\}$) and establish a temporal kinematic vector field. First, to determine which stripe governs any given pixel $(x, y)$ on the screen, we compute the inner product of the relative center coordinates $(x - x_c, y - y_c)$ and the normal vector $(-\sin\theta, \cos\theta)$ perpendicular to the stripe tilt angle $\theta$. This yields the static projection distance $v_{static}$ of the pixel on the orthogonal axis $v$:
\begingroup
\setlength{\abovedisplayskip}{2pt}   
\setlength{\belowdisplayskip}{2pt}   
\begin{equation}
v_{static} = -(x - x_c) \sin\theta + (y - y_c) \cos\theta.
\end{equation}
\endgroup

To simulate the dynamic effects of smartphones' rolling shutter mechanism, we introduce a constant-velocity motion model $(v_x, v_y)$ with frame index $t$. This motion is similarly projected onto the orthogonal axis to form a time-dependent phase shift $\Delta v(t)$:
\begingroup
\setlength{\abovedisplayskip}{2pt}   
\setlength{\belowdisplayskip}{2pt}
\begin{equation}
    \Delta v(t) = -(v_x \cdot t) \sin\theta + (v_y \cdot t) \cos\theta.
\end{equation}
\endgroup
So the coordinate in each frame can be written as $v(t) = v_{static} - \Delta v(t)$. 

\textbf{Dual-layer Banding Fusion}. To better simulate the real-world situation where two types of flicker-banding artifacts exist in one frame, we propose a dual-layer architecture: base stripe layer for high-frequency banding, and thick stripe layer for thick banding.

\begin{figure}[t!]       
    \centering
    \includegraphics[width=\textwidth]{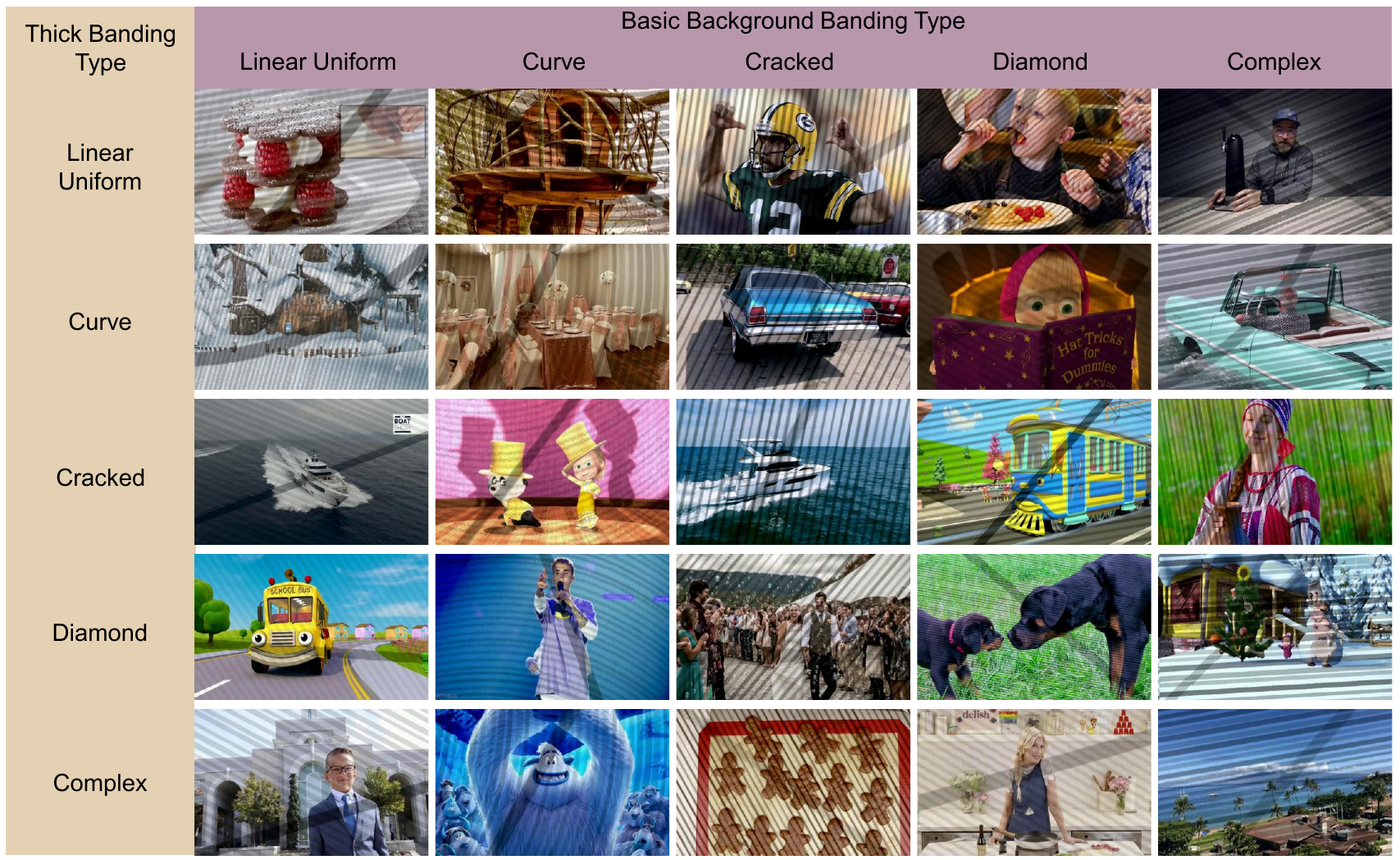}
    \vspace{-6mm}
    \caption{Overview of results of our simulation pipeline. The row and column respectively mean the type of base banding and thick banding. Numbers are different types of thick banding: 1:linear uniform; 2:curve; 3:cracked; 4:diamond; 5:complex.}
    \label{fig:simulation_dataset}
    \vspace{-4mm}
\end{figure}

For any single layer (with period $T$ and width $W$), its generation $O(x,y,t)$ apply a smoothstep function. We first calculate physical boundary distance from pixel to the nearest stripe center $v_c$.
\begingroup
\setlength{\abovedisplayskip}{-1pt}   
\setlength{\belowdisplayskip}{-1pt}   
\begin{equation}
d(x,y,t) = |v(t) - v_c| - \frac{W}{2}.
\end{equation}
\endgroup
Subsequently, a smoothstep function $S(\cdot)$ with an edge feathering threshold $f$ is applied to simulate the soft edges caused by optical defocusing in the real situation:
\begingroup
\setlength{\abovedisplayskip}{2pt}   
\setlength{\belowdisplayskip}{2pt}   
\begin{equation}
O(x,y,t) = 1.0 - S\left(-f, +f, d(x,y,t)\right).
\end{equation}
\endgroup
After obtaining two independent banding $O_{\text{base}}$ and $O_{\text{thick}}$, we adopt maximum for the total banding:
\begingroup
\setlength{\abovedisplayskip}{2pt}   
\setlength{\belowdisplayskip}{2pt}   
\begin{equation}
O_{\text{final}}(x, y, t) = \max(O_{\text{base}}(x, y, t), O_{\text{thick}}(x, y, t)).
\end{equation}
\endgroup
Figure~\ref{fig:simulation_dataset} shows the simulated examples, which cover different base and thick bandings. More details of each type of banding are provided in the Appendix~\ref{section:DFM}.

\begin{figure}[t!]
\centering

\begin{subfigure}[b]{0.23\textwidth}
    \centering
    \includegraphics[width=\linewidth]{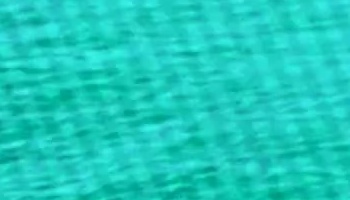}
\end{subfigure}
\hspace{2pt}
\begin{subfigure}[b]{0.23\textwidth}
    \centering
    \includegraphics[width=\linewidth]{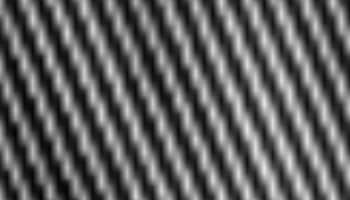}
\end{subfigure}
\hspace{2pt}
\begin{subfigure}[b]{0.23\textwidth}
    \centering
    \includegraphics[width=\linewidth]{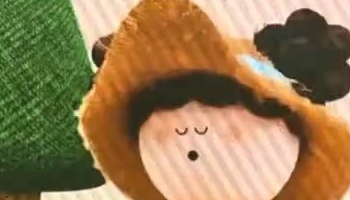}
\end{subfigure}
\hspace{2pt}
\begin{subfigure}[b]{0.23\textwidth}
    \centering
    \includegraphics[width=\linewidth]{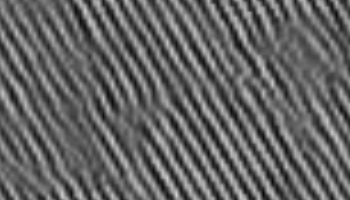}
\end{subfigure}

\begin{subfigure}[b]{0.23\textwidth}
    \centering
    \includegraphics[width=\linewidth]{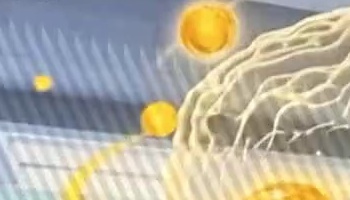}
    \caption*{LQ}
\end{subfigure}
\hspace{2pt}
\begin{subfigure}[b]{0.23\textwidth}
    \centering
    \includegraphics[width=\linewidth]{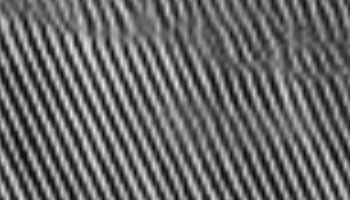}
    \caption*{Confidence Map}
\end{subfigure}
\hspace{2pt}
\begin{subfigure}[b]{0.23\textwidth}
    \centering
    \includegraphics[width=\linewidth]{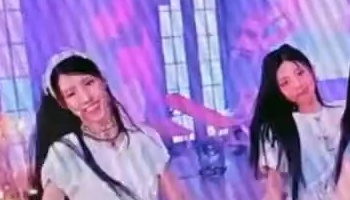}
    \caption*{LQ}
\end{subfigure}
\hspace{2pt}
\begin{subfigure}[b]{0.23\textwidth}
    \centering
    \includegraphics[width=\linewidth]{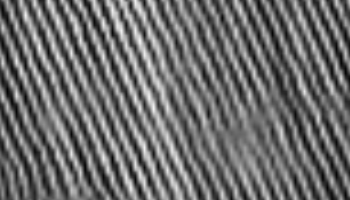}
    \caption*{Confidence Map}
\end{subfigure}

\vspace{-2mm}
\caption{Examples of predicted confidence maps of our real-world dataset (DeViD). Colors in the confidence maps mean that the deeper the color is, the less likely having banding.}
\label{fig:mask_img}
\vspace{-6mm}

\end{figure}

\vspace{-3mm}
\subsection{Spatial-Temporal Continuous Prior Perception (CPP)}
\vspace{-2mm}
To better eliminate the flicker-banding artifacts in the videos, we propose a two-stage restoration framework that decouples artifact prior perception (see Fig.~\ref{fig:mask_img}) from guided generation based on the instance segmentation model Video-SwinUNet~\cite{swinunet} and our baseline STAR~\cite{star}.

\textbf{Flicker-Aware Mean Squared Error (FA-MSE)}. Video banding are characterized by subtle, periodic luminance transitions that discrete binary loss $\{0, 1\}$ fail to represent accurately. To capture full spectrum of these distortions, including the edges of stripes, we define a flicker-intensity loss.

Specifically, though Video-SwinUNet~\cite{swinunet} performs outstandingly in the medical area, to better match our task, we adapt the architecture to output a continuous artifact confidence map $\hat{A}_{flicker} \in [0, 1]$. We optimize the predictor using a Flicker-Aware Mean Squared Error (FA-MSE) $\mathcal{L}_{flicker}$:
\begingroup
\setlength{\abovedisplayskip}{2pt}   
\setlength{\belowdisplayskip}{2pt}   
\begin{equation}
\mathcal{L}_{flicker} = \frac{1}{N} \sum_{i=1}^{N} \| \Psi_{Swin}(X_L)_i - \mathcal{A}_{flicker, i} \|^2,
\end{equation}
\endgroup
where $\Psi_{Swin}(X_L)$ means the predicted artifact spatial-temporal confidence map and $\mathcal{A}_{flicker}$ means the ground-truth luminance-fluctuation amplitude map.

\begin{figure}[t!]     
    \centering
    \includegraphics[width=\textwidth]{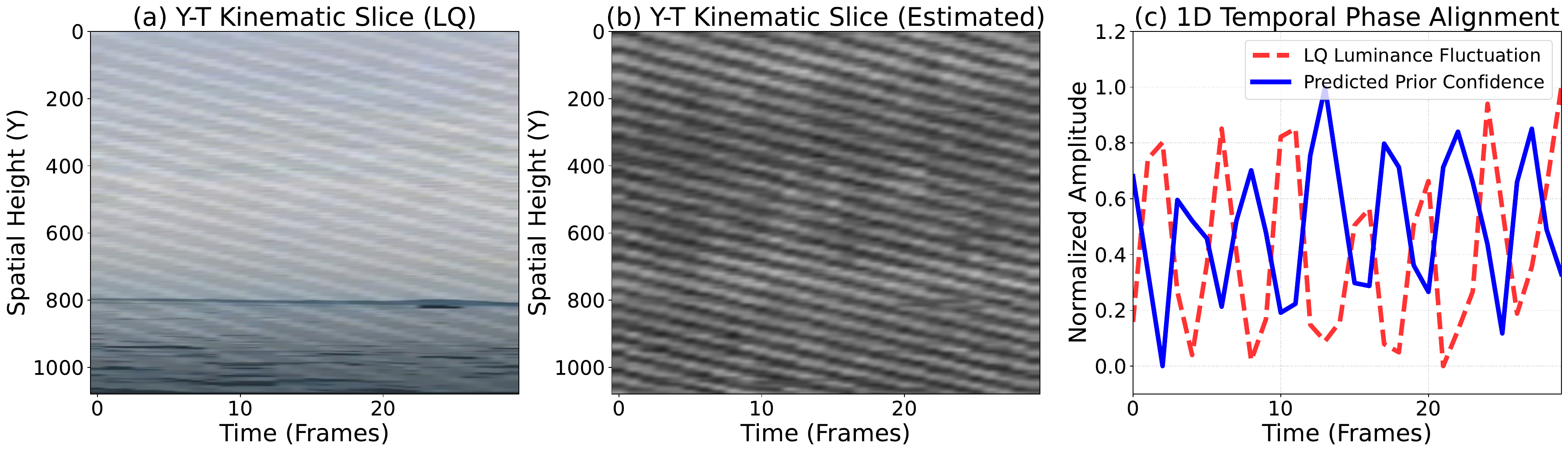}
    \vspace{-7mm}
    \caption{Spatio-temporal kinematic tracking and phase alignment. (a-b) Y-T spatio-temporal slices, extracted from the central vertical line, reveal that our estimated prior successfully tracks downward trajectory of diagonally upward banding. (c) 1D temporal tracking of the center pixel exhibits alignment between the input luminance drops (\textcolor{red}{red valleys}) and our predicted confidence (\textcolor{blue}{blue peaks}).}
    \label{fig:accuracy}
    \vspace{-7mm}
\end{figure}
 
Some examples for artifact prior perception can be seen in Fig.~\ref{fig:mask_img}. Since confidence maps only need to tell the general position and width of banding, they are not completely the same as those in LQ. As proved in Fig.~\ref{fig:accuracy}, our estimated prior perception achieves exceptional temporal accuracy.

\textbf{Banding-Aware Prior Injection}. To introduce this structural prior into the highly optimized pre-trained diffusion model without inducing catastrophic distribution shifts, we propose a Banding-Aware Condition Injection strategy. Let the original weight of the base model be $W_4$, which is designed to process the 4-channel degraded latent feature $X$. We expand the input dimensionality to 5 channels by concatenating $X$ with the 1-channel prior map $\hat{A}_{flicker}$, yielding the augmented input $[X, \hat{A}_{flicker}]$. To ensure optimization stability during the initial training phase, the augmented 5-channel weight matrix $W_5$ is strictly initialized with zeros for the newly appended dimension: $W_5 = [W_4, 0]$.

This zero-initialization mathematically guarantees that, at the very first step of training, this pre-trained generative model can remain entirely undisturbed with original knowledge:
\begingroup
\setlength{\abovedisplayskip}{2pt}   
\setlength{\belowdisplayskip}{2pt}
\begin{equation}
W_5 [X, \hat{A}_{flicker}] = W_4 X + 0 \cdot \hat{A}_{flicker} = W_4 X.
\end{equation}
\endgroup
As training continues, the model learns to utilize the injected prior channel $\hat{A}_{flicker}$ as attention map. 

By explicitly guiding the model with banding priors, the diffusion model selectively suppresses flicker-banding without over-smoothing the clean background contents, thereby striking an optimal balance between the aggressive artifact removal and high-fidelity structural background preservation.

\vspace{-5mm}
\section{Experiments}
\vspace{-3mm}
\subsection{Experiment Setup}
\vspace{-3mm}
\textbf{Data Construction}. The dataset used for our degradation field modeling is HQ-VSR~\cite{chen2025dove}, a large collection of wide range of scenes in our daily life as well as cartoons, which well matches everyday needs. The test dataset we construct, DeViD, collects some videos from famous video super-resolution datasets, SPMCS~\cite{tao2017spmc}, YouHQ~\cite{zhou2024upscaleavideo}, and UDM10~\cite{PFNL}. It's worth mentioning that the videos we collect by ourselves have different resolution to better cover the real-life situation.

\vspace{-0.5mm}
\noindent \textbf{Evaluation Metrics}. The reference-based evaluation metrics used here are SSIM~\cite{wang2004image}, LPIPS~\cite{zhang2018perceptual}, VMAF~\cite{VMAF}, and DISTS~\cite{DISTS}. No-reference evaluation metrics used here are MUSIQ~\cite{MUSIQ}, CLIP-IQA~\cite{CLIP-IQA}, BRISQUE~\cite{BRISQUE}, $E_{warp}^*$ (specifically, refers to $E_{warp}$($\times 10^{-3}$))~\cite{Lai-ECCV-2018}, and DOVER~\cite{wu2023dover}. Among these metrics, $E_{warp}^*$~\cite{Lai-ECCV-2018} and DOVER~\cite{wu2023dover} are metrics evaluating videos.

\vspace{-0.5mm}
\noindent \textbf{Implementation Details}. In our experiments, we freeze pre-trained text-to-video (T2V) model and exclusively finetune the Video-ControlNet along with the Local Information Enhancement Modules (LIEM), just like what STAR~\cite{star} does. We utilize the AdamW~\cite{AdamW} optimizer with a learning rate of 2$\times$10$^{-5}$. The training process is performed on the video sequences consisting of sixteen frames at a resolution of 1,280$\times$720. Experiments are implemented on two A6000 GPUs.

\vspace{-0.5mm}
\noindent \textbf{Comparison Methods}. We select methods from a wide range of other tasks, such as DLoRAL~\cite{DLoRAL} in video super-resolution, CompEvent~\cite{zhong2025compevent} in video deblurring, and STBN~\cite{STBN} in video denoising. Moreover, since there are no existing models available with accurately the same deflickering task as we do, the task to eliminate flicker-banding caused by the mismatch between screens and smartphone cameras, we select models similar to our tasks, such as FPANet~\cite{oh2025fpanet} in video demoiréing, which also deals with the problem confronted when photographing display matrices, and Flickerformer~\cite{flickerformer} in burst flicker removal, which has similar phenomenon to deal with as deflickering. 

These models mentioned above for comparison not only cover a wide variety of video restoration tasks, but also include different types of model structures, such as diffusion models, transformers, and one-step models to better verify the effectiveness of our model.

\begin{table}[t!]
\centering
\footnotesize
\caption{Comparison results of our model (VDFP) and other methods trained on our simulation pipeline and tested on our dataset (DeViD). Specifically, Flickerformer* means testing with the official checkpoint. The best and the second best results are respectively colored with \textcolor{red}{red} and \textcolor{blue}{blue}. } 
\resizebox{\linewidth}{!}{

\begin{tabular}{lcccccccc} 
\toprule

\rowcolor{color3} {Method}& {SSIM$\uparrow$} & {LPIPS$\downarrow$} & {MUSIQ$\uparrow$} & {BRISQUE$\downarrow$} & {DISTS$\downarrow$} & {DOVER$\uparrow$} & {$E_{warp}^*$$\downarrow$} \\
\midrule
LQ & 0.6471 & 0.3815 & 43.5214 & 42.9104 & 0.1807 & 0.4850 & 16.5998 \\
STBN~\cite{STBN} & 0.6502 & 0.3865 & 38.2374 & 53.3685 & 0.1911 & 0.3199 & 13.6938 \\
DLoRAL~\cite{DLoRAL} & \textcolor{blue}{0.6745} & \textcolor{blue}{0.3159} & 44.5751 & 41.4641 & \textcolor{blue}{0.1401} & \textcolor{blue}{0.5301} & \textcolor{blue}{12.9694} \\
FPANet~\cite{oh2025fpanet} & 0.6691 & 0.3433 & \textcolor{blue}{45.8296} & 38.8692 & 0.1698 & 0.5219 & 15.2090 \\
CompEvent~\cite{zhong2025compevent} & 0.6564 & 0.3708 & 44.6865 & 41.6570 & 0.1752 & 0.4801 & 16.5878 \\
Flickerformer*~\cite{flickerformer} & 0.6445 & 0.3872 & 42.1644 & 40.5255 & 0.1818 & 0.4647 & 15.1741 \\
Flickerformer~\cite{flickerformer} & 0.6545 & 0.4041 & 43.7394 & \textcolor{blue}{37.6982} & 0.1929 & 0.4446 & 13.5410 \\
\midrule
\rowcolor{violet!20}
VDFP (ours) & \textcolor{red}{0.6974} & \textcolor{red}{0.2958} & \textcolor{red}{48.8910} & \textcolor{red}{36.4535} & \textcolor{red}{0.1378} & \textcolor{red}{0.6484} & \textcolor{red}{10.5442} \\
\bottomrule
\end{tabular}
}
\label{quantitative result}
\vspace{-5mm}
\end{table}

\begin{figure*}[t]
\scriptsize

\centering
\begin{tabular}{ccc}

\hspace{-0.45cm}
\begin{adjustbox}{valign=t}
\begin{tabular}{c}
\includegraphics[width=0.216\textwidth,height=0.231\textwidth]{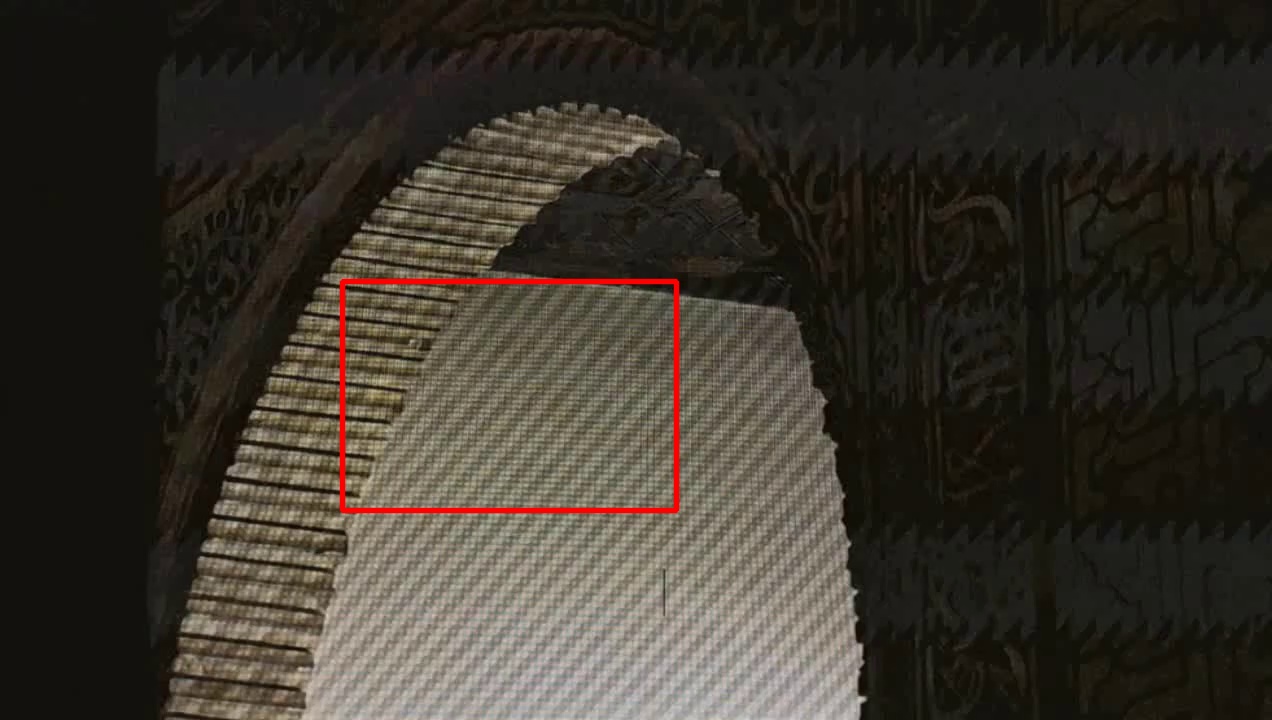}
\\
DeViD: 011
\end{tabular}
\end{adjustbox}
\hspace{-0.46cm}
\begin{adjustbox}{valign=t}
\begin{tabular}{cccccc}
\includegraphics[width=0.149\textwidth]{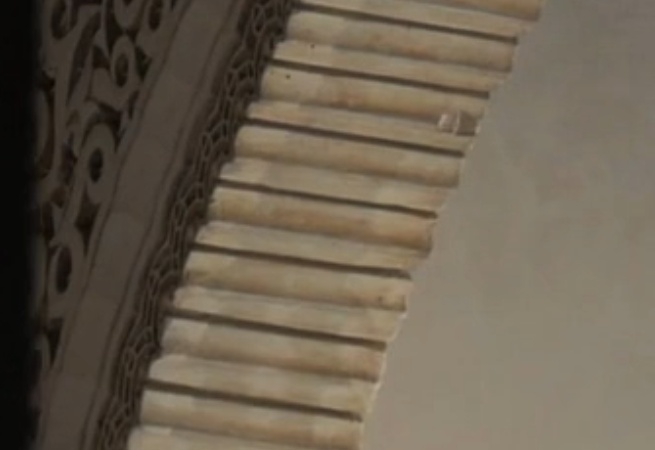} \hspace{-4mm} &
\includegraphics[width=0.149\textwidth]{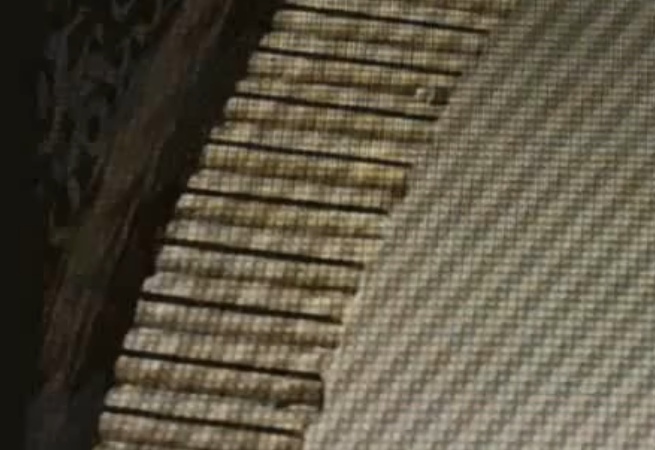} \hspace{-4mm} &
\includegraphics[width=0.149\textwidth]{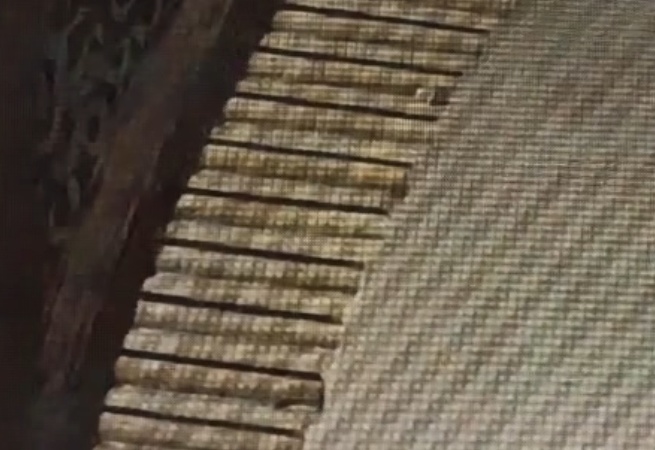} \hspace{-4mm} &
\includegraphics[width=0.149\textwidth]{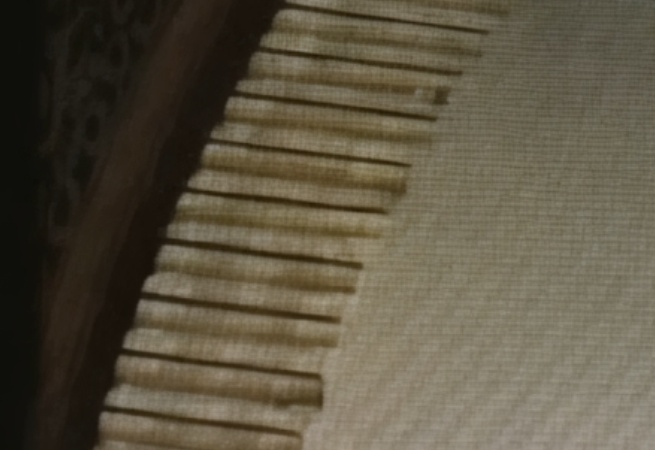} \hspace{-4mm} &
\includegraphics[width=0.149\textwidth]{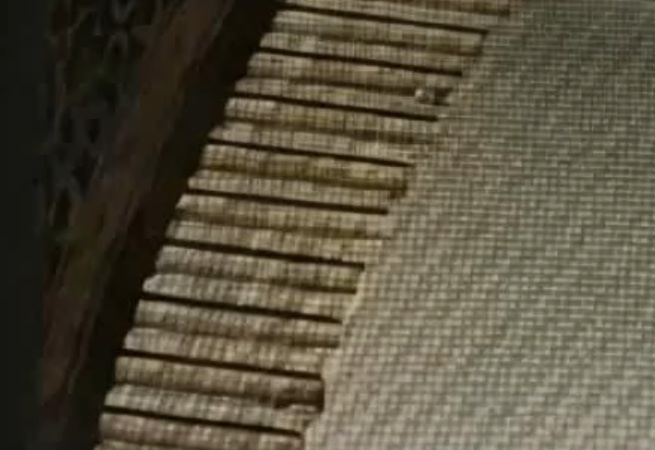} \hspace{-4mm} 
\\
GT \hspace{-4mm} &
LQ \hspace{-4mm} &
STBN~\cite{STBN} \hspace{-4mm} &
STAR~\cite{star} \hspace{-4mm} &
DLoRAL~\cite{DLoRAL} \hspace{-4mm}
\\
\includegraphics[width=0.149\textwidth]{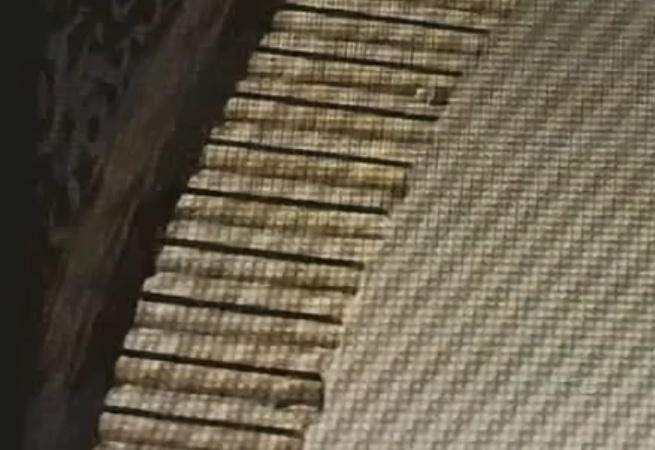} \hspace{-4mm} &
\includegraphics[width=0.149\textwidth]{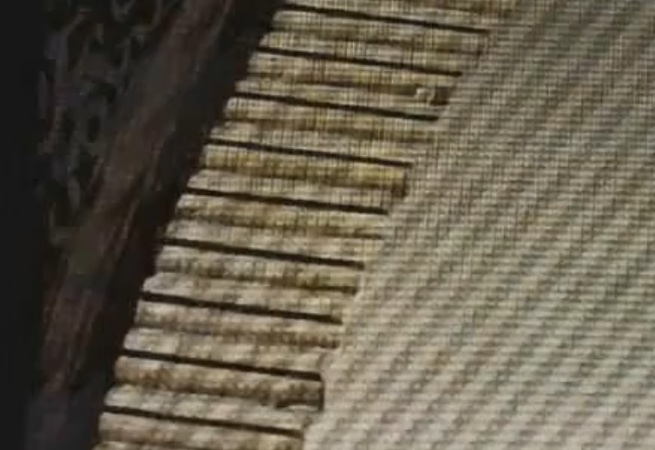} \hspace{-4mm} &
\includegraphics[width=0.149\textwidth]{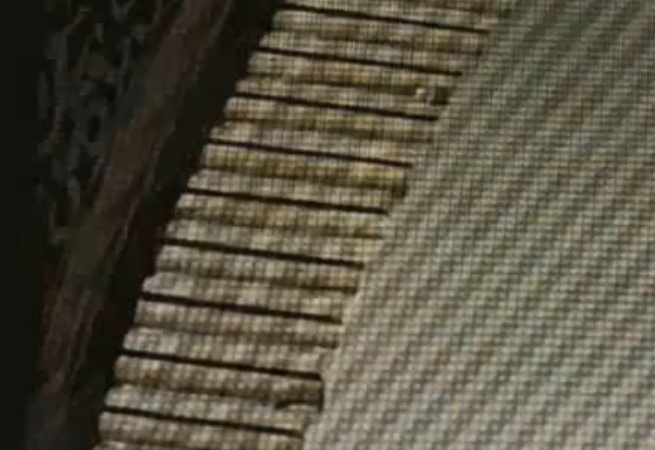} \hspace{-4mm} &
\includegraphics[width=0.149\textwidth]{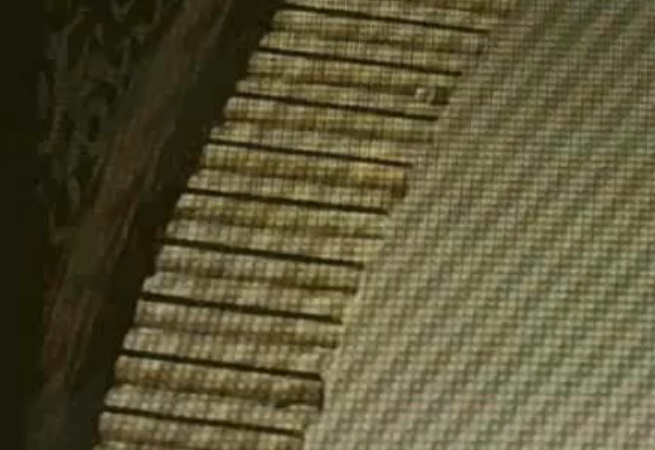} \hspace{-4mm} &
\includegraphics[width=0.149\textwidth]{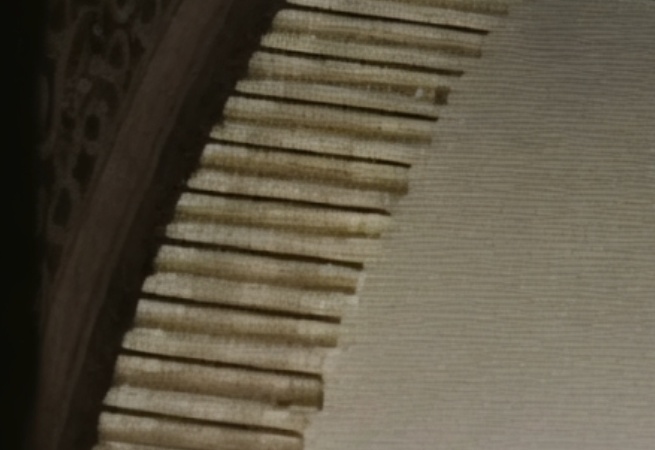} \hspace{-4mm}  
\\ 
FPANet~\cite{oh2025fpanet} \hspace{-4mm} &
CompEvent~\cite{zhong2025compevent} \hspace{-4mm} &
Flickerformer*~\cite{flickerformer} \hspace{-4mm} &
Flickerformer~\cite{flickerformer}  \hspace{-4mm} &
VDFP (ours) \hspace{-4mm}
\\
\end{tabular}
\end{adjustbox}
\\

\hspace{-0.42cm}
\begin{adjustbox}{valign=t}
\begin{tabular}{c}
\includegraphics[width=0.216\textwidth,height=0.231\textwidth]{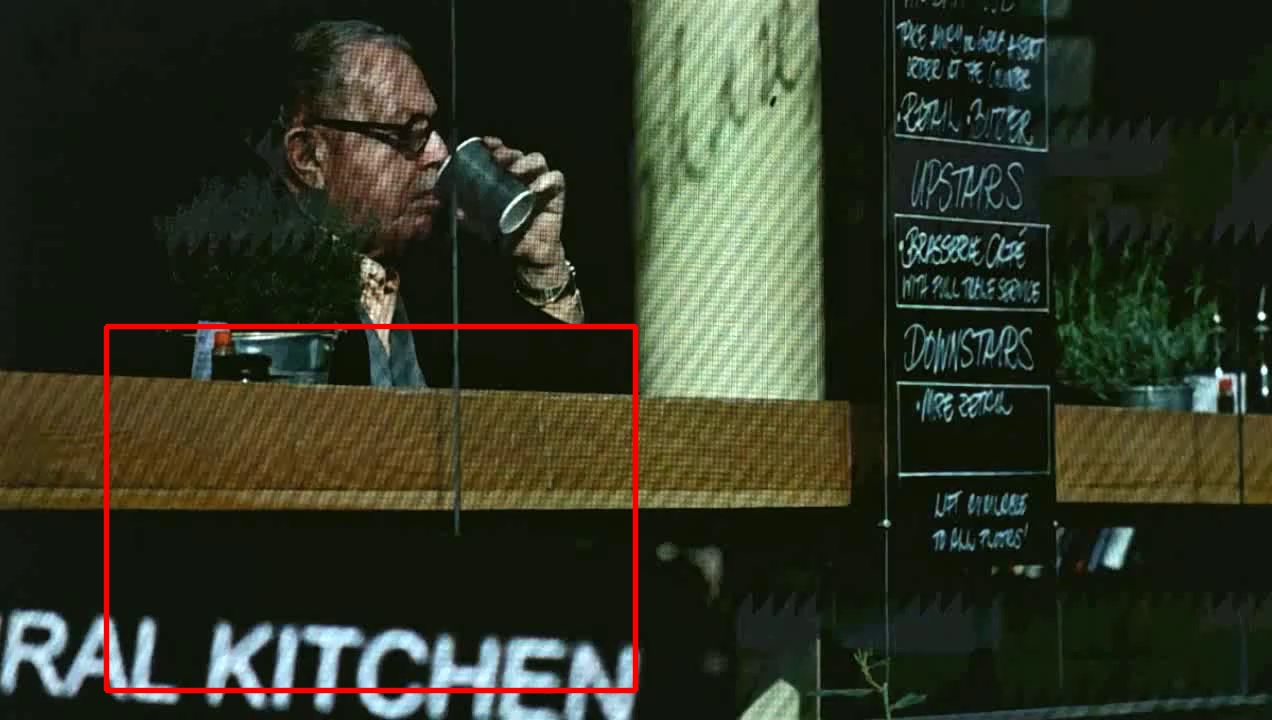}
\\
DeViD: 013
\end{tabular}
\end{adjustbox}
\hspace{-0.46cm}
\begin{adjustbox}{valign=t}
\begin{tabular}{cccccc}
\includegraphics[width=0.149\textwidth]{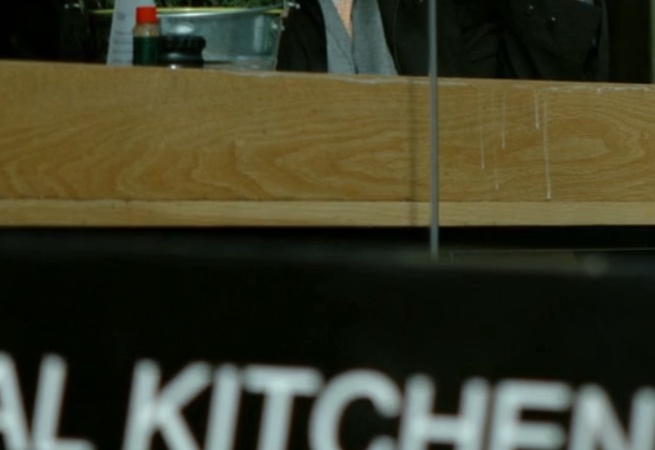} \hspace{-4mm} &
\includegraphics[width=0.149\textwidth]{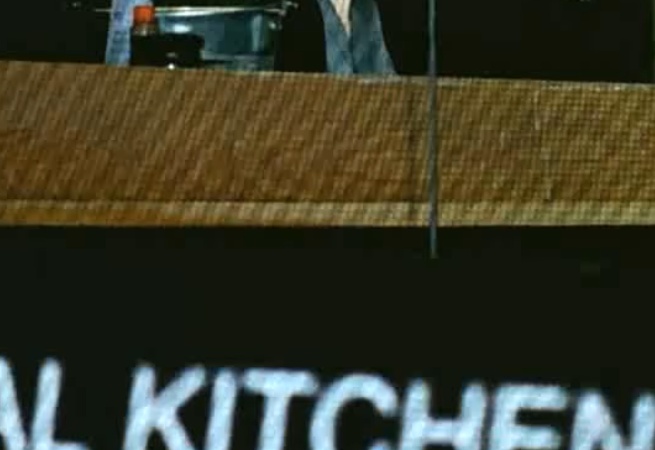} \hspace{-4mm} &
\includegraphics[width=0.149\textwidth]{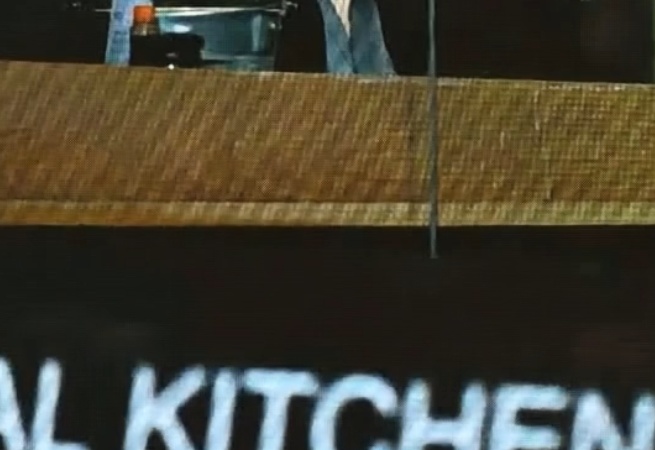} \hspace{-4mm} &
\includegraphics[width=0.149\textwidth]{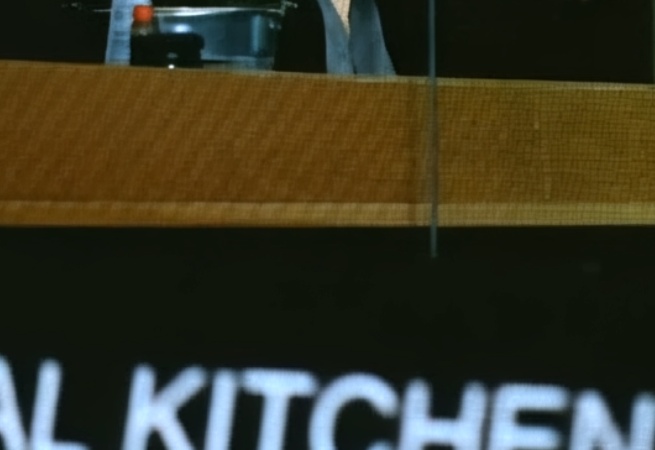} \hspace{-4mm} &
\includegraphics[width=0.149\textwidth]{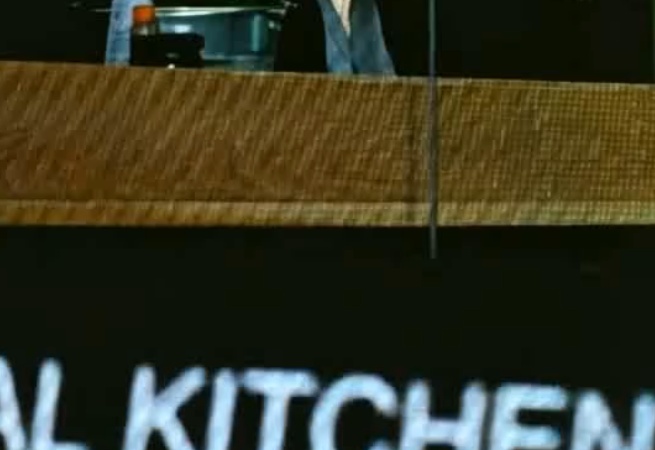} \hspace{-4mm} 
\\
GT \hspace{-4mm} &
LQ \hspace{-4mm} &
STBN~\cite{STBN} \hspace{-4mm} &
STAR~\cite{star} \hspace{-4mm} &
DLoRAL~\cite{DLoRAL} \hspace{-4mm}
\\
\includegraphics[width=0.149\textwidth]{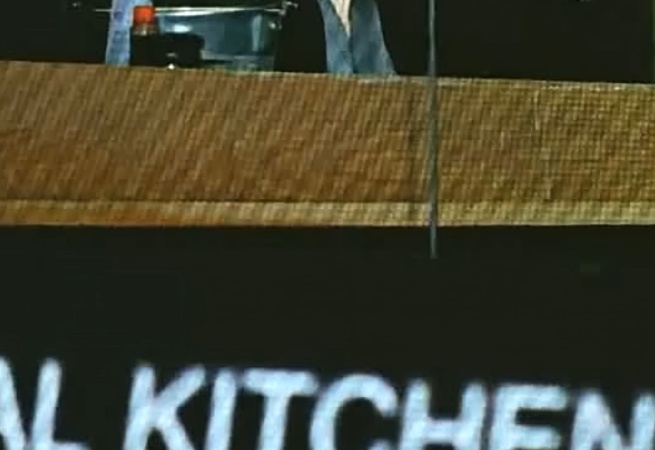} \hspace{-4mm} &
\includegraphics[width=0.149\textwidth]{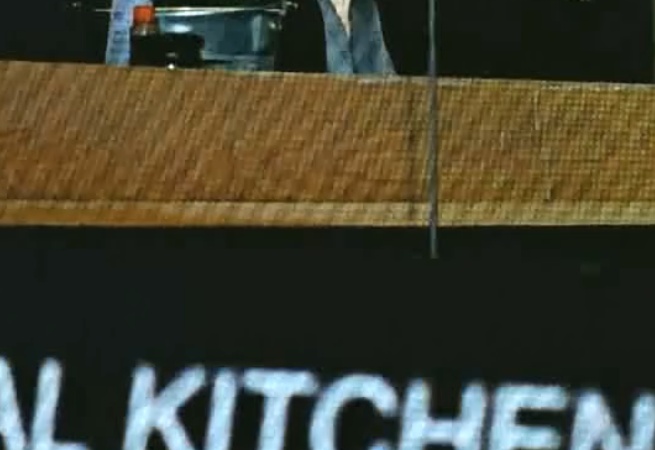} \hspace{-4mm} &
\includegraphics[width=0.149\textwidth]{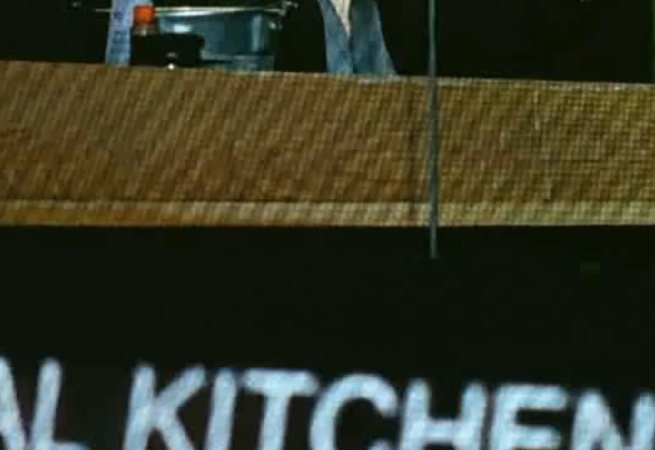} \hspace{-4mm} &
\includegraphics[width=0.149\textwidth]{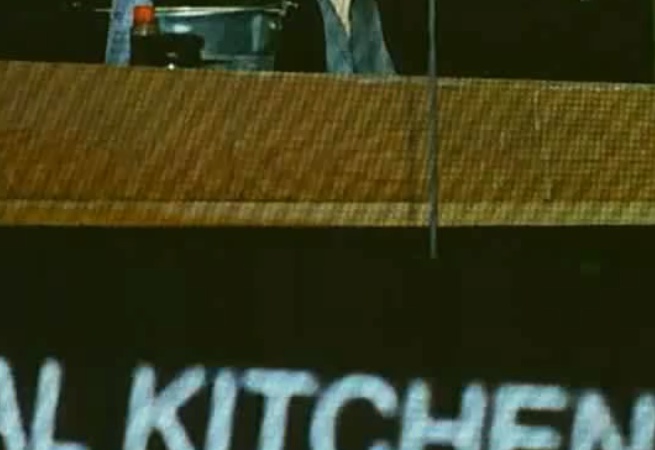} \hspace{-4mm} &
\includegraphics[width=0.149\textwidth]{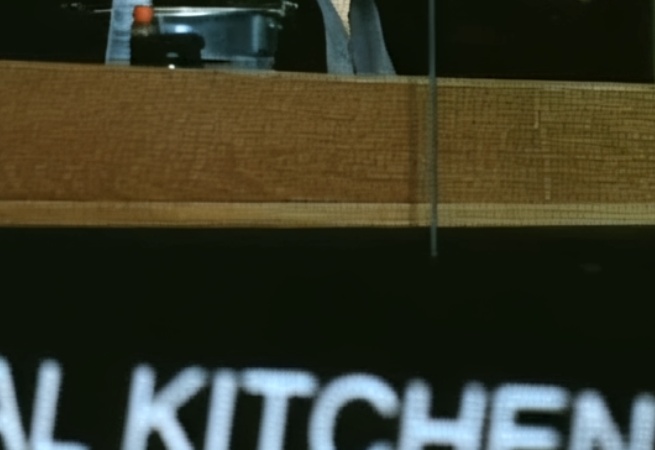} \hspace{-4mm}  
\\ 
FPANet~\cite{oh2025fpanet} \hspace{-4mm} &
CompEvent~\cite{zhong2025compevent} \hspace{-4mm} &
Flickerformer*~\cite{flickerformer} \hspace{-4mm} &
Flickerformer~\cite{flickerformer}  \hspace{-4mm} &
VDFP (ours) \hspace{-4mm}
\\
\end{tabular}
\end{adjustbox}
\\

\hspace{-0.42cm}
\begin{adjustbox}{valign=t}
\begin{tabular}{c}
\includegraphics[width=0.216\textwidth,height=0.231\textwidth]{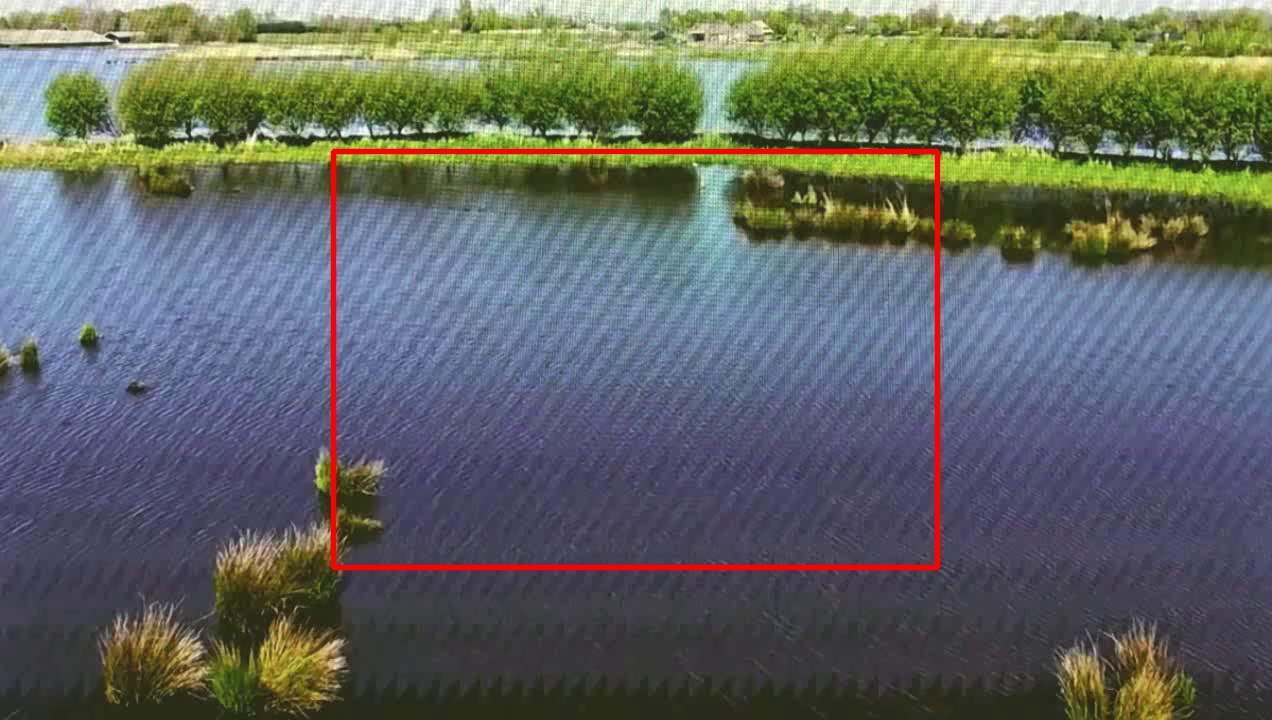}
\\
DeViD: 015
\end{tabular}
\end{adjustbox}
\hspace{-0.46cm}
\begin{adjustbox}{valign=t}
\begin{tabular}{cccccc}
\includegraphics[width=0.149\textwidth]{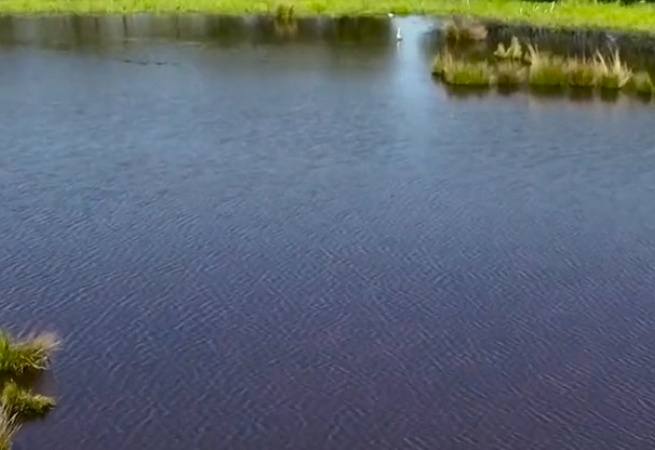} \hspace{-4mm} &
\includegraphics[width=0.149\textwidth]{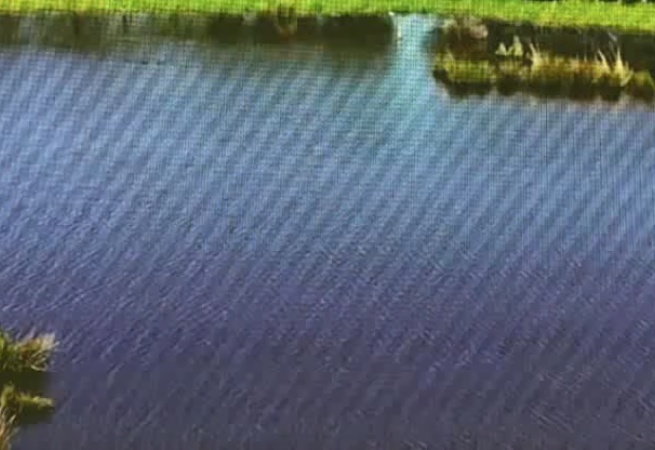} \hspace{-4mm} &
\includegraphics[width=0.149\textwidth]{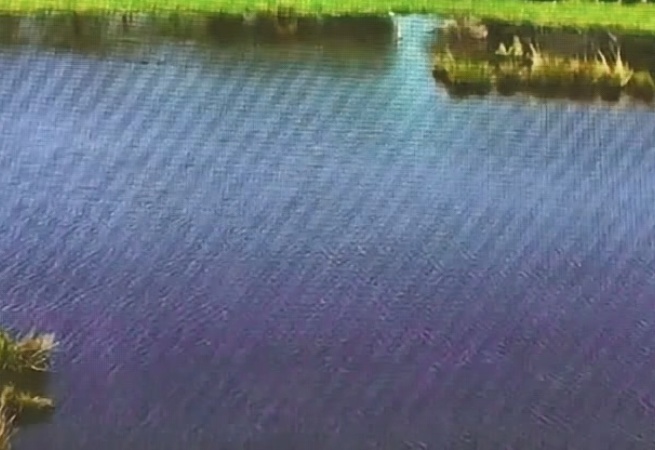} \hspace{-4mm} &
\includegraphics[width=0.149\textwidth]{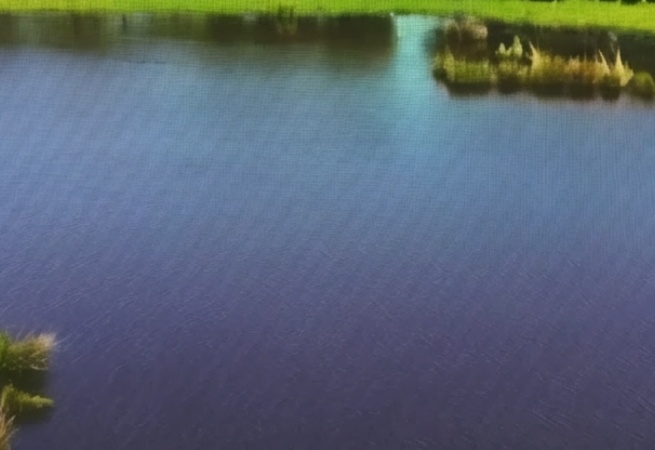} \hspace{-4mm} &
\includegraphics[width=0.149\textwidth]{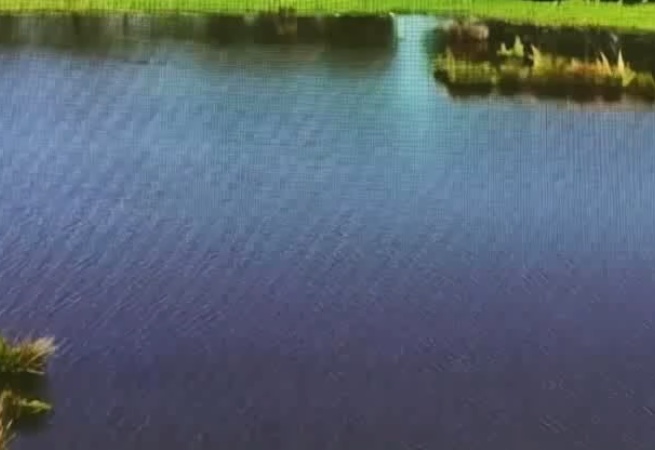} \hspace{-4mm} 
\\
HQ \hspace{-4mm} &
LQ \hspace{-4mm} &
STBN~\cite{STBN} \hspace{-4mm} &
STAR~\cite{star} \hspace{-4mm} &
DLoRAL~\cite{DLoRAL} \hspace{-4mm}
\\
\includegraphics[width=0.149\textwidth]{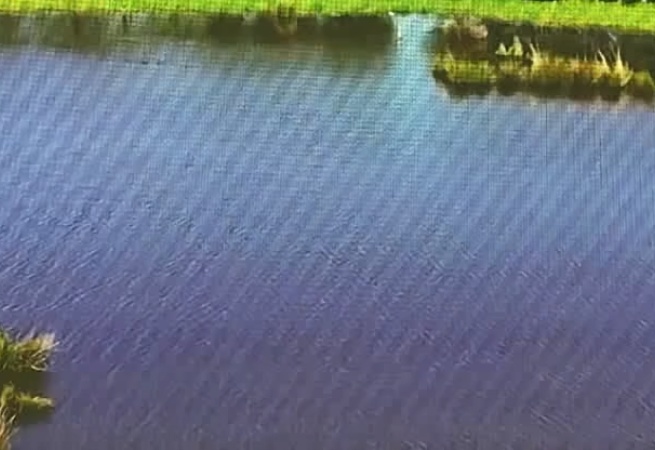} \hspace{-4mm} &
\includegraphics[width=0.149\textwidth]{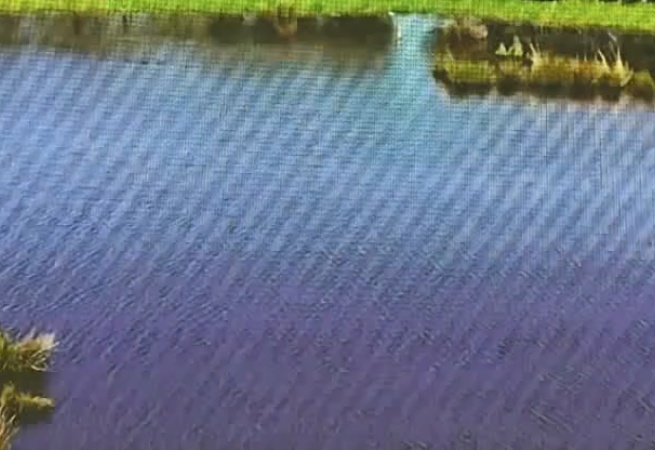} \hspace{-4mm} &
\includegraphics[width=0.149\textwidth]{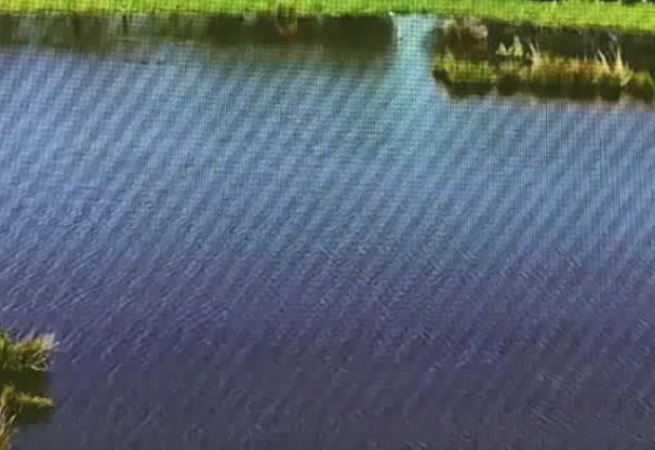} \hspace{-4mm} &
\includegraphics[width=0.149\textwidth]{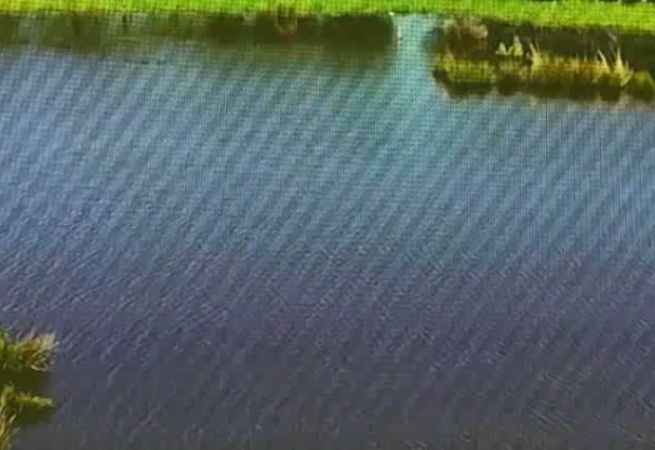} \hspace{-4mm} &
\includegraphics[width=0.149\textwidth]{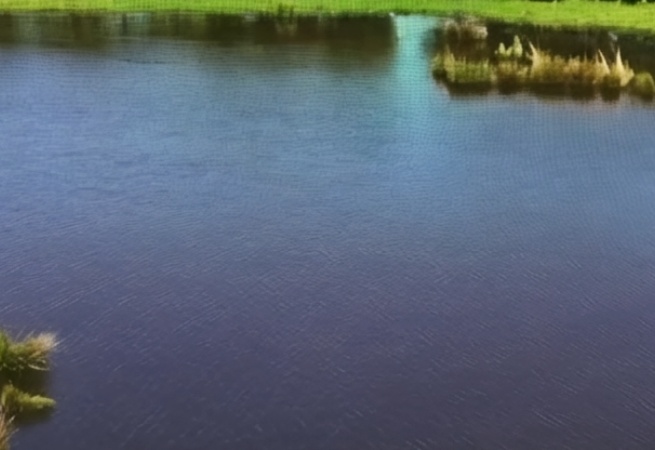} \hspace{-4mm}  
\\ 
FPANet~\cite{oh2025fpanet} \hspace{-4mm} &
CompEvent~\cite{zhong2025compevent} \hspace{-4mm} &
Flickerformer*~\cite{flickerformer} \hspace{-4mm} &
Flickerformer~\cite{flickerformer}  \hspace{-4mm} &
VDFP (ours) \hspace{-4mm}
\\
\end{tabular}
\end{adjustbox}
\end{tabular}
\vspace{-2mm}
\caption{Visual comparison between VDFP and other methods on our real-world dataset DeViD. Specifically, Flickerformer* represents using the official checkpoint to test.}
\label{fig:visual_comparison}
\vspace{-8mm}
\end{figure*}

\vspace{-1mm}
\subsection{Main Results}
\vspace{-2mm}
\textbf{Quantitative Results}. Quantitative results of the comparison between other methods and our model (VDFP) are shown in Tab.~\ref{quantitative result}. It's obvious that VDFP outperforms every model in every metric. Following VDFP, DLoRAL~\cite{DLoRAL} performs secondly well with nearly all of metrics second well. Then FPANet~\cite{oh2025fpanet} and Flickerformer~\cite{flickerformer} follows with only one metric second well. Though VDFP performs well visually, metrics may only improve a little because flicker-banding artifacts are local only on some parts of videos rather than affecting the overall pixels across the frames like video denoising. More explanations of our results are in the Appendix~\ref{section:explanation}.

\vspace{-1mm}
\textbf{Visual Comparison}. Visual comparison is in Fig.~\ref{fig:visual_comparison}. Obviously, our model eliminate most banding and in most videos, both thick and background banding are removed nearly completely. Then DLoRAL follows and in some videos, it can even nearly achieve the removal effect of VDFP. However, even with similar removal visual effect, DLoRAL blurs frames or remains heavy banding when seeing as a whole video rather than a series of frames. Other methods perform not well in most of videos. More visual comparisons are in the Appendix~\ref{section:visual_comparison}.

\vspace{-8pt}
\subsection{Ablation Study}
\begin{figure*}[t!]
    \centering
    \footnotesize
    \begin{subfigure}[b]{0.19\linewidth}
        \includegraphics[width=\linewidth]{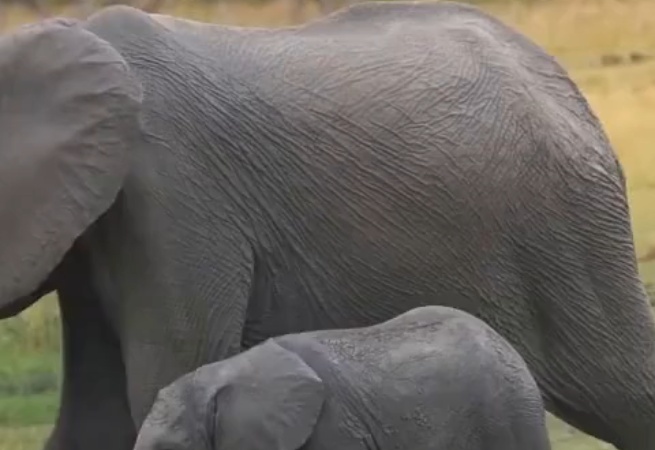}
    \end{subfigure}\hfill
    \begin{subfigure}[b]{0.19\linewidth}
        \includegraphics[width=\linewidth]{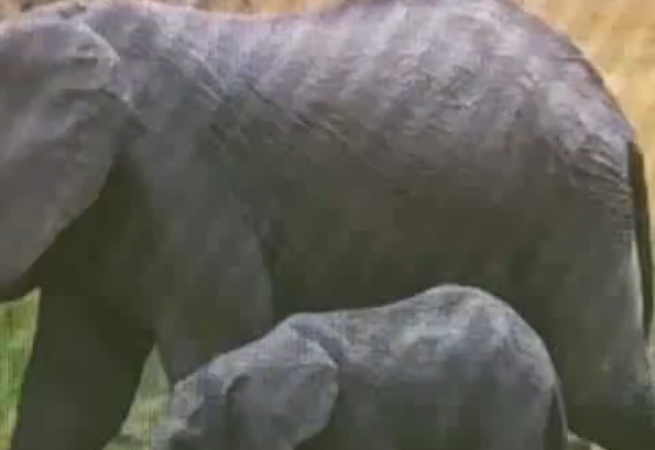}
    \end{subfigure}\hfill
    \begin{subfigure}[b]{0.19\linewidth}
        \includegraphics[width=\linewidth]{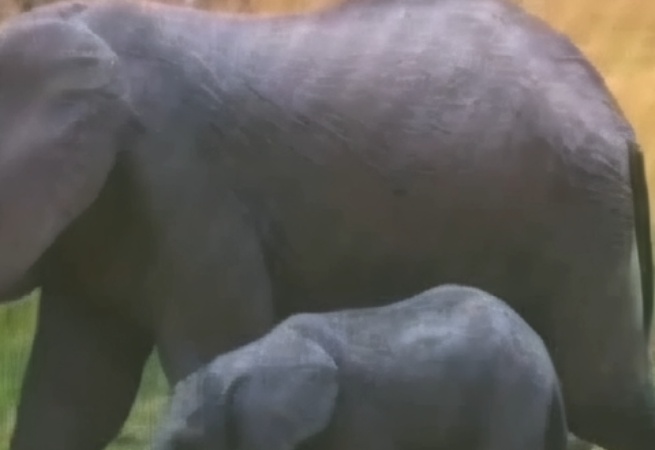}
    \end{subfigure}\hfill
    \begin{subfigure}[b]{0.19\linewidth}
        \includegraphics[width=\linewidth]{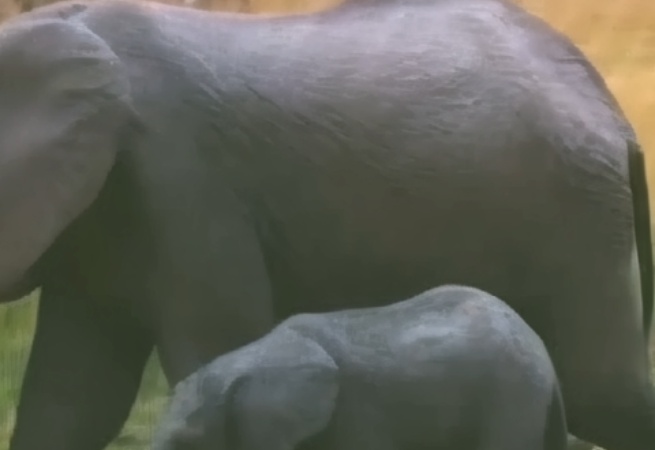}
    \end{subfigure}\hfill
    \begin{subfigure}[b]{0.19\linewidth}
        \includegraphics[width=\linewidth]{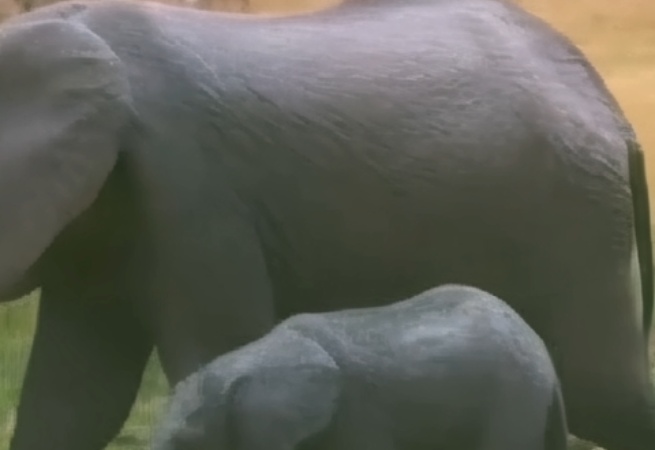}
    \end{subfigure}
    
    \vspace{2pt} 
    \begin{subfigure}[b]{0.19\linewidth}
        \includegraphics[width=\linewidth]{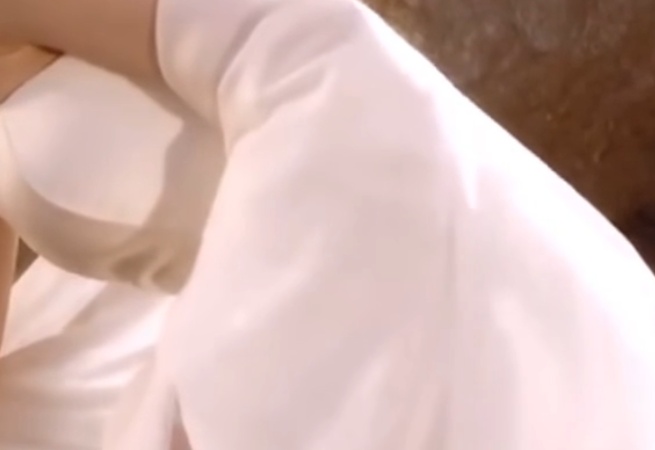}
    \end{subfigure}\hfill
    \begin{subfigure}[b]{0.19\linewidth}
        \includegraphics[width=\linewidth]{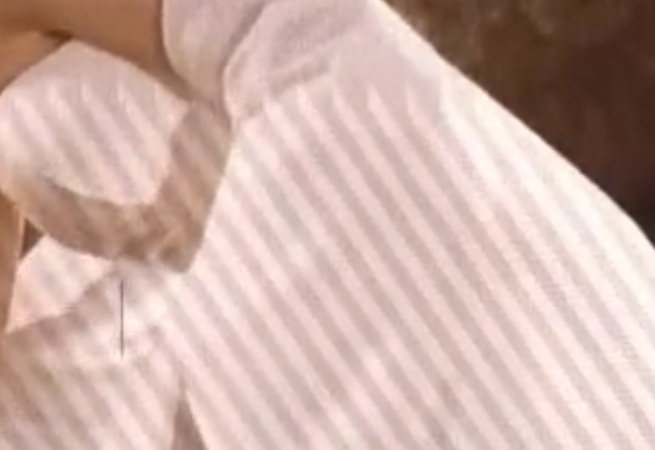}
    \end{subfigure}\hfill
    \begin{subfigure}[b]{0.19\linewidth}
        \includegraphics[width=\linewidth]{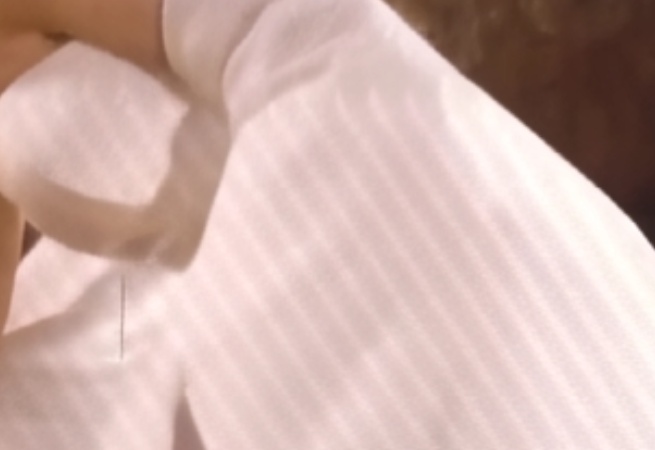}
    \end{subfigure}\hfill
    \begin{subfigure}[b]{0.19\linewidth}
        \includegraphics[width=\linewidth]{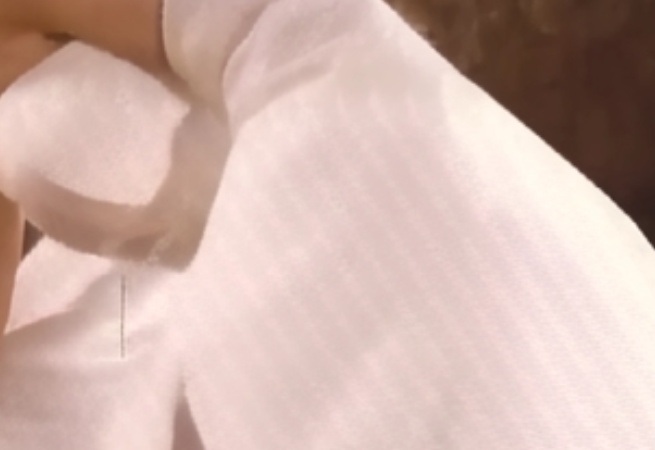}
    \end{subfigure}\hfill
    \begin{subfigure}[b]{0.19\linewidth}
        \includegraphics[width=\linewidth]{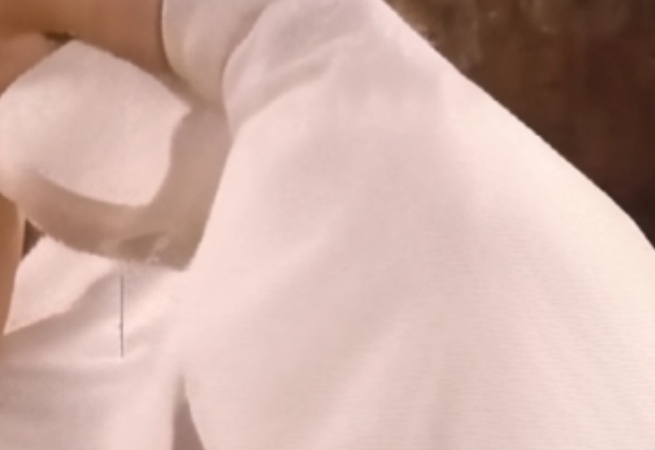}
    \end{subfigure}
    
    \vspace{2pt} 
    \begin{subfigure}[b]{0.19\linewidth}
        \includegraphics[width=\linewidth]{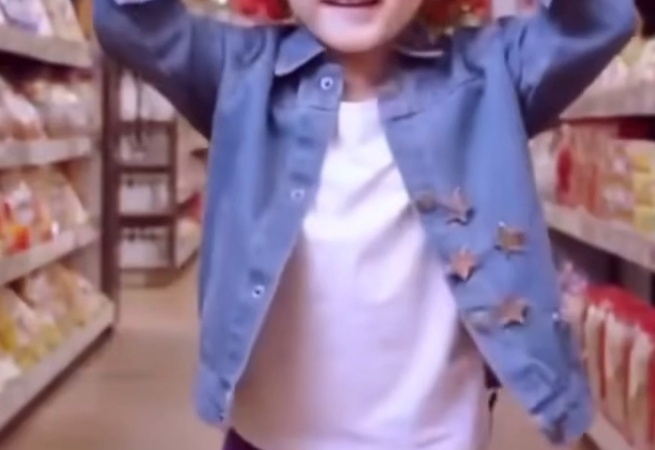}
    \end{subfigure}\hfill
    \begin{subfigure}[b]{0.19\linewidth}
        \includegraphics[width=\linewidth]{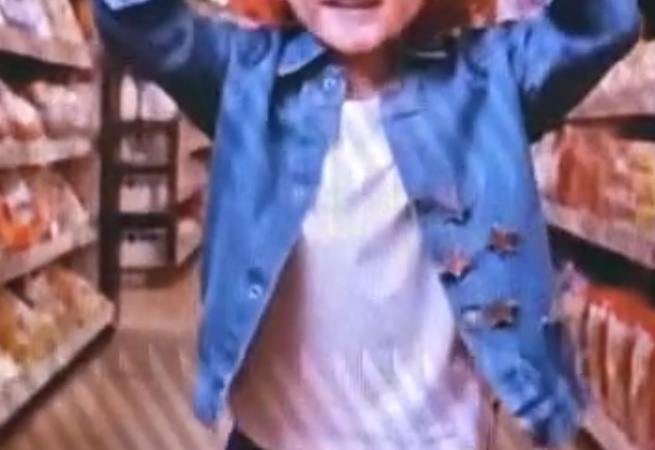}
    \end{subfigure}\hfill
    \begin{subfigure}[b]{0.19\linewidth}
        \includegraphics[width=\linewidth]{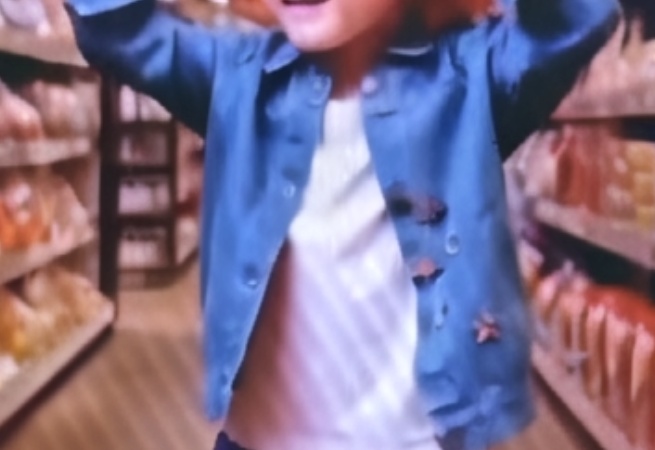}
    \end{subfigure}\hfill
    \begin{subfigure}[b]{0.19\linewidth}
        \includegraphics[width=\linewidth]{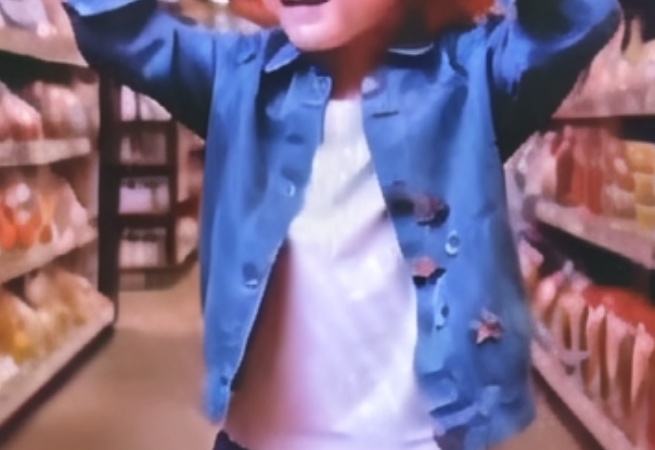}
    \end{subfigure}\hfill
    \begin{subfigure}[b]{0.19\linewidth}
        \includegraphics[width=\linewidth]{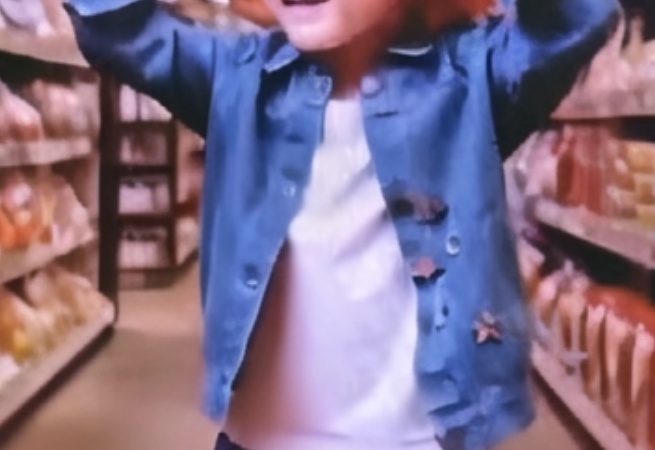}
    \end{subfigure}
    \makebox[0.19\linewidth][c]{GT}\hfill
    \makebox[0.19\linewidth][c]{LQ}\hfill
    \makebox[0.19\linewidth][c]{Baseline}\hfill
    \makebox[0.19\linewidth][c]{DFM}\hfill
    \makebox[0.19\linewidth][c]{DFM+CPP}
    \vspace{-6pt}
    \caption{Visual comparisons of ablation study. The figures show three sets of examples, and from left to right are: GT, LQ, Baseline, DFM, and DFM+CPP.}
    \label{fig:ablation_visual}
    \vspace{-5mm}
\end{figure*}

\vspace{-4pt}
Ablation study is taken in each stage: LQ, Baseline, DFM, and DFM+CPP. Visual comparisons and metrics of our ablation study in each step are shown respectively in Fig.~\ref{fig:ablation_visual} and Tab.~\ref{ablation}. It can be concluded from the table and visual comparisons that every module added improves deflickering performance. More visual comparisons of ablation study are in the Appendix~\ref{abl}.

\noindent \textbf{Effectiveness of the DFM Stage}. As shown in Tab.~\ref{ablation}, integrating the DFM module yields a significant performance boost over the baseline, not only visually but also in the metrics. The baseline stage has most metrics better than LQ videos with VMAF~\cite{VMAF} worse than LQ, which means that flicker-banding in our baseline, STAR~\cite{star}, is removed only a little. This indicates that the DFM design effectively makes flicker-banding artifacts fade greatly compared with banding becoming only a little weaker in the results of the baseline by simulating the real-world degradation field of banding.

\noindent \textbf{Contribution of the CPP Stage}. The addition of the CPP stage further improves the results and works synergistically with the DFM stage, achieving the highest overall quantitative metrics in Tab.~\ref{ablation} and the best elimination effect visually in Fig.~\ref{fig:ablation_visual}. It can be observed from the figures that the CPP module successfully completely removes the residual artifacts remained by the DFM stage.

\noindent \textbf{Detail Preservation and Structural Integrity}. Beyond simply eliminating the flicker-banding artifacts, the complete pipeline ensures that the original video content is not ruined as shown in Fig.~\ref{fig:ablation_visual}. The combined DFM+CPP architecture successfully restores smooth color transitions while strictly preserving the underlying sharp details and structural integrity of the frames.

\begin{table}[t!]
\centering
\footnotesize
\caption{Comparison of metric results of ablation study with different modules: the baseline, DFM, and DFM+CPP. Best results are marked with \textcolor{red}{red}. Overall, every step has improvement.}
\setlength{\tabcolsep}{4pt}

\begin{tabular}{cccccc}
\toprule
\rowcolor{color3} Stage & LPIPS$\downarrow$ & CLIP-IQA$\uparrow$ & BRISQUE$\downarrow$ & VMAF$\uparrow$ & DISTS$\downarrow$ \\ 
\midrule
LQ & 0.3815 & 0.2703 & 42.9104 & 19.2861 & 0.1807 \\
Baseline & 0.3102 & 0.2728 & 38.0108 & 18.1278 & 0.1468 \\
DFM & 0.3050 & 0.2790 & 37.7922 & 19.3047 & 0.1437 \\
\midrule
\rowcolor{violet!20}
DFM+CPP & \textcolor{red}{0.2958} & \textcolor{red}{0.2799} & \textcolor{red}{36.4535} & \textcolor{red}{20.0290} & \textcolor{red}{0.1378} \\
\bottomrule
\end{tabular}
\label{ablation}
\vspace{-14pt}
\end{table}
\vspace{-3mm}
\section{Conclusion}
\vspace{-3mm}
In this paper, we propose VDFP (\textbf{V}ideo \textbf{D}eflickering with \textbf{F}licker-banding \textbf{P}riors) to tackle video deflickering, a prevalent yet under-explored degradation in real-world scenarios. To address the lack of data available on the Internet, we construct DeViD, the real-world dataset and introduce a Degradation Field Modeling (DFM) strategy to synthesize complex multi-banding conditions. Crucially, our architecture features a spatial-temporal Continuous Prior Perception (CPP) module, specifically designed to track banding. Extensive experiments demonstrate that while existing video restoration models struggle with this task, VDFP achieves outstanding performance. Ultimately, our work establishes a highly effective framework for real-world video deflickering. Considering societal impacts, positively, our work can help people restore videos and save precious memory.

\bibliographystyle{plain}
\bibliography{reference}

\begin{thebibliography}{10}

\bibitem{LDM}
Andreas Blattmann, Robin Rombach, Huan Ling, Tim Dockhorn, Seung~Wook Kim, Sanja Fidler, and Karsten Kreis.
\newblock Align your latents: High-resolution video synthesis with latent diffusion models.
\newblock In {\em CVPR}, 2023.

\bibitem{chen2025dove}
Zheng Chen, Zichen Zou, Kewei Zhang, Xiongfei Su, Xin Yuan, Yong Guo, and Yulun Zhang.
\newblock Dove: Efficient one-step diffusion model for real-world video super-resolution.
\newblock In {\em NeurIPS}, 2025.

\bibitem{STBN}
Zikang Chen, Tao Jiang, Xiaowan Hu, Wang Zhang, Huaqiu Li, and Haoqian Wang.
\newblock Spatiotemporal blind-spot network with calibrated flow alignment for self-supervised video denoising.
\newblock In {\em AAAI}, 2025.

\bibitem{2022ECCVdemoire}
Peng Dai, Xin Yu, Lan Ma, Baoheng Zhang, Jia Li, Wenbo Li, Jiajun Shen, and Xiaojuan Qi.
\newblock Video demoireing with relation-based temporal consistency.
\newblock In {\em CVPR}, 2022.

\bibitem{DISTS}
Keyan Ding, Kede Ma, Shiqi Wang, and Eero~P Simoncelli.
\newblock Image quality assessment: Unifying structure and texture similarity.
\newblock {\em TPAMI}, 2022.

\bibitem{transformer2}
Ming Ding, Wendi Zheng, Wenyi Hong, and Jie Tang.
\newblock Cogview2: faster and better text-to-image generation via hierarchical transformers.
\newblock In {\em NeurIPS}, 2022.

\bibitem{DuDemoire2025}
Xuan Dong, Xiangyuan Sun, Xia Wang, Jian Song, Ya~Li, and Weixin Li.
\newblock Video demoireing using focused-defocused dual-camera system.
\newblock {\em TPAMI}, 2025.

\bibitem{durini2019high}
Daniel Durini.
\newblock {\em High performance silicon imaging: Fundamentals and applications of CMOS and CCD sensors}.
\newblock Woodhead Publishing, 2019.

\bibitem{transformer}
Patrick Esser, Robin Rombach, and Bj{\"o}rn Ommer.
\newblock Taming transformers for high-resolution image synthesis.
\newblock In {\em CVPR}, 2021.

\bibitem{geffroy2006organic}
Bernard Geffroy, Philippe Le~Roy, and Christophe Prat.
\newblock Organic light-emitting diode (oled) technology: materials, devices and display technologies.
\newblock {\em Polymer international}, 2006.

\bibitem{GANs}
Ian~J. Goodfellow, Jean Pouget-Abadie, Mehdi Mirza, Bing Xu, David Warde-Farley, Sherjil Ozair, Aaron Courville, and Yoshua Bengio.
\newblock Generative adversarial nets.
\newblock In {\em NeurIPS}, 2014.

\bibitem{digital_demoire}
Bin He, Ce~Wang, Boxin Shi, and Ling-Yu Duan.
\newblock Fhde2net: Full high definition demoireing network.
\newblock In {\em ECCV}, 2020.

\bibitem{VDM}
Jonathan Ho, Tim Salimans, Alexey Gritsenko, William Chan, Mohammad Norouzi, and David~J. Fleet.
\newblock Video diffusion models.
\newblock In {\em NeurIPS}, 2022.

\bibitem{GANs_3}
Tero Karras, Miika Aittala, Samuli Laine, Erik H\"{a}rk\"{o}nen, Janne Hellsten, Jaakko Lehtinen, and Timo Aila.
\newblock Alias-free generative adversarial networks.
\newblock In {\em NeurIPS}, 2021.

\bibitem{GANs_4}
Tero Karras, Samuli Laine, and Timo Aila.
\newblock A style-based generator architecture for generative adversarial networks.
\newblock {\em TPAMI}, 2021.

\bibitem{MUSIQ}
Junjie Ke, Qifei Wang, Yilin Wang, Peyman Milanfar, and Feng Yang.
\newblock Musiq: Multi-scale image quality transformer.
\newblock In {\em ICCV}, 2021.

\bibitem{GANs_2}
Moez Krichen.
\newblock Generative adversarial networks.
\newblock In {\em ICCCNT}, 2023.

\bibitem{Lai-ECCV-2018}
Wei-Sheng Lai, Jia-Bin Huang, Oliver Wang, Eli Shechtman, Ersin Yumer, and Ming-Hsuan Yang.
\newblock Learning blind video temporal consistency.
\newblock In {\em ECCV}, 2018.

\bibitem{VMAF}
Zhi Li, Christos Bampis, Julie Novak, Anne Aaron, Kyle Swanson, Anush Moorthy, and JD~Cock.
\newblock Vmaf: The journey continues.
\newblock {\em Netflix Technology Blog}, 25(1), 2018.

\bibitem{Sora}
Yixin Liu, Kai Zhang, Yuan Li, Zhiling Yan, Chujie Gao, Ruoxi Chen, Zhengqing Yuan, Yue Huang, Hanchi Sun, Jianfeng Gao, Lifang He, and Lichao Sun.
\newblock Sora: A review on background, technology, limitations, and opportunities of large vision models.
\newblock {\em arXiv preprint arXiv:2402.17177}, 2024.

\bibitem{AdamW}
Ilya Loshchilov and Frank Hutter.
\newblock Decoupled weight decay regularization.
\newblock In {\em ICLR}, 2017.

\bibitem{LCDdisplay}
Robert~A. Meyers, editor.
\newblock {\em Encyclopedia of Physical Science and Technology}.
\newblock Academic Press, third edition, 2001.

\bibitem{BRISQUE}
Anish Mittal, Anush~Krishna Moorthy, and Alan~Conrad Bovik.
\newblock No-reference image quality assessment in the spatial domain.
\newblock {\em TIP}, 2012.

\bibitem{Unet}
B.~Murugesakumar, Aakash S, A~Arunmanikandan, M~Chithra, and M~Dhasaranjan.
\newblock U-net convolutional networks for real-time biomedical image segmentation and anomaly detection.
\newblock In {\em ICONSTEM}, 2025.

\bibitem{oh2025fpanet}
Gyeongrok Oh, Sungjune Kim, Heon Gu, Sang~Ho Yoon, Jinkyu Kim, and Sangpil Kim.
\newblock Fpanet: Frequency-based video demoireing using frame-level post alignment.
\newblock In {\em Neural Networks}, 2025.

\bibitem{BurstDeflicker_lishenqu}
Lishen Qu, Zhihao Liu, Shihao Zhou, Yaqi Luo, Jie Liang, Hui Zeng, Lei Zhang, and Jufeng Yang.
\newblock Burstdeflicker: A benchmark dataset for flicker removal in dynamic scenes.
\newblock In {\em NeurIPS}, 2025.

\bibitem{flickerformer}
Lishen Qu, Shihao Zhou, Jie Liang, Hui Zeng, Lei Zhang, and Jufeng Yang.
\newblock It takes two: A duet of periodicity and directionality for burst flicker removal.
\newblock In {\em CVPR}, 2026.

\bibitem{transformer3}
Aditya Ramesh, Mikhail Pavlov, Gabriel Goh, Scott Gray, Chelsea Voss, Alec Radford, Mark Chen, and Ilya Sutskever.
\newblock Zero-shot text-to-image generation.
\newblock In {\em ICML}, 2021.

\bibitem{DLoRAL}
Yujing Sun, Lingchen Sun, Shuaizheng Liu, Rongyuan Wu, Zhengqiang Zhang, and Lei Zhang.
\newblock One-step diffusion for detail-rich and temporally consistent video super-resolution.
\newblock In {\em NeurlPS}, 2025.

\bibitem{sung2025mocha}
Jeahun Sung, Changhyun Roh, Chanho Eom, and Jihyong Oh.
\newblock Mocha-former: Moir{\'e}-conditioned hybrid adaptive transformer for video demoir{\'e}ing.
\newblock {\em Neurocomputing}, 2025.

\bibitem{tao2017spmc}
Xin Tao, Hongyun Gao, Renjie Liao, Jue Wang, and Jiaya Jia.
\newblock Detail-revealing deep video super-resolution.
\newblock In {\em ICCV}, 2017.

\bibitem{CLIP-IQA}
Jianyi Wang, Kelvin~CK Chan, and Chen~Change Loy.
\newblock Exploring clip for assessing the look and feel of images.
\newblock In {\em AAAI}, 2023.

\bibitem{wang2004image}
Zhou Wang, Alan~C Bovik, Hamid~R Sheikh, and Eero~P Simoncelli.
\newblock Image quality assessment: from error visibility to structural similarity.
\newblock {\em TIP}, 2004.

\bibitem{wu2023dover}
Haoning Wu, Erli Zhang, Liang Liao, Chaofeng Chen, Jingwen~Hou Hou, Annan Wang, Wenxiu~Sun Sun, Qiong Yan, and Weisi Lin.
\newblock Exploring video quality assessment on user generated contents from aesthetic and technical perspectives.
\newblock In {\em ICCV}, 2023.

\bibitem{star}
Rui Xie, Yinhong Liu, Penghao Zhou, Chen Zhao, Jun Zhou, Kai Zhang, Zhenyu Zhang, Jian Yang, Zhenheng Yang, and Ying Tai.
\newblock Star: Spatial-temporal augmentation with text-to-video models for real-world video super-resolution.
\newblock In {\em ICCV}, 2025.

\bibitem{Survey_On_Diffusion}
Zhen Xing, Qijun Feng, Haoran Chen, Qi~Dai, Han Hu, Hang Xu, Zuxuan Wu, and Yu-Gang Jiang.
\newblock A survey on video diffusion models.
\newblock {\em arXiv preprint arXiv:2310.10647}, 2024.

\bibitem{xu2025alignmentfreerawvideodemoireing}
Shuning Xu, Xina Liu, Binbin Song, Xiangyu Chen, Qiubo Chen, and Jiantao Zhou.
\newblock Alignment-free raw video demoireing.
\newblock {\em arXiv preprint arXiv:2408.10679}, 2025.

\bibitem{DTNet}
Shuning Xu, Binbin Song, Xiangyu Chen, and Jiantao Zhou.
\newblock Direction-aware video demoireing with temporal-guided bilateral learning.
\newblock {\em arXiv preprint arXiv:2308.13388}, 2023.

\bibitem{YCb}
Qirui Yang, Fangpu Zhang, Yeying Jin, Qihua Cheng, Pengtao Jiang, Huanjing Yue, and Jingyu Yang.
\newblock Dsdnet: Raw domain demoir\'{e}ing via dual color-space synergy.
\newblock In {\em ACM MM}, 2025.

\bibitem{PFNL}
Peng Yi, Zhongyuan Wang, Kui Jiang, Junjun Jiang, and Jiayi Ma.
\newblock Progressive fusion video super-resolution network via exploiting non-local spatio-temporal correlations.
\newblock In {\em ICCV}, 2019.

\bibitem{UHDM}
Xin Yu, Peng Dai, Wenbo Li, Lan Ma, Jiajun Shen, Jia Li, and Xiaojuan Qi.
\newblock Towards efficient and scale-robust ultra-high-definition image demoir{\'e}ing.
\newblock In {\em ECCV}, 2022.

\bibitem{yue2023recaptured}
Huanjing Yue, Yijia Cheng, Xin Liu, and Jingyu Yang.
\newblock Recaptured raw screen image and video demoireing via channel and spatial modulations.
\newblock {\em arXiv preprint arXiv:2310.20332}, 2023.

\bibitem{swinunet}
Chengxi Zeng, Xinyu Yang, David Smithard, Majid Mirmehdi, Alberto~M Gambaruto, and Tilo Burghardt.
\newblock Video-swinunet: Spatio-temporal deep learning framework for vfss instance segmentation.
\newblock {\em arXiv preprint arXiv:2302.11325}, 2023.

\bibitem{zhang2018perceptual}
Richard Zhang, Phillip Isola, Alexei~A Efros, Eli Shechtman, and Oliver Wang.
\newblock The unreasonable effectiveness of deep features as a perceptual metric.
\newblock In {\em CVPR}, 2018.

\bibitem{zhong2025compevent}
Mingchen Zhong, Xin Lu, Dong Li, Senyan Xu, Ruixuan Jiang, Xueyang Fu, and Baocai Yin.
\newblock Compevent: Complex-valued event-rgb fusion for low-light video enhancement and deblurring.
\newblock In {\em AAAI}, 2026.

\bibitem{zhou2024upscaleavideo}
Shangchen Zhou, Peiqing Yang, Jianyi Wang, Yihang Luo, and Chen~Change Loy.
\newblock {Upscale-A-Video}: Temporal-consistent diffusion model for real-world video super-resolution.
\newblock In {\em CVPR}, 2024.

\bibitem{RIFLE}
Libo Zhu, Zihan Zhou, Xiaoyang Liu, Weihang Zhang, Keyu Shi, Yifan Fu, and Yulun Zhang.
\newblock Rifle: Removal of image flicker-banding via latent diffusion enhancement.
\newblock {\em arXiv preprint arXiv:2509.24644}, 2025.

\end{thebibliography}

\newpage
\appendix

\section{Technical Appendices and Supplementary Materials}
\subsection{Mechanism of Flicker-banding Artifact Formation}
\label{FB}
Flicker-banding is fundamentally a visual manifestation of temporal aliasing, emerging as alternating luminance striations, such as curve stripes, cracked banding or diamond-like artifacts, across frames. The intensity and structural clarity of these banding artifacts are highly sensitive to camera exposure settings. Specifically, a narrower exposure window typically makes the flicker-banding artifacts more serious. Ultimately, this degradation is driven by a severe asynchronous interaction between the recording device's sampling mechanism and the emitting source's modulation strategy~\cite{RIFLE}.

On the acquisition front, the majority of the modern smartphone cameras employ the Complementary Metal-Oxide Semiconductor (CMOS) sensors operating under an electronic rolling shutter paradigm~\cite{durini2019high}. Unlike global shutters that capture an entire scene instantaneously, the rolling shutters integrate light progressively on a row-by-row basis. This sequential readout mechanism inherently maps different temporal instances of the scene to discrete spatial rows on the sensor pane, introducing a continuous time offset across the vertical axis of the frame.

From illumination perspective, active electronic displays do not output a constant flux of light. Instead, they rely on high-frequency temporal modulation to render images and control perceived brightness. For instance, Liquid Crystal Displays (LCD) dynamically alter the polarization state of backlight illumination through voltage-controlled crystal alignments~\cite{LCDdisplay}. Similarly, Organic Light-Emitting Diode (OLED) screens commonly employ Pulse-Width Modulation (PWM) to regulate pixel luminance by adjusting duty cycle of light pulses~\cite{geffroy2006organic}. Furthermore, LED matrices typically utilize multiplexed scanning, sequentially driving distinct rows or columns at high frequencies.

Consequently, both the photon accumulation process of the camera and the photon emission process of the display screens are discrete and time-varying events rather than continuous. When the progressive and row-wise sampling rate of the rolling shutter falls out of synchronization with the periodic modulation cycle of the display, the sensor inadvertently captures varying phases of the screen's illumination waveform. It is this fundamental spatiotemporal misalignment that is permanently recorded as the periodic light and dark flicker-banding artifacts in the final output.
\begin{figure}[t!]
    \centering
    \includegraphics[width=\textwidth]{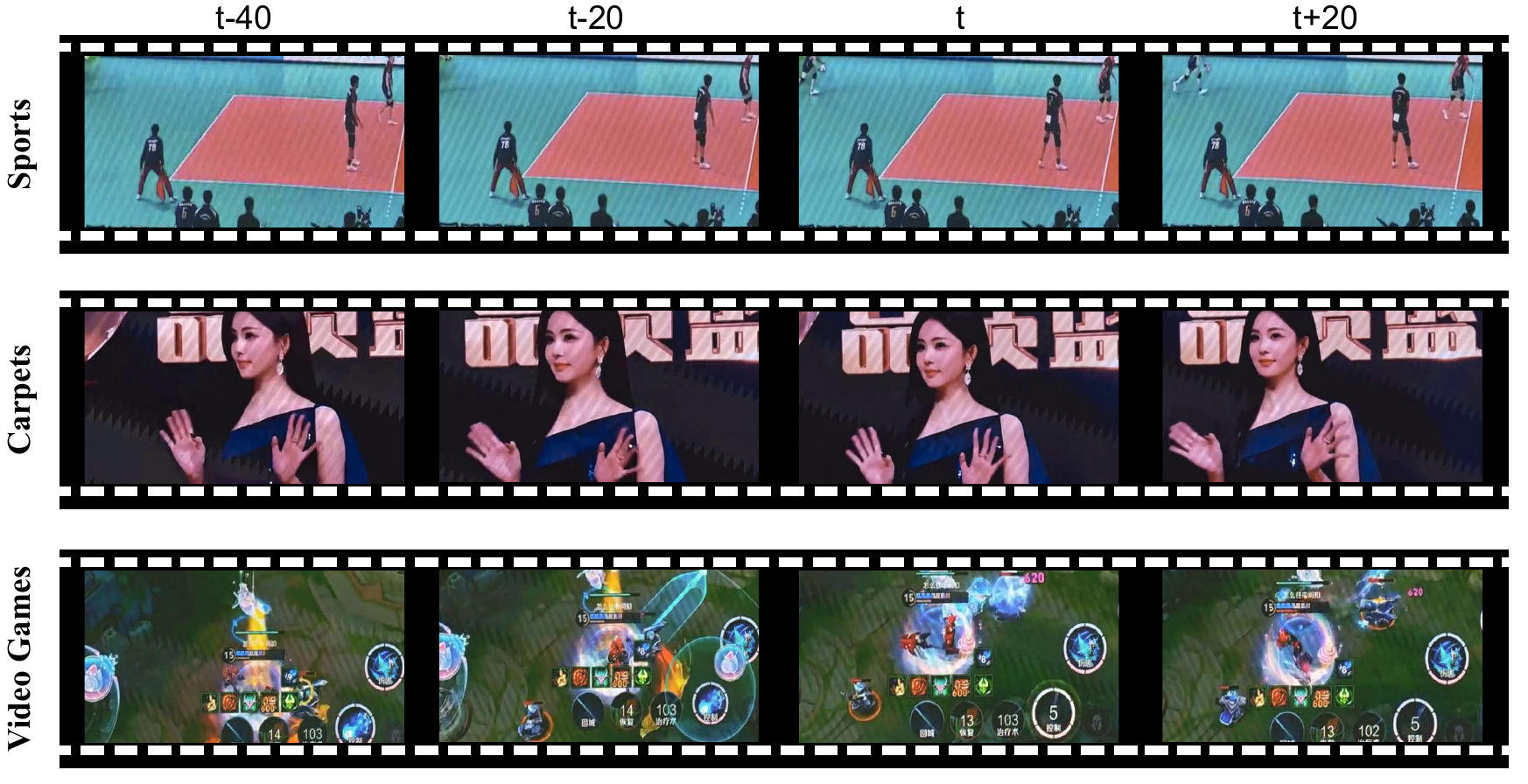}  

    \vspace{1mm}
    \includegraphics[width=\textwidth]{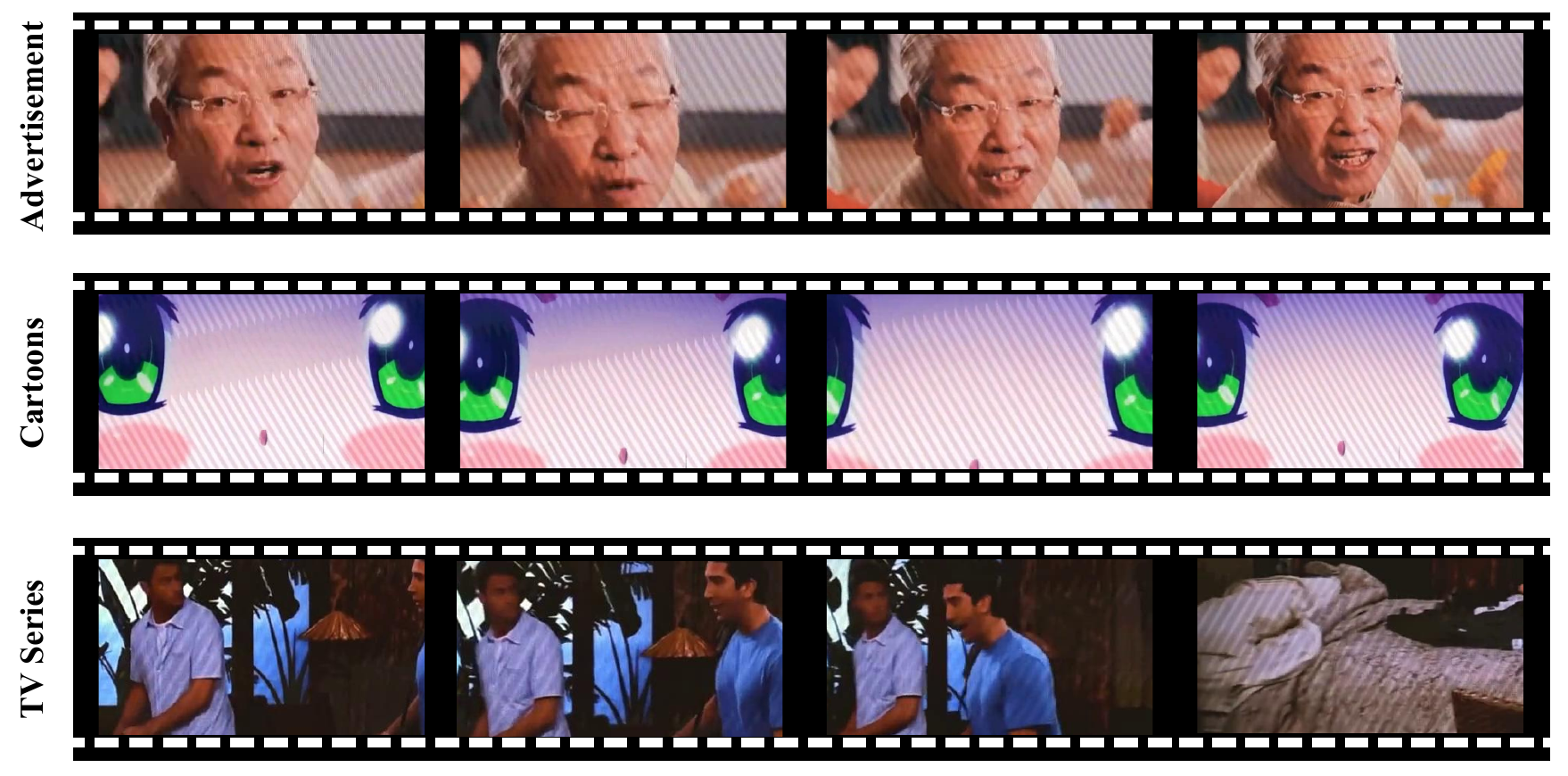}   
    \caption{More examples of our real-world dataset (DeViD) in various scenes, such as sports, carpets, video games, streets, cartoons, and TV series.}
    \label{fig:supp_data}
\end{figure}

\subsection{Visual Examples of Our Real-world Dataset (DeViD)}
\label{section:DeViD}
More sets of visual examples of our real-world dataset DeViD are shown in Fig.~\ref{fig:supp_data}, which are shown in the form of four continuous frames. These figures contain scenes in sports, carpets, video games, cartoons, and TV series, which are parts of scenes contained in DeViD.
\subsection{Degradation Field Modeling}
\label{section:DFM}
In this section, we detail the spatial generation functions for the five types of fundamental specific banding proposed in our dual-layer degradation field modeling: Linear Uniform Stripe, Curve Stripe, Cracked Stripe, Diamond Stripe, and Complex Stripe. Since the thick banding is also based on our background banding, we only introduce the formulas for background banding here.

For a given frame at time $t$, we map the spatial coordinates to a localized orthogonal coordinate system $(u, v)$, where $u$ represents the axis parallel to the stripe direction, and $v(t)$ is the orthogonal axis including the kinematic temporal shift as defined in the main text. The periodicity of the stripe is defined as $T = W + G$, where $W$ is the base stripe width and $G$ is the gap width. The stripe index for any pixel is determined by $k = \text{round}(v(t) / T)$, and center coordinate of the $k$-th stripe is $v_c = k \cdot T$.
\subsubsection{Linear Uniform Stripe}
The linear uniform stripe serves as the ideal baseline, representing standard light flickering without severe structural degradation. The stripe maintains a constant width $W$ along the $u$-axis, and its edges are perfectly straight. The physical boundary distance is calculated as:
\begin{equation}
    d_{uniform}(u, v) = |v(t) - v_c| - \frac{W}{2}.
\end{equation}
The occupancy banding is then generated exclusively by the smoothstep function to simulate standard optical defocus and $f$ is the edge feathering threshold.:
\begin{equation}
    O_{uniform}(u, v) = 1.0 - S\left(-f, +f, d_{uniform}(u,v)\right).
\end{equation}
\subsubsection{Curve Stripe}
Rolling shutter effects combined with non-linear camera lens distortion often yield curved banding. We introduce a quadratic spatial mapping $C(u)$ along the orthogonal axis:
\begin{equation}
    C(u) = \pm A \cdot \left( \frac{u - u_{mid}}{u_{span}} \right)^2 - \mu_C,
\end{equation}where $u_{mid}$ is the spatial center of the $u$-axis, $u_{span}$ is the half-span of the screen, $A$ is the curve amplitude, and $\mu_C$ centers the deformation. To generate the curved mask, this deformation $C(u)$ is directly subtracted within the physical boundary distance formula defined in the main text:
\begin{equation}
d_{curve}(u, v, t) = |v(t) - C(u) - v_c| - \frac{W}{2}.
\end{equation}
The occupancy $O_{curve}$ is then generated by feeding $d_{curve}$ into the smoothstep function $S(\cdot)$.
\subsubsection{Cracked Stripe}
To simulate extreme signal interference resulting in split or fractured bands, the Cracked Stripe model generates randomized longitudinal cracks. We define a preserved central width $v_{keep} = R_{keep} \cdot (W / 2)$, where $R_{keep}$ is the random ratio of the uncracked banding occupying the whole banding, forming the backbone $O_{main} = \mathbb{I}[|v(t) - v_c| \le v_{keep}]$.
For the $k$-th stripe, $n$ random crack centers $O_i$ are sampled in the remaining margin regions. The local width of each crack is modulated by a 1D smoothed noise $\omega_i(u) = \omega_{base} \cdot (1 + \rho \cdot N_{1D}(u))$, where $\omega_{base}$ is the base width of the crack, $\rho$ is the jitter ratio and $N_{1D}(u)$ represents a smoothed, band-limited 1D Gaussian noise designed to simulate the continuous structural fluctuation of the crack width, rather than discontinuous high-frequency static. We first generate a discrete standard white noise signal $X(u) \sim \mathcal{N}(0, 1)$ along the parallel axis $u$. To eliminate unrealistic high-frequency discontinuities, $X(u)$ is convolved with a 1D Gaussian kernel $G_\sigma(u)$ parameterized by a smoothing factor $\sigma$:
\begin{equation}
X'(u) = X(u) * G_\sigma(u).
\end{equation}
We apply Z-score normalization to ensure consistent perturbation amplitude across different synthesized stripes. Final structural noise is normalized as:
\begin{equation}
N_{1D}(u) = \frac{X'(u)}{\sqrt{\text{Var}(X')}}.
\end{equation}
The union of sub-masks creates the fracture geometry:
\begin{equation}
O_{sub}(u, v, t) = \bigcup_{i=1}^{n} \mathbb{I}\left[ |v(t) - (v_c \pm O_i)| \le 0.5 \cdot \omega_i(u) \right].
\end{equation}
The final banding is written as $O_{cracked} = \max(O_{main}, O_{sub})$.
\subsubsection{Diamond Stripe}
We utilize local modulo operations along the $u$-axis to make banding appear periodically: $u_{mod} = u \pmod L$, where $L$ is the diamond length.The tangent angle of diamonds is defined as $\tan\alpha = \frac{W (1 - R_w)}{L}$, where $R_w$ is the size of diamonds. The dynamic upper and lower boundaries of the sheared band are modeled as:
\begin{equation}
    v_{top}(u) = v_c + 0.5 \cdot W \cdot R_w - \tan\alpha \cdot \left( u_{mod} - \frac{L}{2} \right).
\end{equation}
\begin{equation}
    v_{bot}(u) = v_c - 0.5 \cdot W \cdot R_w - \tan\alpha \cdot \left( u_{mod} - \frac{L}{2} \right).
\end{equation}
The sheared occupancy $O_{shear} = \mathbb{I}[v_{bot}(u) \le v(t) \le v_{top}(u)]$ is then multiplied by the base stripe mask $O_{base}$. This geometric intersection exactly crops the slanted bands into diamond grids.
\subsubsection{Complex Stripe}
Real-world banding often suffers from irregular sensor readout noise and signal jitter, leading to torn edges and uneven widths. Building upon the uniform model, we introduce parametric structural jitters. The actual width of the $k$-th stripe varies dynamically along the $u$-axis:
\begin{equation}
    W_k(u) = \max(1.0, W + \delta_{w,k} + \omega_{wiggle}(u)),
\end{equation}
where $\delta_{w,k} \sim \mathcal{U}(-j_w, j_w)$ is the base width jitter per stripe, and $\omega_{wiggle}(u)$ is a smoothed 1D noise array simulating low-frequency width fluctuation.The effective upper and lower boundaries ($v_{top}$ and $v_{bot}$) are further perturbed by independent high-frequency edge noises $N_{top}(u)$ and $N_{bot}(u)$:
\begin{equation}
    v_{top}(u) = (v_c + \delta_{s,k}) + 0.5 \cdot W_k(u) + \eta \cdot \gamma_{t} \cdot N_{top}(u).
\end{equation}
\begin{equation}
    v_{bot}(u) = (v_c + \delta_{s,k}) - 0.5 \cdot W_k(u) + \eta \cdot \gamma_{b} \cdot N_{bot}(u).
\end{equation}
Here, $\delta_{s,k} \sim \mathcal{U}(-j_s, j_s)$ is the spacing jitter, $\eta$ is the maximum perpendicular jitter amplitude, and $\gamma_{t,b} \in [0.6, 1.0]$ are per-stripe random scaling factors. The initial signed distance $d_0$ is calculated as $d_0(u,v) = -\min(v_{top}(u) - v(t), v(t) - v_{bot}(u))$. Finally, a 2D global blur noise field $J(u,v)$ is subtracted to form the final distance $d_{complex}(u,v) = d_0(u,v) - \epsilon \cdot J(u,v)$, which is fed into the smoothstep function $S(\cdot)$ to output the final complex mask.
\subsection{Explanations for Other Comparison Methods' Weakness.}
\label{section:explanation}
This section will in detail describe the results of the comparison methods and explain why their results are not so satisfying. Since the DLoRAL~\cite{DLoRAL} model has been discussed in the main text in detail, the supplementary material here will skip this model.

It is interesting that the Flickerformer~\cite{flickerformer} model deals with a very similar visual effect, burst flickering removal caused by the alternating current light, but it performs badly no matter with the official checkpoint posted on the Internet or even after being trained on datasets processed by our degradation field modeling module. The reason may be that this model tackles with a far easier situation than our model, where the background is still with banding only horizontal and in only one shape. On the contrary, VDFP needs to eliminate two kinds of banding in one video with random directions while different videos may contain different kinds of banding respectively. Moreover, the banding artifacts targeted by Flickerformer are relatively mild and transient, meaning that a specific spatial location corrupted by banding at the current moment may become completely clean in the next frame. In stark contrast, VDFP addresses a significantly more severe degradation scenario where dense banding artifacts persistently cover the entire frame across the temporal dimension. Under these complex conditions, the assumption of finding clean reference pixels in adjacent frames fails entirely, as the corresponding spatial locations in neighboring frames remain heavily contaminated.

CompEvent~\cite{zhong2025compevent} fails to resolve flicker-banding artifacts, resulting in fragmented stripes and severe temporal flickering, because the screen's high-frequency refresh rate floods the event stream with non-kinematic noise that breaks the network's temporal alignment mechanism. Furthermore, its reliance on localized convolutions lacks the global structural priors necessary to capture and completely eliminate continuous banding patterns. FPANet~\cite{oh2025fpanet}'s Frequency Selection Module relies on soft confidence maps that merely reduce the extreme high-frequency energy of display flicker-banding, leaving residual faded stripes instead of completely eliminating them. Consequently, its downstream Post Align Module erroneously tracks the phase motion of these residual stripes rather than the underlying scene, forcefully locking the artifacts into the temporally fused output. STBN~\cite{STBN} relies on a spatiotemporal blind-spot assumption that strictly requires noise to be pixel-independent, which fundamentally fails when applied to the highly structural flicker-banding artifacts, causing the network to arbitrarily break continuous stripes into random high-frequency noise. Furthermore, the strong phase motion of the banding severely misleads its optical flow-based alignment, resulting in erroneous frame warping and temporal fusion, blurring the underlying content in videos.

\begin{figure*}[htbp]

\scriptsize
\centering
\begin{tabular}{ccc}
\hspace{-0.45cm}
\begin{adjustbox}{valign=t}
\begin{tabular}{c}
\includegraphics[width=0.216\textwidth,height=0.231\textwidth]{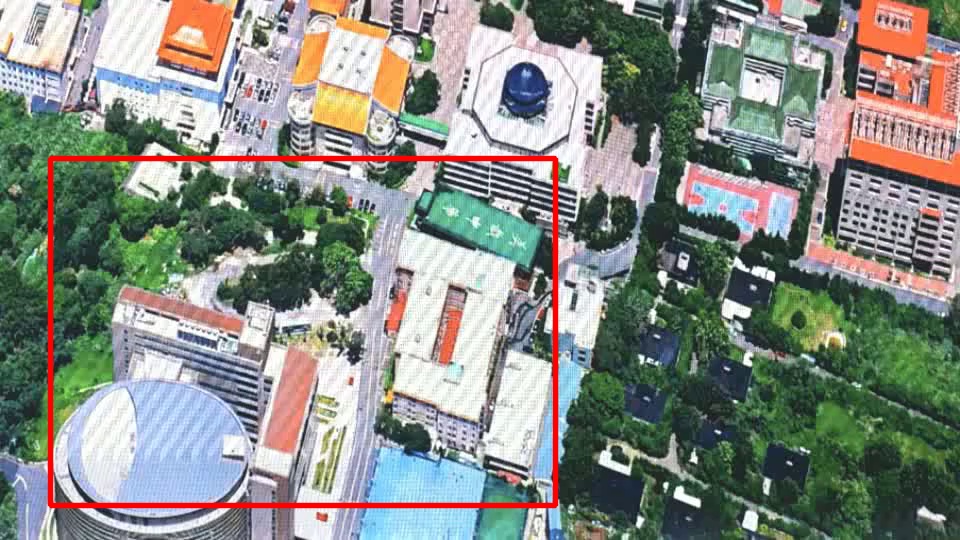}
\\
DeViD: 002
\end{tabular}
\end{adjustbox}
\hspace{-0.46cm}
\begin{adjustbox}{valign=t}
\begin{tabular}{cccccc}
\includegraphics[width=0.149\textwidth]{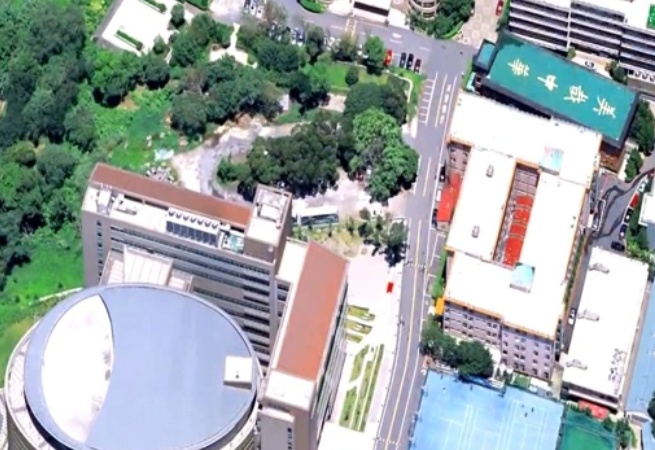} \hspace{-4mm} &
\includegraphics[width=0.149\textwidth]{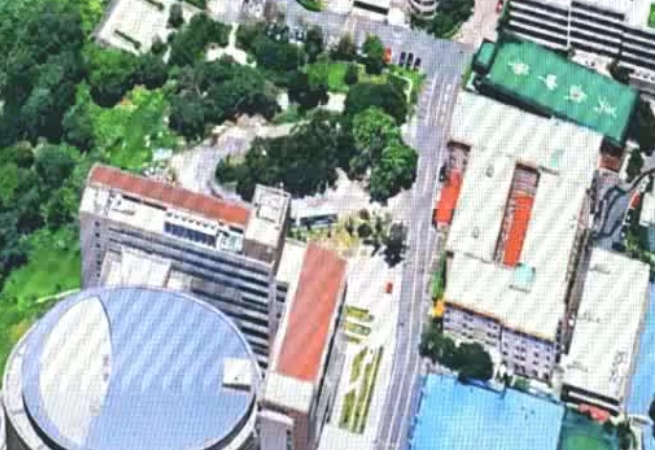} \hspace{-4mm} &
\includegraphics[width=0.149\textwidth]{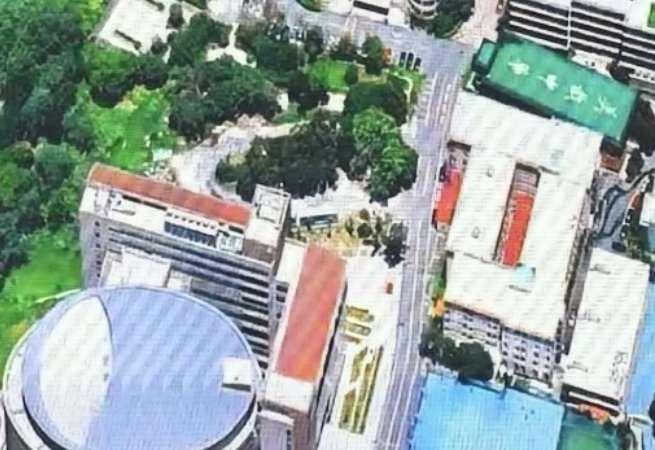} \hspace{-4mm} &
\includegraphics[width=0.149\textwidth]{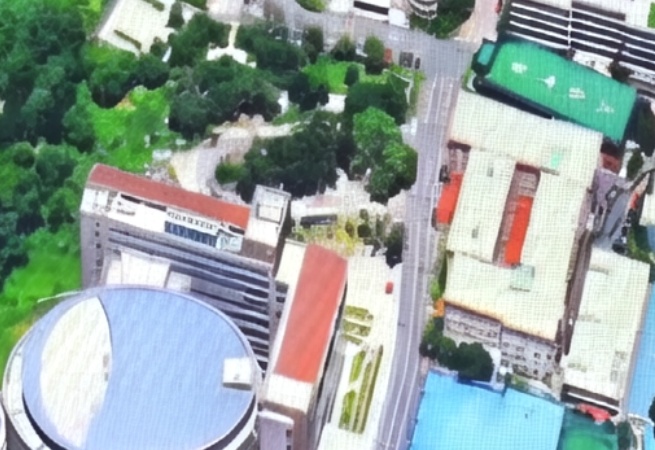} \hspace{-4mm} &
\includegraphics[width=0.149\textwidth]{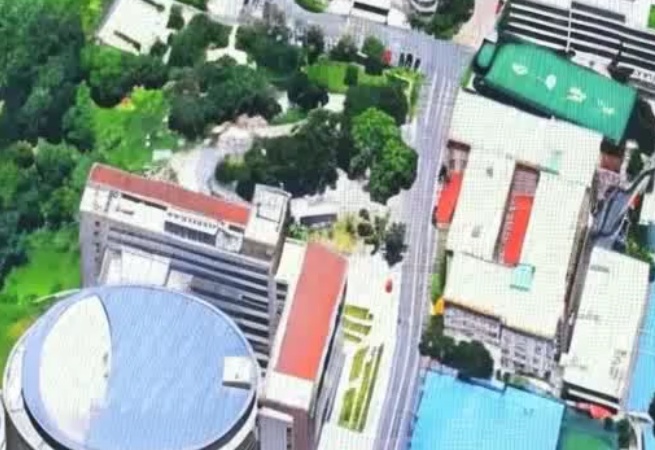} \hspace{-4mm} 
\\
GT \hspace{-4mm} &
LQ \hspace{-4mm} &
STBN~\cite{STBN} \hspace{-4mm} &
STAR~\cite{star} \hspace{-4mm} &
DLoRAL~\cite{DLoRAL} \hspace{-4mm}
\\
\includegraphics[width=0.149\textwidth]{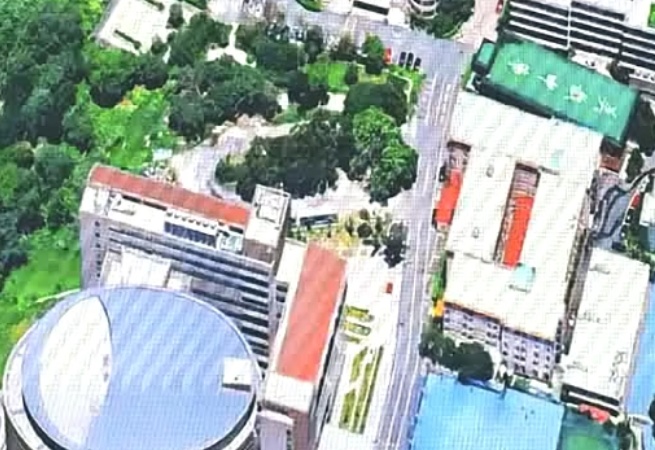} \hspace{-4mm} &
\includegraphics[width=0.149\textwidth]{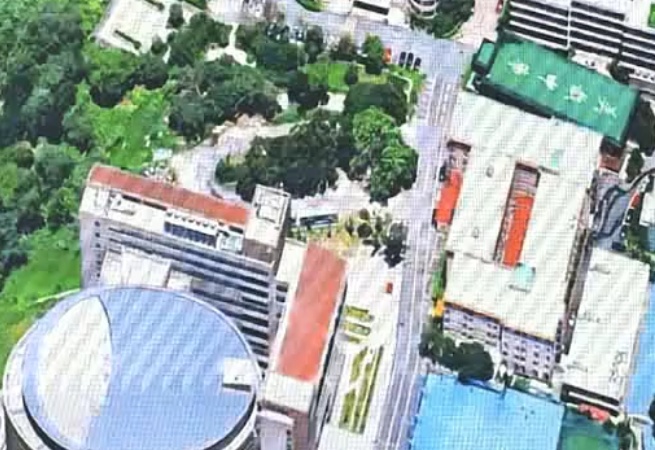} \hspace{-4mm} &
\includegraphics[width=0.149\textwidth]{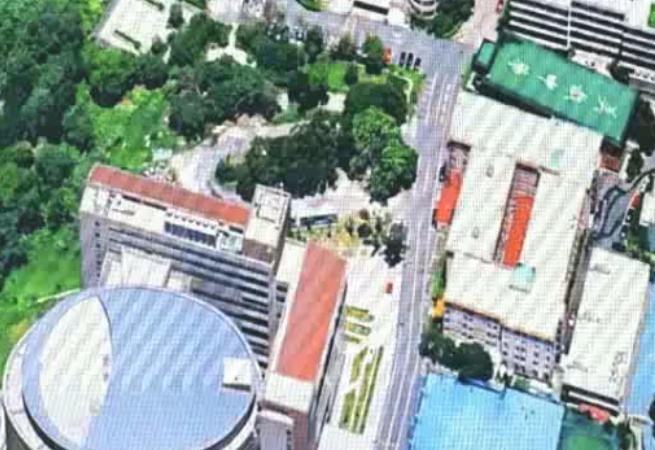} \hspace{-4mm} &
\includegraphics[width=0.149\textwidth]{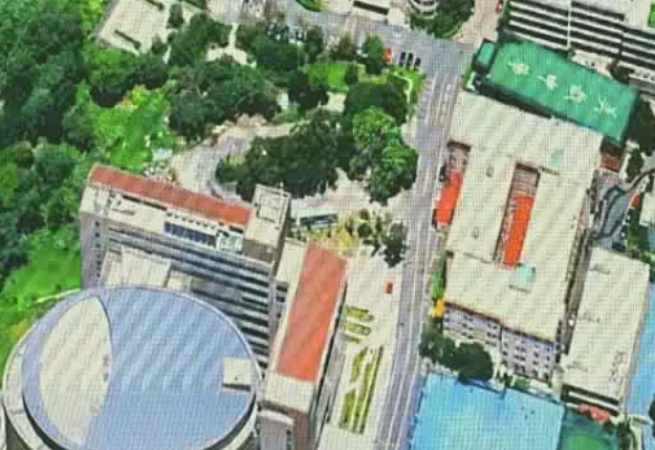} \hspace{-4mm} &
\includegraphics[width=0.149\textwidth]{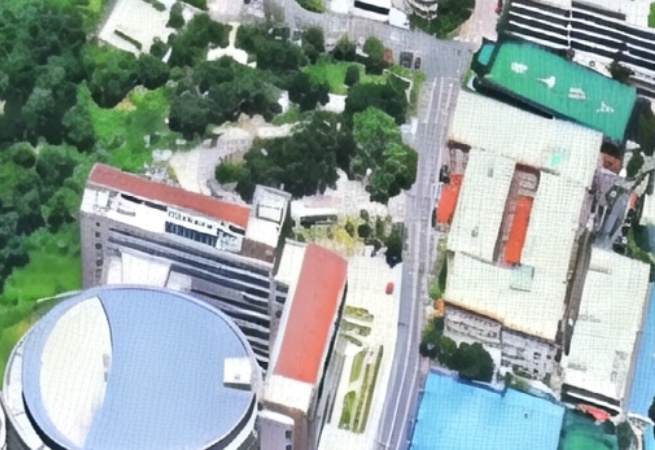} \hspace{-4mm}  
\\ 
FPANet~\cite{oh2025fpanet} \hspace{-4mm} &
CompEvent~\cite{zhong2025compevent} \hspace{-4mm} &
Flickerformer*~\cite{flickerformer} \hspace{-4mm} &
Flickerformer~\cite{flickerformer}  \hspace{-4mm} &
VDFP (ours) \hspace{-4mm}
\\
\end{tabular}
\end{adjustbox}
\\
\hspace{-0.42cm}
\begin{adjustbox}{valign=t}
\begin{tabular}{c}
\includegraphics[width=0.216\textwidth,height=0.231\textwidth]{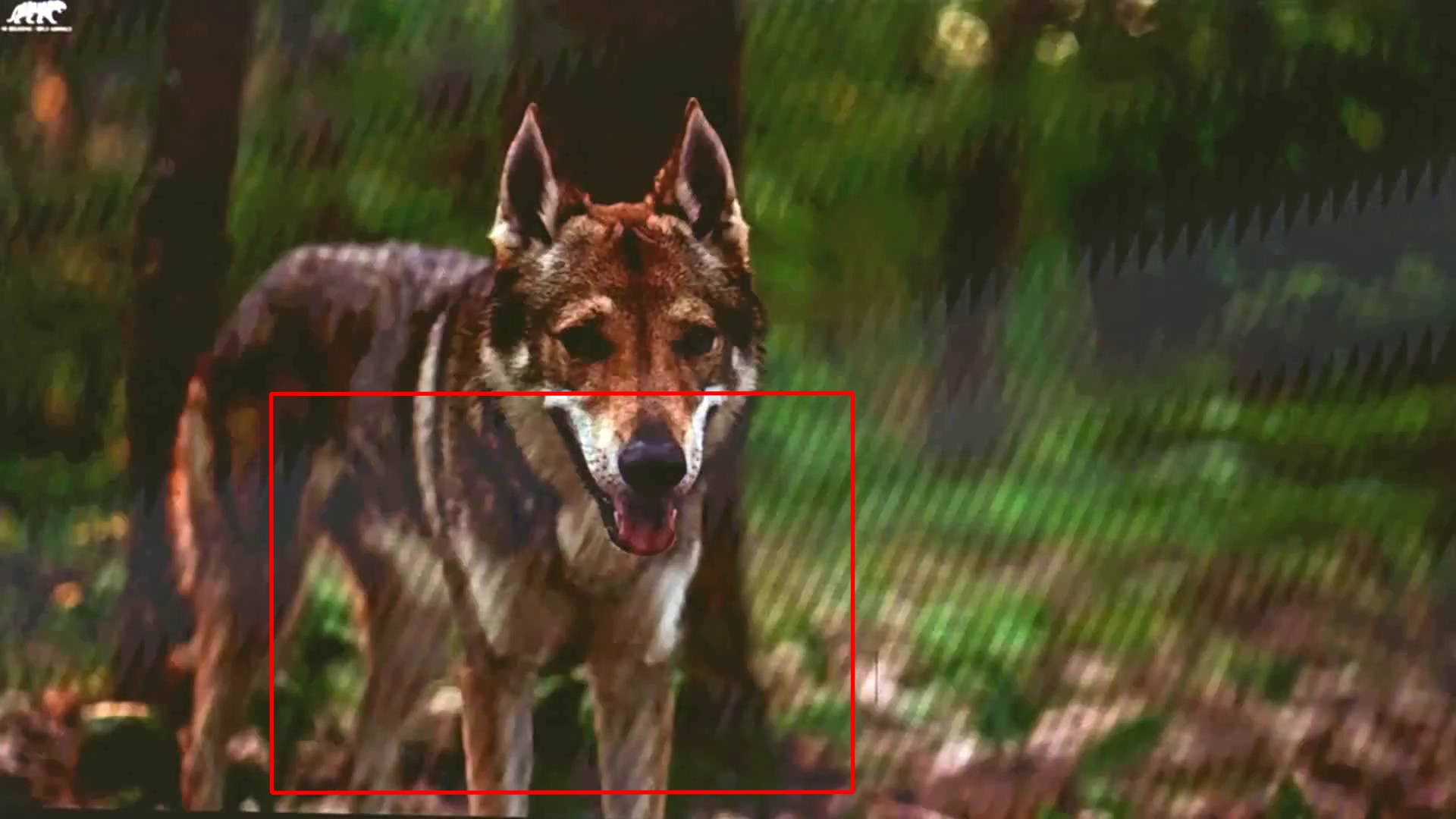}
\\
DeViD: 005
\end{tabular}
\end{adjustbox}
\hspace{-0.46cm}
\begin{adjustbox}{valign=t}
\begin{tabular}{cccccc}
\includegraphics[width=0.149\textwidth]{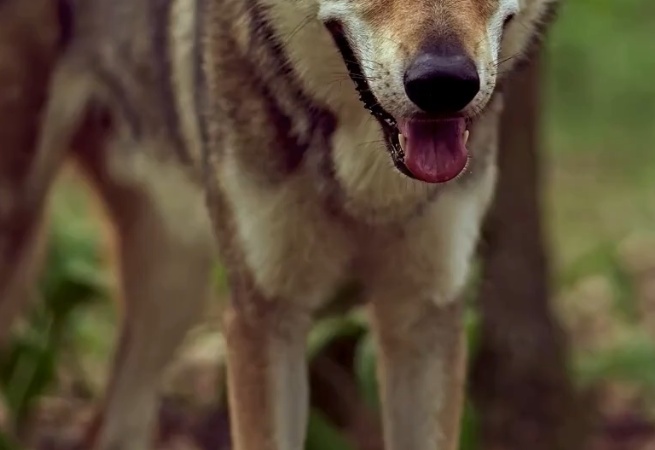} \hspace{-4mm} &
\includegraphics[width=0.149\textwidth]{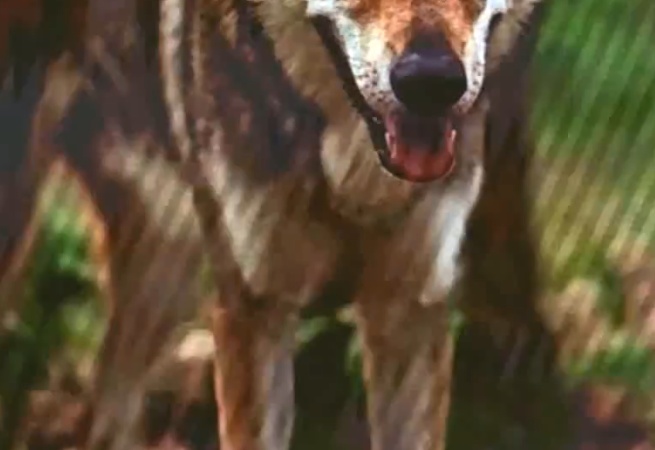} \hspace{-4mm} &
\includegraphics[width=0.149\textwidth]{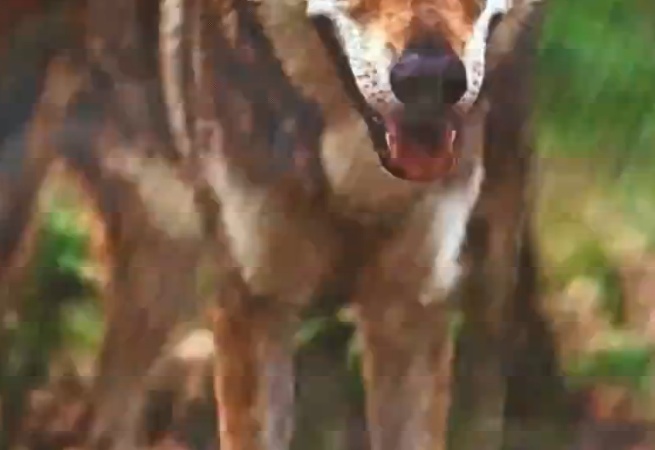} \hspace{-4mm} &
\includegraphics[width=0.149\textwidth]{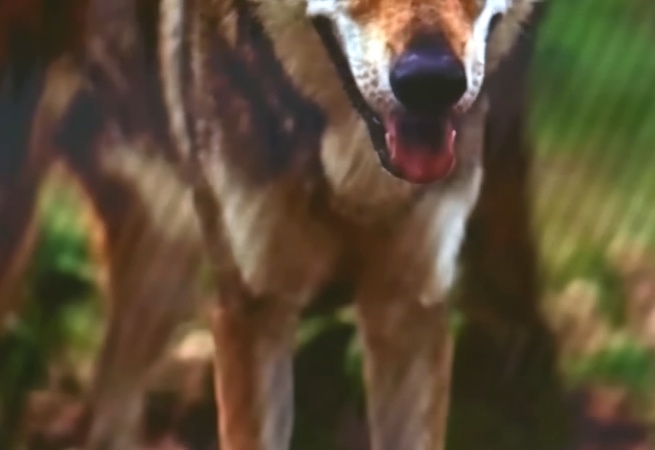} \hspace{-4mm} &
\includegraphics[width=0.149\textwidth]{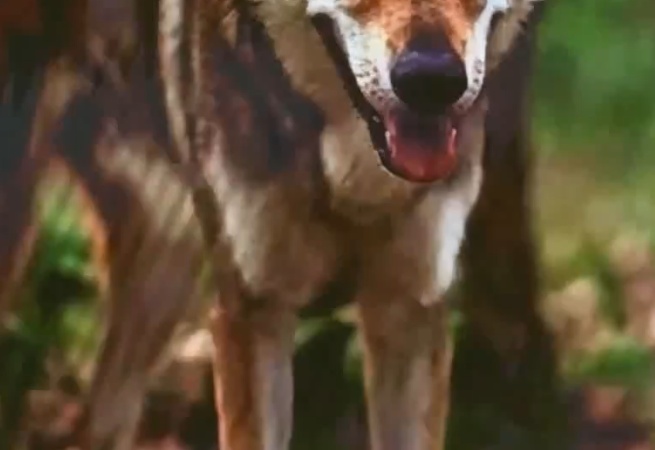} \hspace{-4mm} 
\\
GT \hspace{-4mm} &
LQ \hspace{-4mm} &
STBN~\cite{STBN} \hspace{-4mm} &
STAR~\cite{star} \hspace{-4mm} &
DLoRAL~\cite{DLoRAL} \hspace{-4mm}
\\
\includegraphics[width=0.149\textwidth]{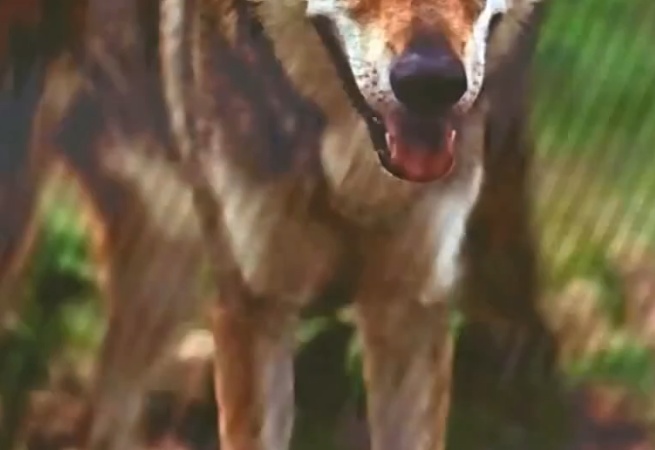} \hspace{-4mm} &
\includegraphics[width=0.149\textwidth]{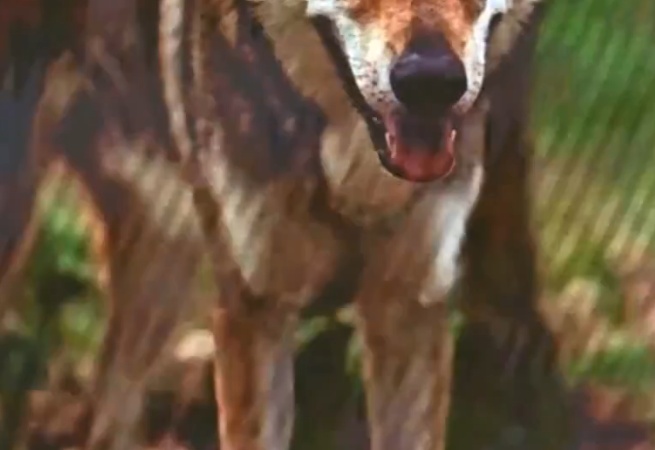} \hspace{-4mm} &
\includegraphics[width=0.149\textwidth]{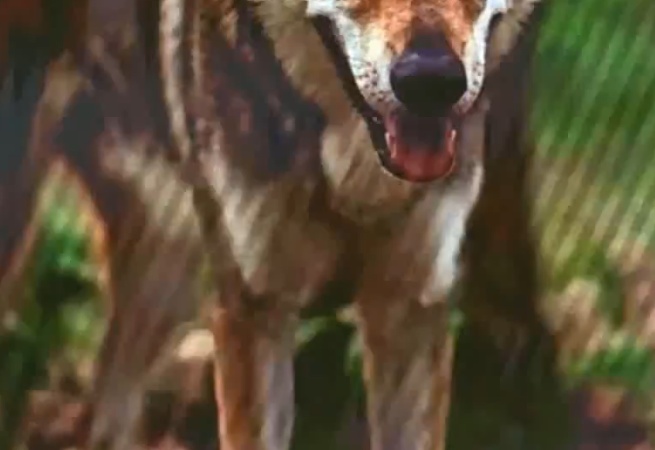} \hspace{-4mm} &
\includegraphics[width=0.149\textwidth]{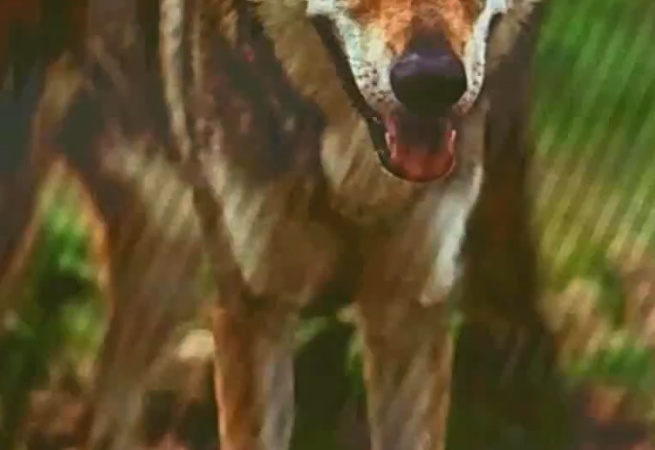} \hspace{-4mm} &
\includegraphics[width=0.149\textwidth]{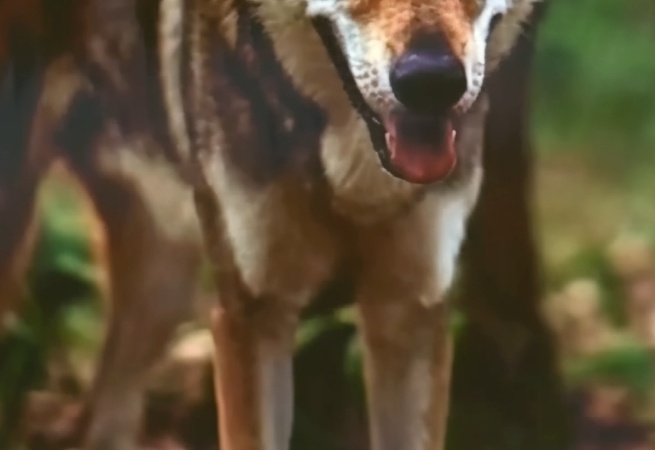} \hspace{-4mm}  
\\ 
FPANet~\cite{oh2025fpanet} \hspace{-4mm} &
CompEvent~\cite{zhong2025compevent} \hspace{-4mm} &
Flickerformer*~\cite{flickerformer} \hspace{-4mm} &
Flickerformer~\cite{flickerformer}  \hspace{-4mm} &
VDFP (ours) \hspace{-4mm}
\\
\end{tabular}
\end{adjustbox}
\\
\hspace{-0.42cm}
\begin{adjustbox}{valign=t}
\begin{tabular}{c}
\includegraphics[width=0.216\textwidth,height=0.231\textwidth]{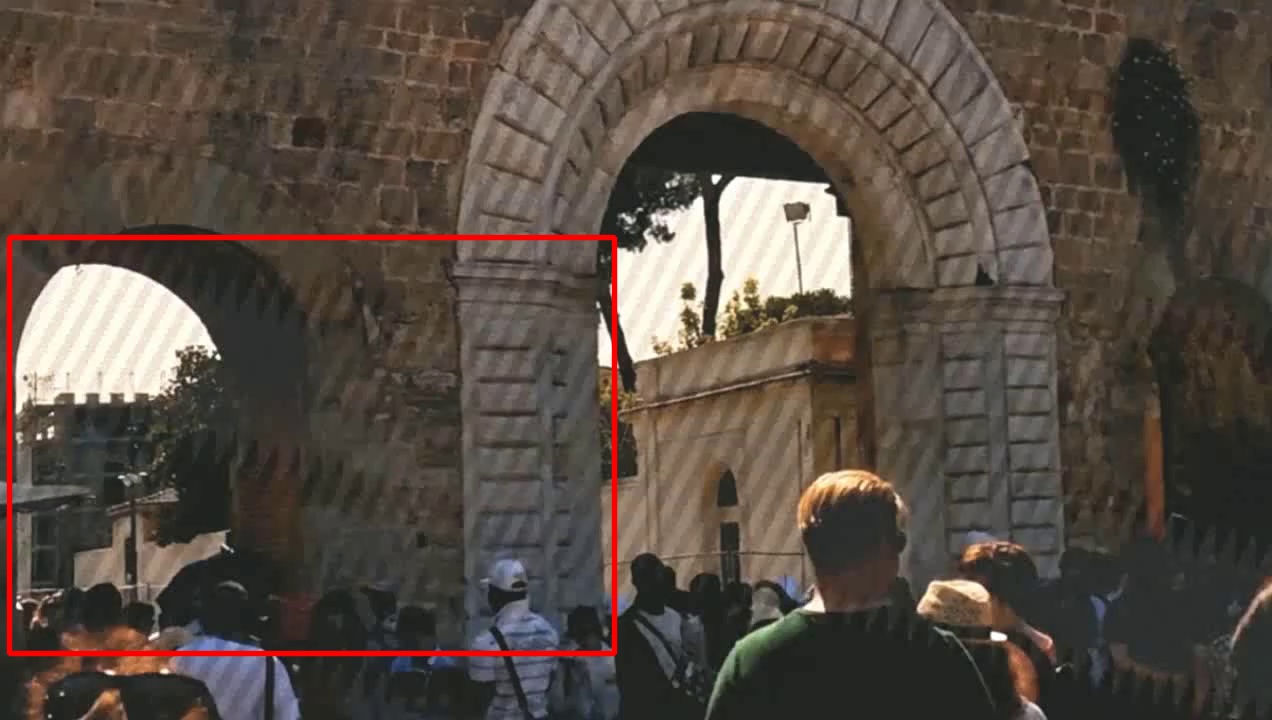}
\\
DeViD: 010
\end{tabular}
\end{adjustbox}
\hspace{-0.46cm}
\begin{adjustbox}{valign=t}
\begin{tabular}{cccccc}
\includegraphics[width=0.149\textwidth]{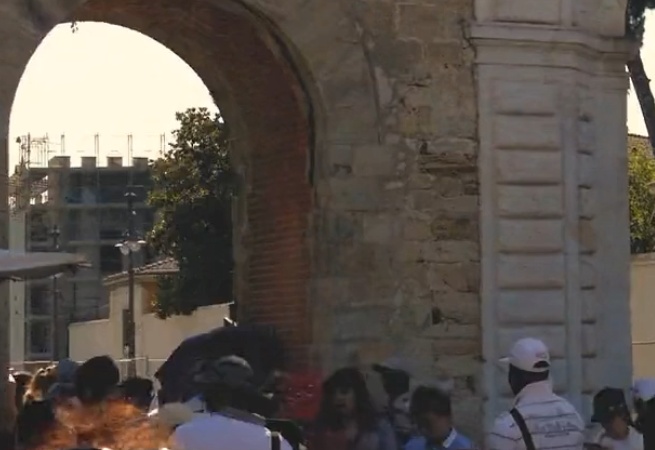} \hspace{-4mm} &
\includegraphics[width=0.149\textwidth]{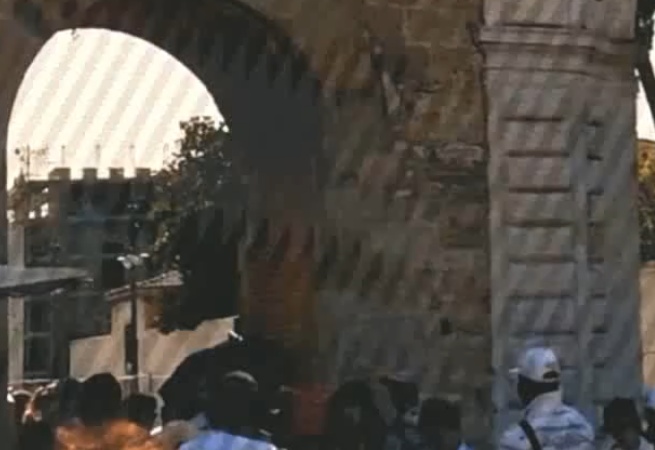} \hspace{-4mm} &
\includegraphics[width=0.149\textwidth]{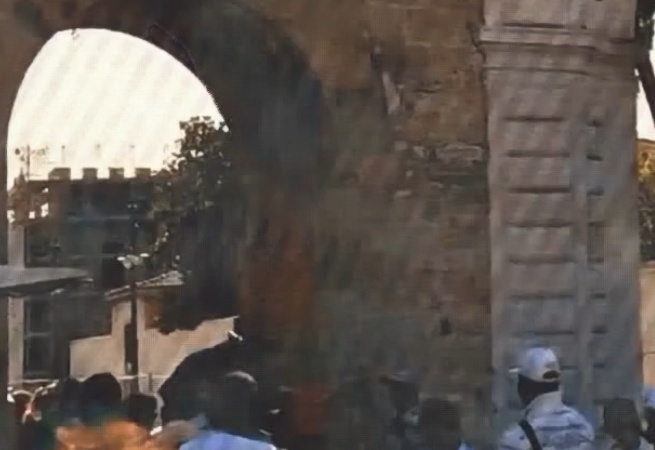} \hspace{-4mm} &
\includegraphics[width=0.149\textwidth]{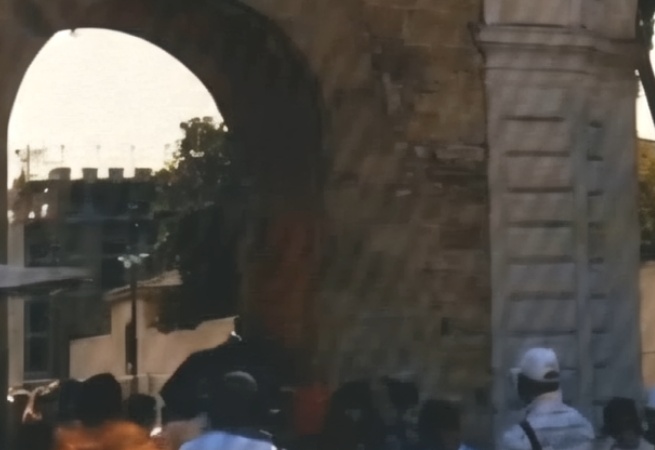} \hspace{-4mm} &
\includegraphics[width=0.149\textwidth]{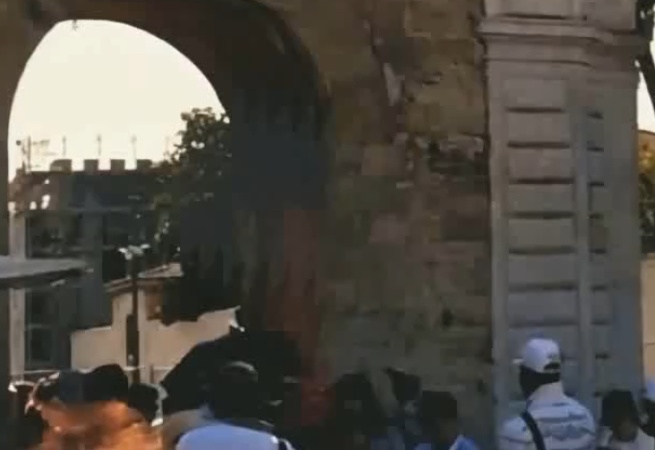} \hspace{-4mm} 
\\
HQ \hspace{-4mm} &
LQ \hspace{-4mm} &
STBN~\cite{STBN} \hspace{-4mm} &
STAR~\cite{star} \hspace{-4mm} &
DLoRAL~\cite{DLoRAL} \hspace{-4mm}
\\
\includegraphics[width=0.149\textwidth]{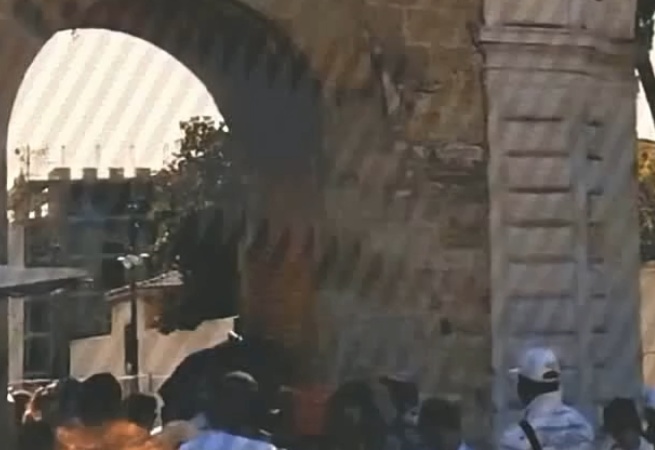} \hspace{-4mm} &
\includegraphics[width=0.149\textwidth]{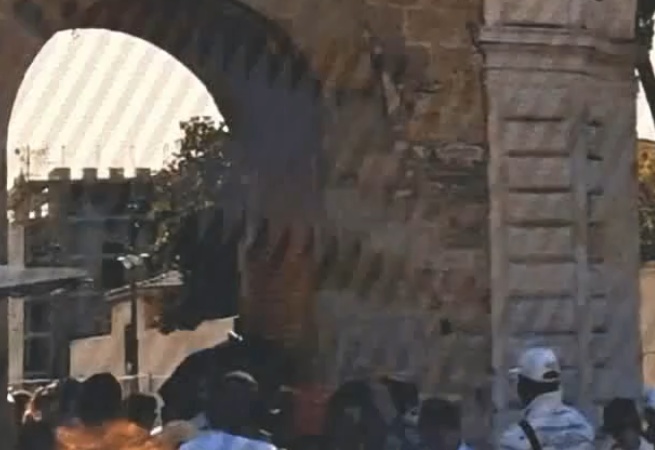} \hspace{-4mm} &
\includegraphics[width=0.149\textwidth]{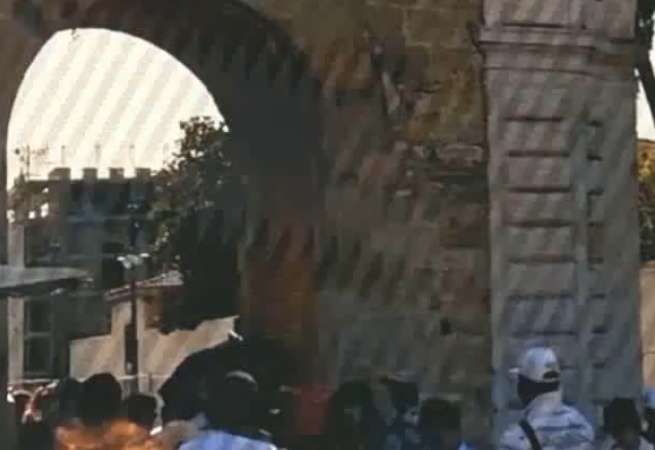} \hspace{-4mm} &
\includegraphics[width=0.149\textwidth]{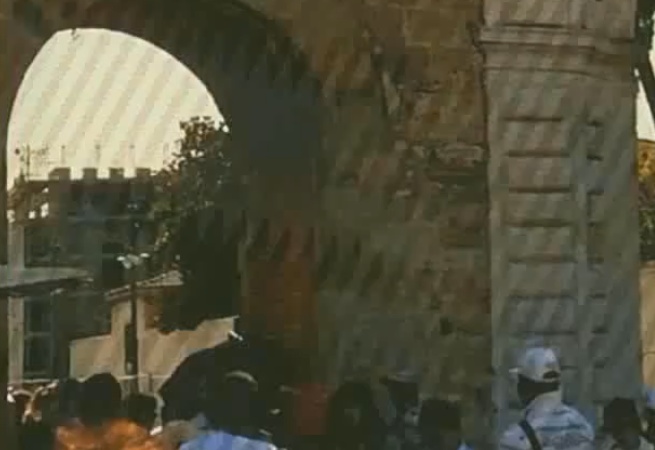} \hspace{-4mm} &
\includegraphics[width=0.149\textwidth]{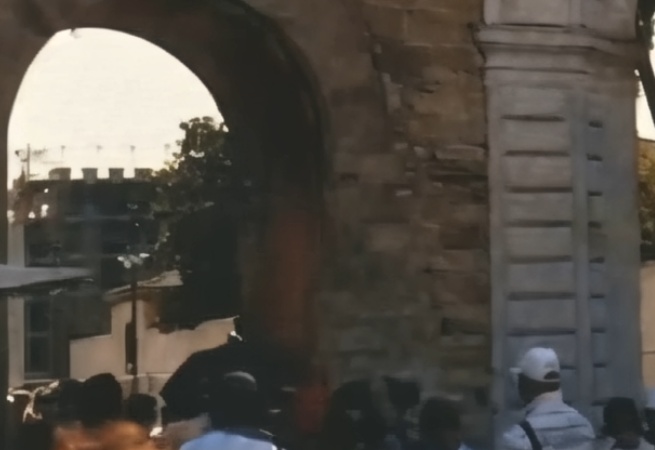} \hspace{-4mm}  
\\ 
FPANet~\cite{oh2025fpanet} \hspace{-4mm} &
CompEvent~\cite{zhong2025compevent} \hspace{-4mm} &
Flickerformer*~\cite{flickerformer} \hspace{-4mm} &
Flickerformer~\cite{flickerformer}  \hspace{-4mm} &
VDFP (ours) \hspace{-4mm}
\\
\end{tabular}
\end{adjustbox}
\\

\hspace{-0.42cm}
\begin{adjustbox}{valign=t}
\begin{tabular}{c}
\includegraphics[width=0.216\textwidth,height=0.231\textwidth]{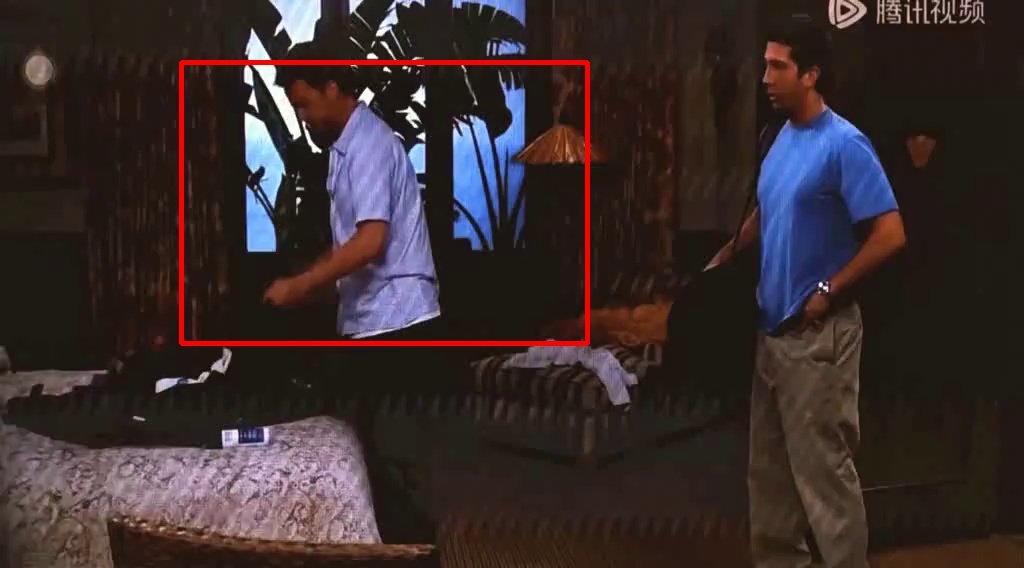}
\\
DeViD: 099
\end{tabular}
\end{adjustbox}
\hspace{-0.46cm}
\begin{adjustbox}{valign=t}
\begin{tabular}{cccccc}
\includegraphics[width=0.149\textwidth]{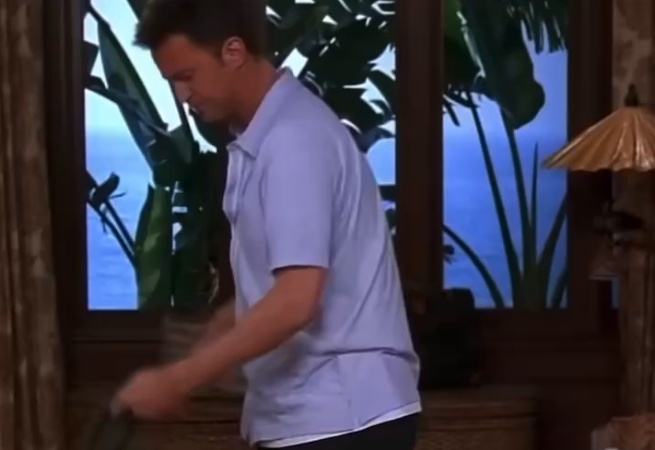} \hspace{-4mm} &
\includegraphics[width=0.149\textwidth]{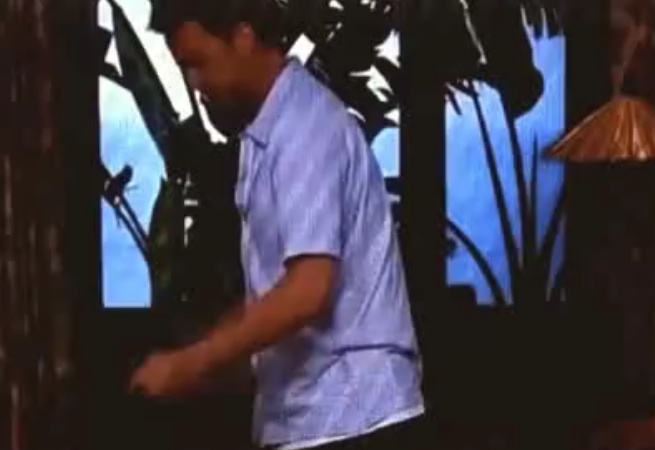} \hspace{-4mm} &
\includegraphics[width=0.149\textwidth]{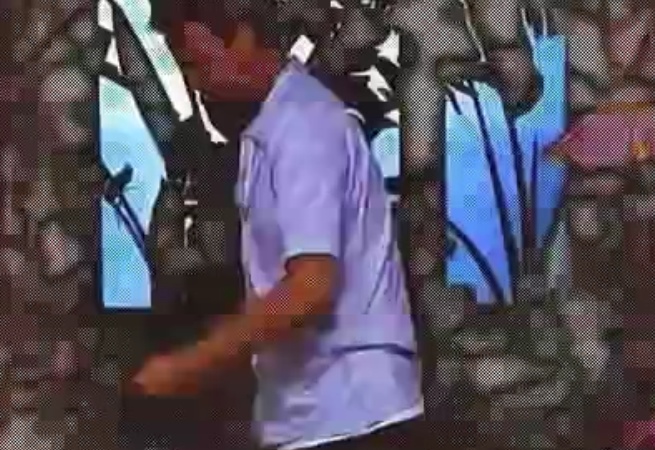} \hspace{-4mm} &
\includegraphics[width=0.149\textwidth]{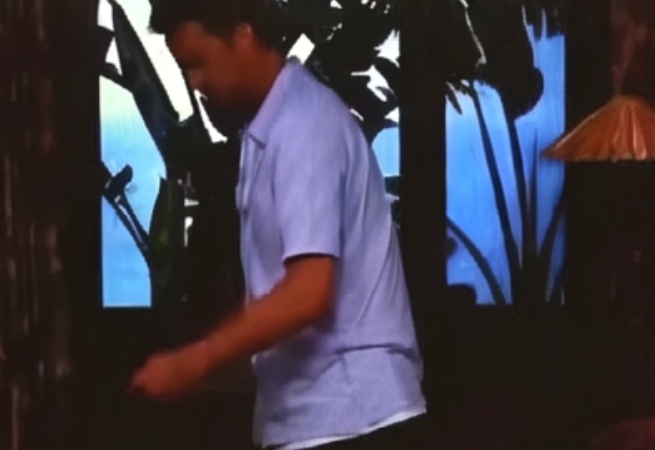} \hspace{-4mm} &
\includegraphics[width=0.149\textwidth]{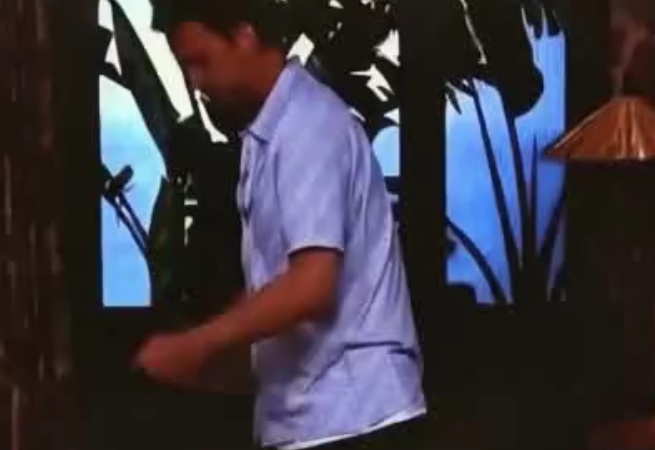} \hspace{-4mm} 
\\
GT \hspace{-4mm} &
LQ \hspace{-4mm} &
STBN~\cite{STBN} \hspace{-4mm} &
STAR~\cite{star} \hspace{-4mm} &
DLoRAL~\cite{DLoRAL} \hspace{-4mm}
\\
\includegraphics[width=0.149\textwidth]{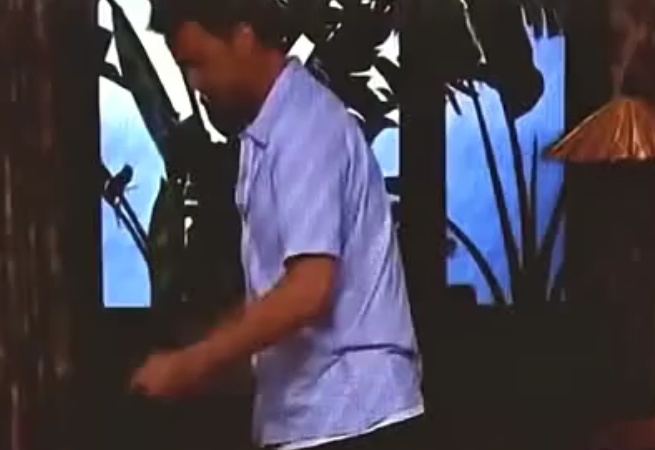} \hspace{-4mm} &
\includegraphics[width=0.149\textwidth]{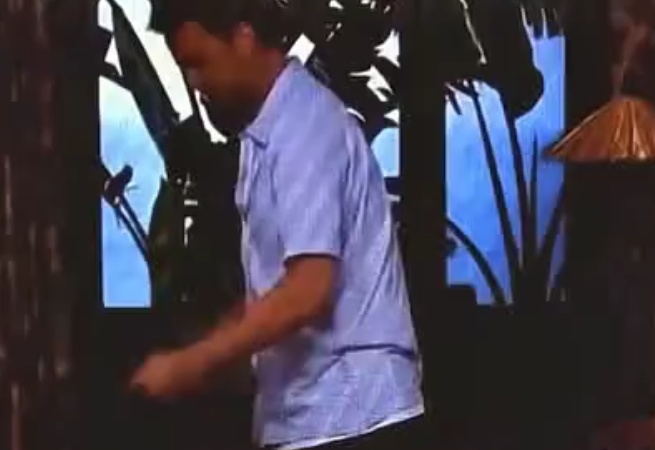} \hspace{-4mm} &
\includegraphics[width=0.149\textwidth]{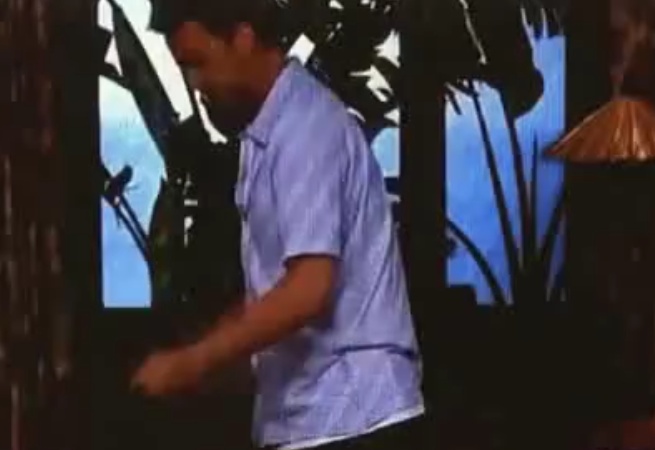} \hspace{-4mm} &
\includegraphics[width=0.149\textwidth]{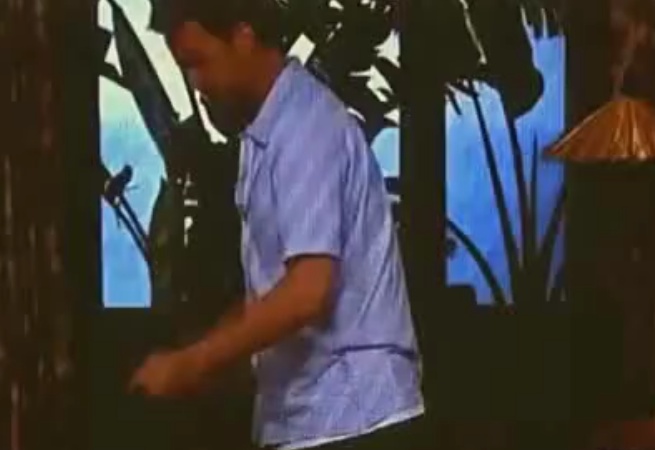} \hspace{-4mm} &
\includegraphics[width=0.149\textwidth]{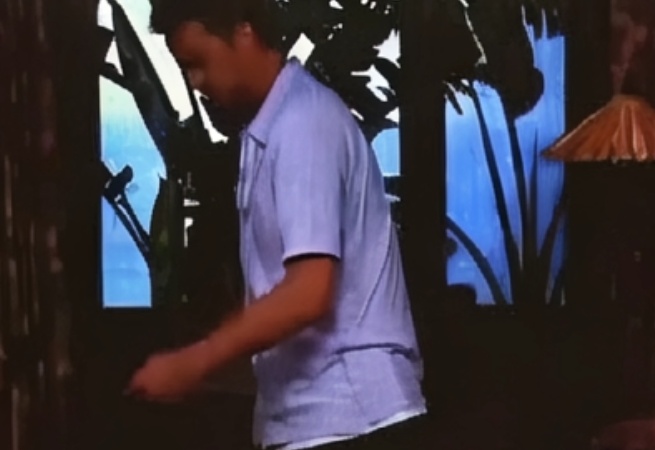} \hspace{-4mm}  
\\ 
FPANet~\cite{oh2025fpanet} \hspace{-4mm} &
CompEvent~\cite{zhong2025compevent} \hspace{-4mm} &
Flickerformer*~\cite{flickerformer} \hspace{-4mm} &
Flickerformer~\cite{flickerformer}  \hspace{-4mm} &
VDFP (ours) \hspace{-4mm}
\\
\end{tabular}
\end{adjustbox}
\\
\hspace{-0.42cm}
\begin{adjustbox}{valign=t}
\begin{tabular}{c}
\includegraphics[width=0.216\textwidth,height=0.231\textwidth]{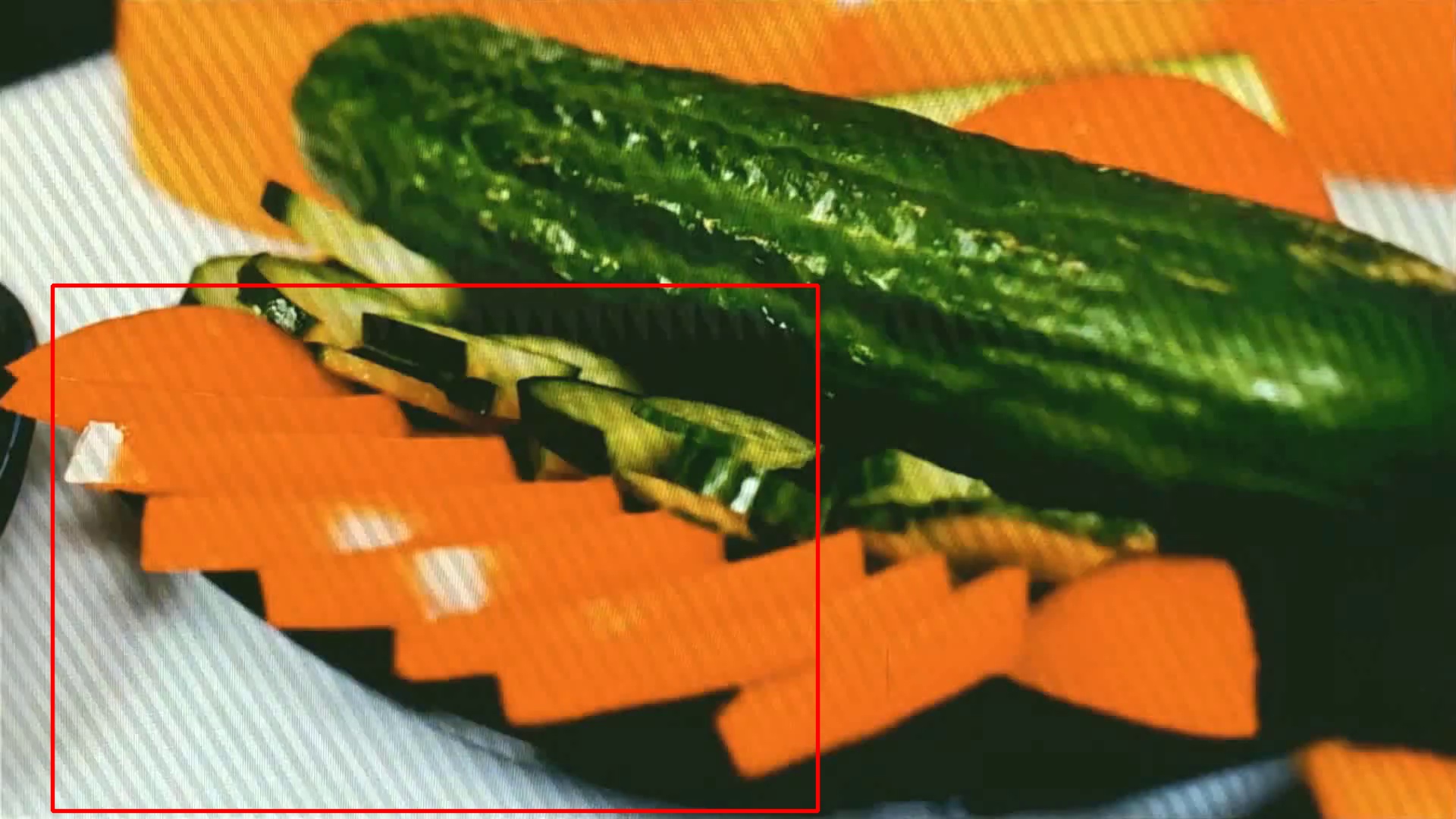}
\\
DeViD: 056
\end{tabular}
\end{adjustbox}
\hspace{-0.46cm}
\begin{adjustbox}{valign=t}
\begin{tabular}{cccccc}
\includegraphics[width=0.149\textwidth]{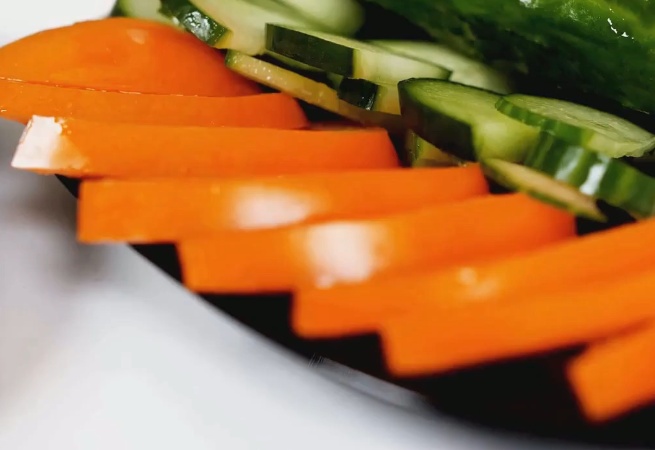} \hspace{-4mm} &
\includegraphics[width=0.149\textwidth]{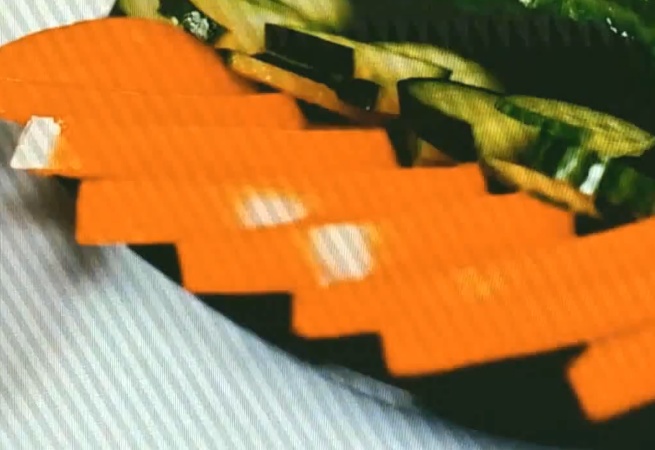} \hspace{-4mm} &
\includegraphics[width=0.149\textwidth]{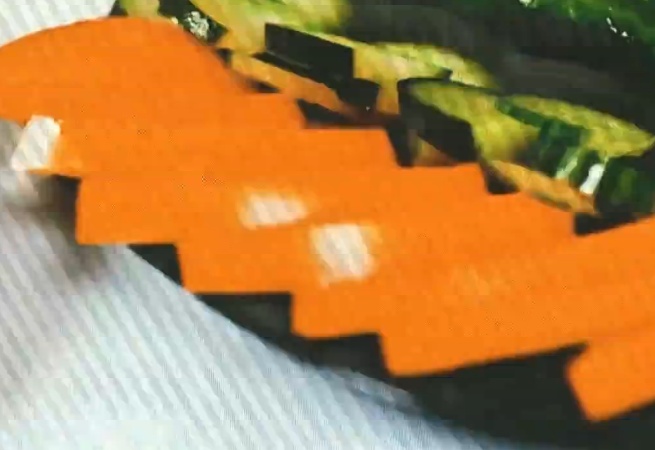} \hspace{-4mm} &
\includegraphics[width=0.149\textwidth]{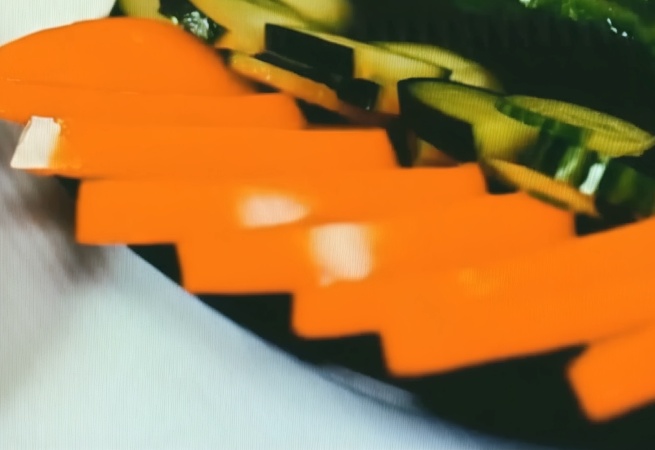} \hspace{-4mm} &
\includegraphics[width=0.149\textwidth]{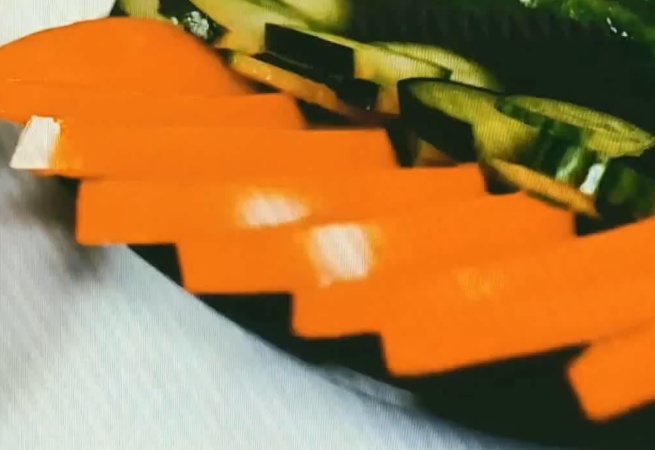} \hspace{-4mm} 
\\
GT \hspace{-4mm} &
LQ \hspace{-4mm} &
STBN~\cite{STBN} \hspace{-4mm} &
STAR~\cite{star} \hspace{-4mm} &
DLoRAL~\cite{DLoRAL} \hspace{-4mm}
\\
\includegraphics[width=0.149\textwidth]{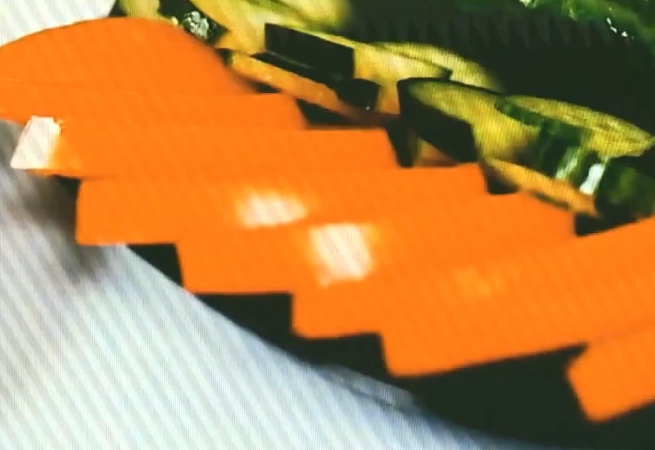} \hspace{-4mm} &
\includegraphics[width=0.149\textwidth]{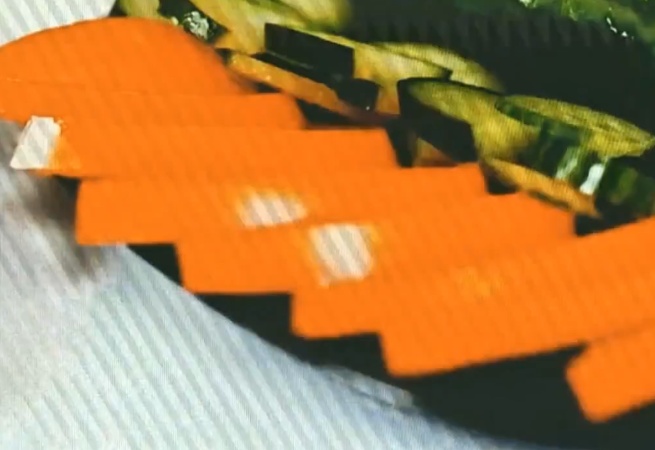} \hspace{-4mm} &
\includegraphics[width=0.149\textwidth]{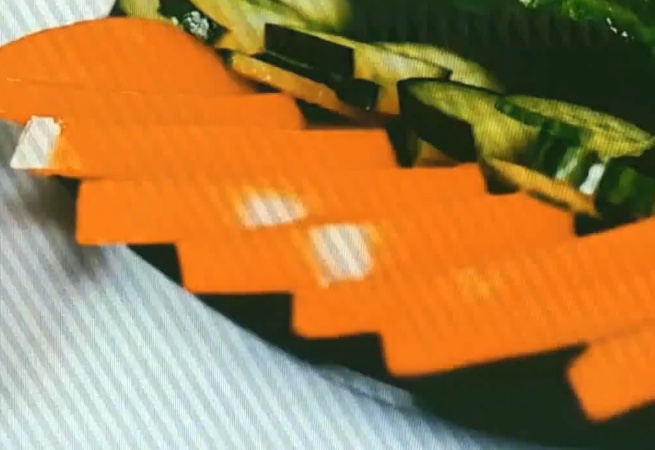} \hspace{-4mm} &
\includegraphics[width=0.149\textwidth]{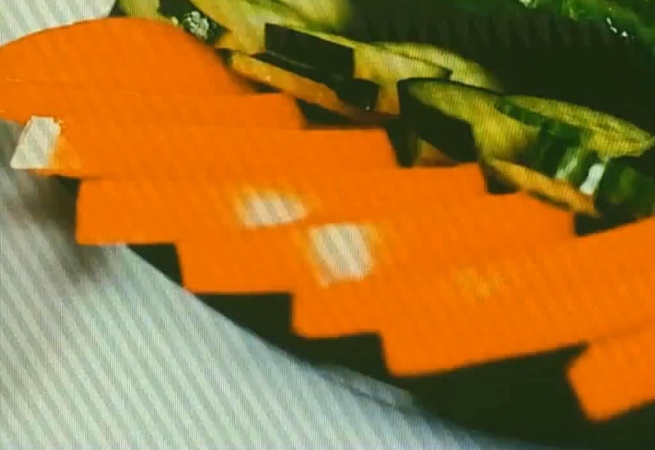} \hspace{-4mm} &
\includegraphics[width=0.149\textwidth]{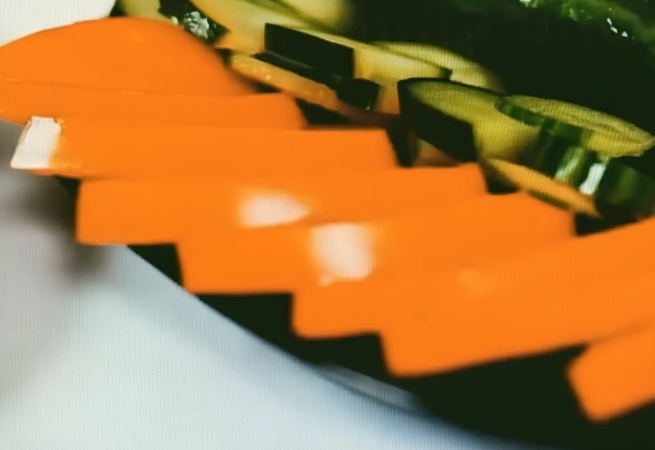} \hspace{-4mm}  
\\ 
FPANet~\cite{oh2025fpanet} \hspace{-4mm} &
CompEvent~\cite{zhong2025compevent} \hspace{-4mm} &
Flickerformer*~\cite{flickerformer} \hspace{-4mm} &
Flickerformer~\cite{flickerformer}  \hspace{-4mm} &
VDFP (ours) \hspace{-4mm}
\\
\end{tabular}
\end{adjustbox}
\\
\hspace{-0.42cm}
\begin{adjustbox}{valign=t}
\begin{tabular}{c}
\includegraphics[width=0.216\textwidth,height=0.231\textwidth]{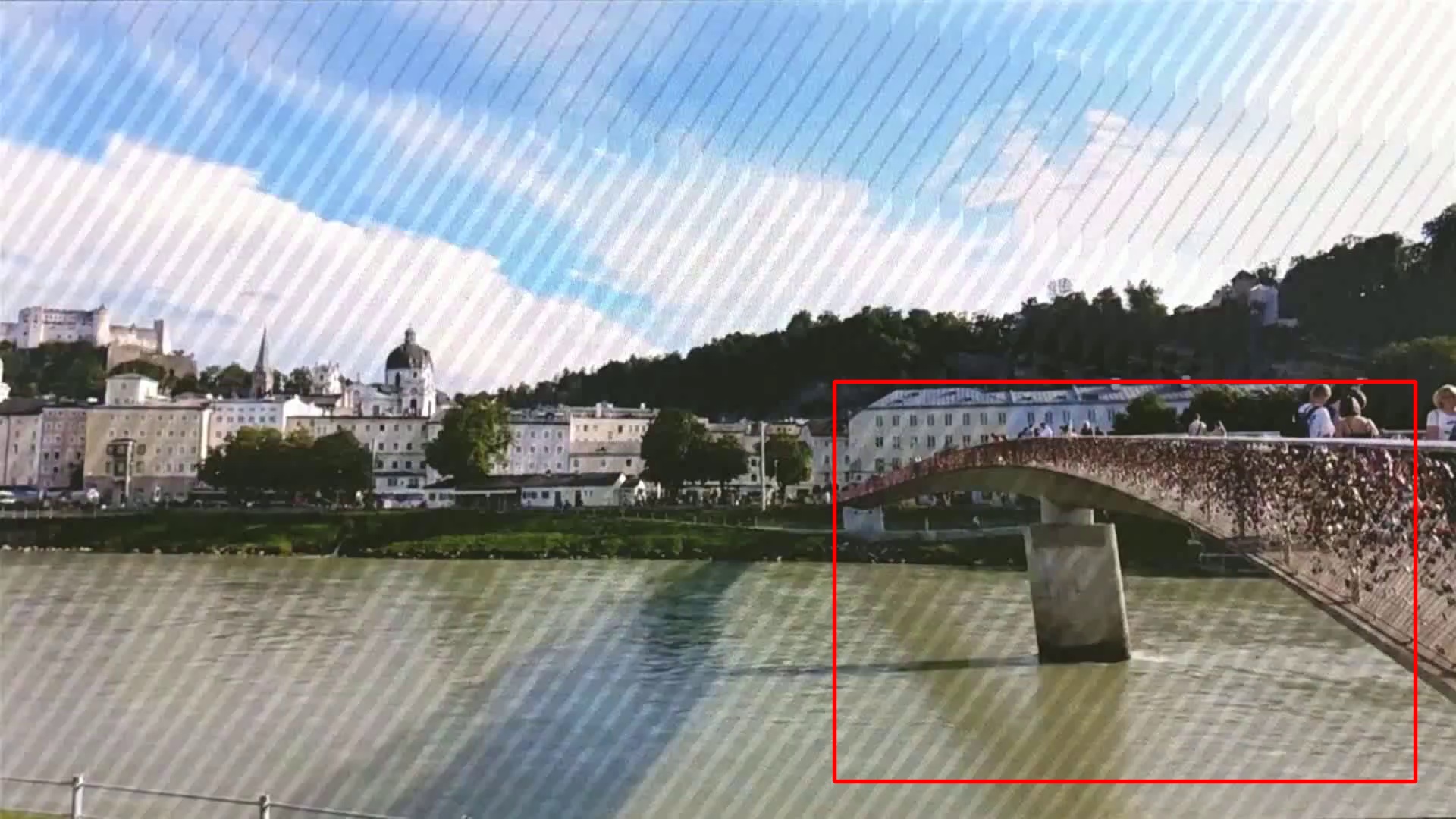}
\\
DeViD: 032
\end{tabular}
\end{adjustbox}
\hspace{-0.46cm}
\begin{adjustbox}{valign=t}
\begin{tabular}{cccccc}
\includegraphics[width=0.149\textwidth]{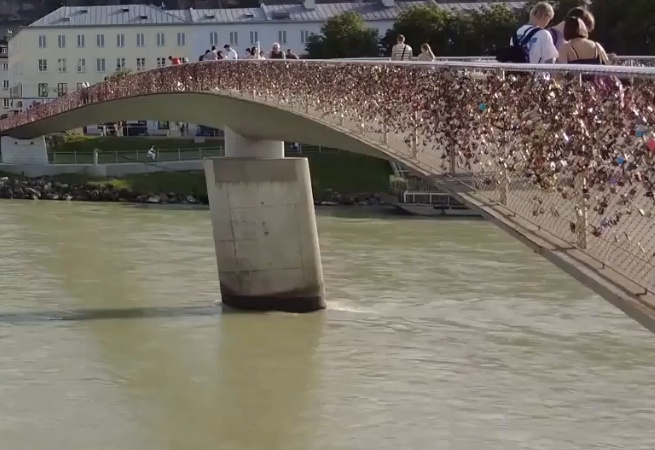} \hspace{-4mm} &
\includegraphics[width=0.149\textwidth]{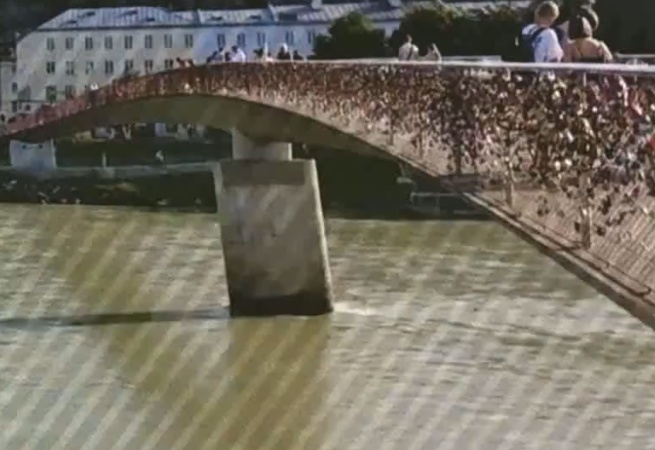} \hspace{-4mm} &
\includegraphics[width=0.149\textwidth]{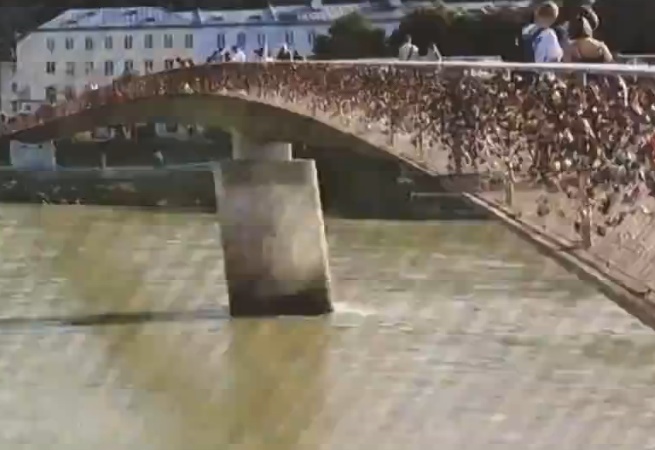} \hspace{-4mm} &
\includegraphics[width=0.149\textwidth]{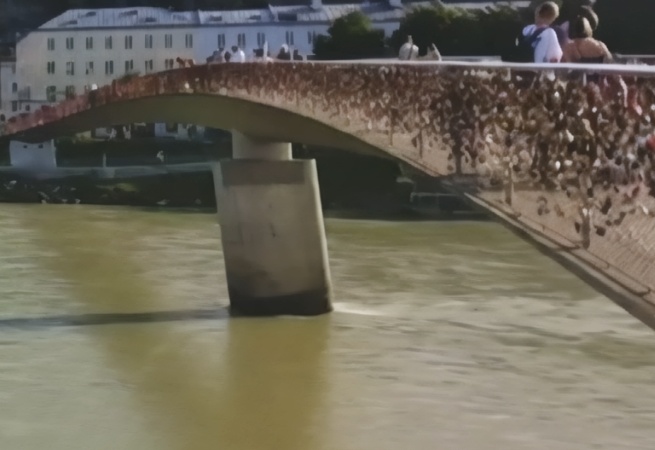} \hspace{-4mm} &
\includegraphics[width=0.149\textwidth]{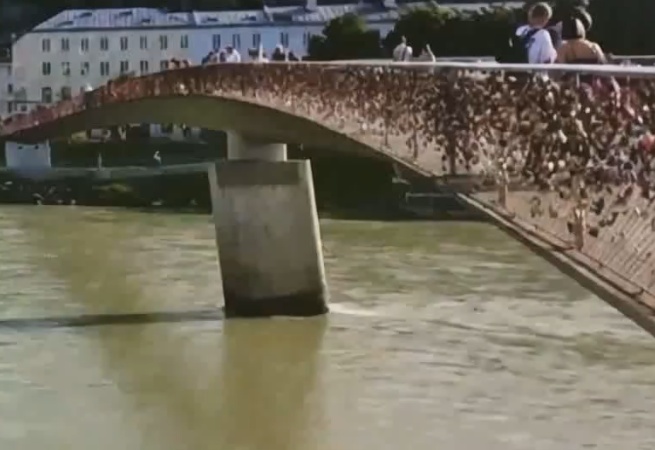} \hspace{-4mm} 
\\
GT \hspace{-4mm} &
LQ \hspace{-4mm} &
STBN~\cite{STBN} \hspace{-4mm} &
STAR~\cite{star} \hspace{-4mm} &
DLoRAL~\cite{DLoRAL} \hspace{-4mm}
\\
\includegraphics[width=0.149\textwidth]{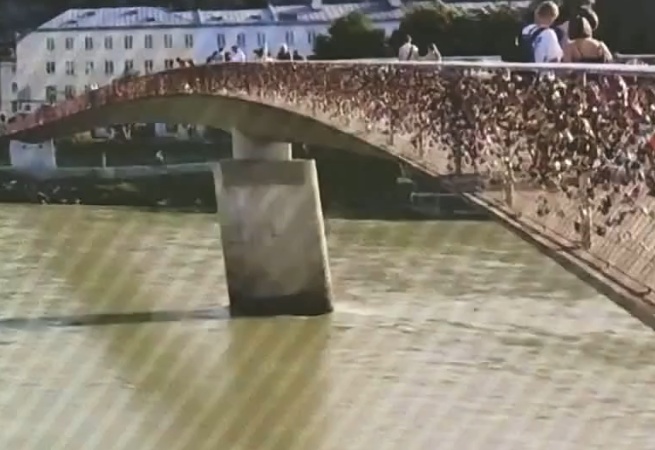} \hspace{-4mm} &
\includegraphics[width=0.149\textwidth]{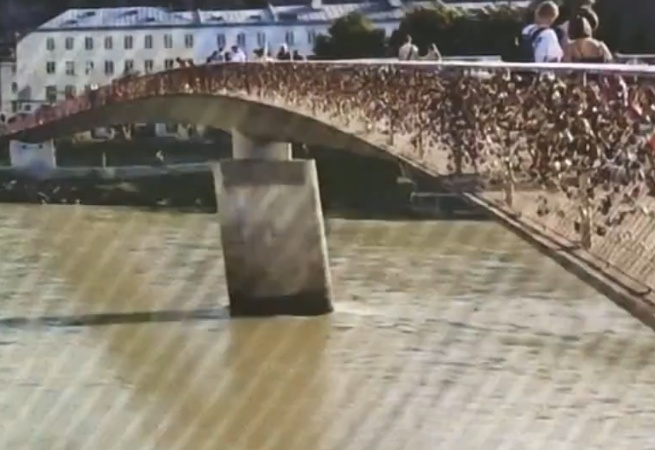} \hspace{-4mm} &
\includegraphics[width=0.149\textwidth]{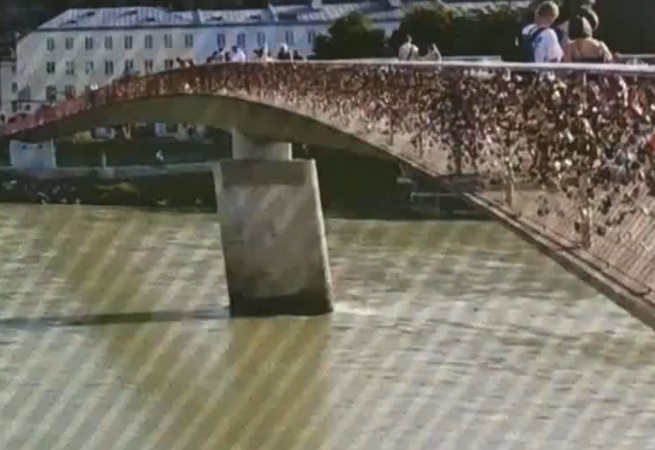} \hspace{-4mm} &
\includegraphics[width=0.149\textwidth]{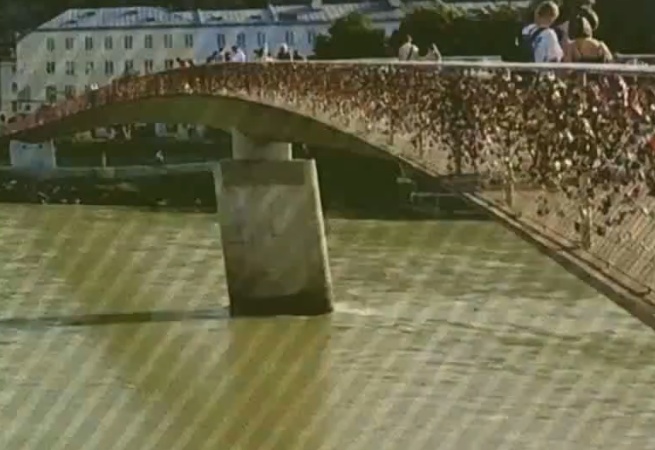} \hspace{-4mm} &
\includegraphics[width=0.149\textwidth]{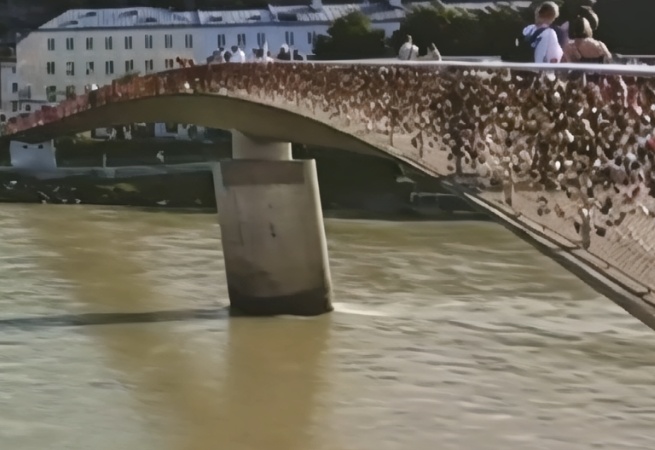} \hspace{-4mm}  
\\ 
FPANet~\cite{oh2025fpanet} \hspace{-4mm} &
CompEvent~\cite{zhong2025compevent} \hspace{-4mm} &
Flickerformer*~\cite{flickerformer} \hspace{-4mm} &
Flickerformer~\cite{flickerformer}  \hspace{-4mm} &
VDFP (ours) \hspace{-4mm}
\\
\end{tabular}
\end{adjustbox}
\end{tabular}
\caption{More visual comparison between VDFP and other methods on our real-world dataset DeViD. Specifically, Flickerformer* represents using the official checkpoint to test.}
\label{fig:more_visual_comparison}
\vspace{-6mm}
\end{figure*}

\begin{figure*}[htbp]
    \centering
    \footnotesize
    \begin{subfigure}[b]{0.19\linewidth}
        \includegraphics[width=\linewidth]{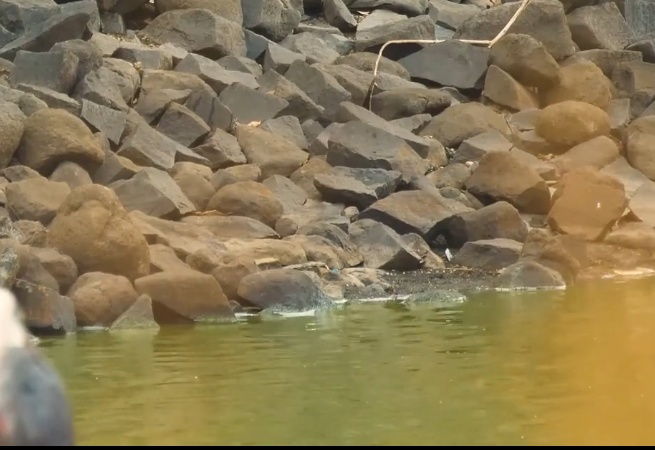}
    \end{subfigure}\hfill
    \begin{subfigure}[b]{0.19\linewidth}
        \includegraphics[width=\linewidth]{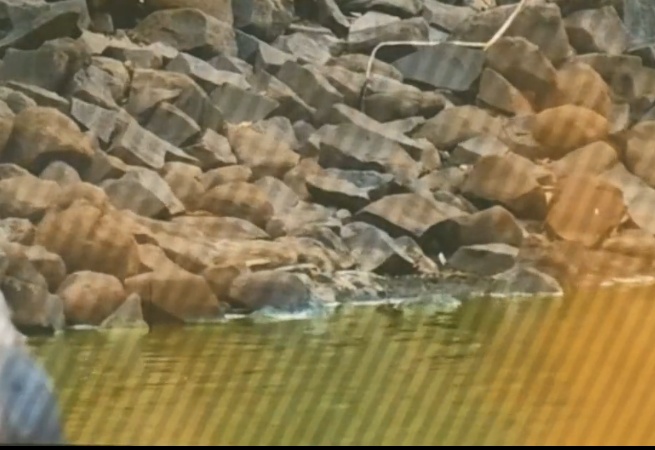}
    \end{subfigure}\hfill
    \begin{subfigure}[b]{0.19\linewidth}
        \includegraphics[width=\linewidth]{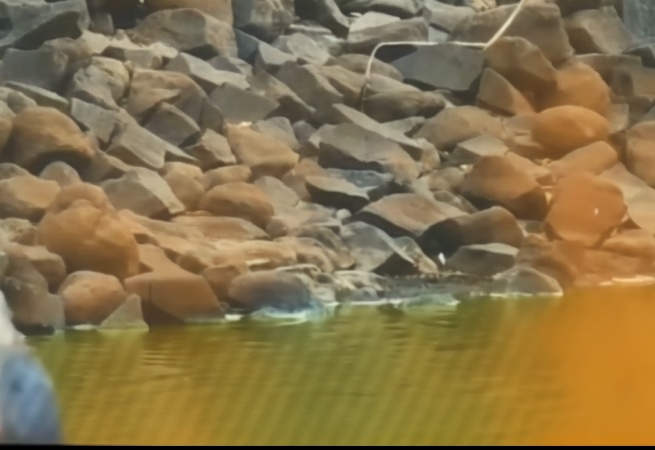}
    \end{subfigure}\hfill
    \begin{subfigure}[b]{0.19\linewidth}
        \includegraphics[width=\linewidth]{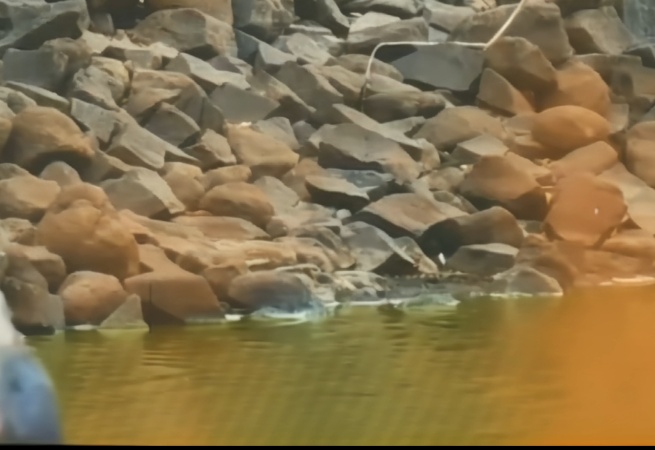}
    \end{subfigure}\hfill
    \begin{subfigure}[b]{0.19\linewidth}
        \includegraphics[width=\linewidth]{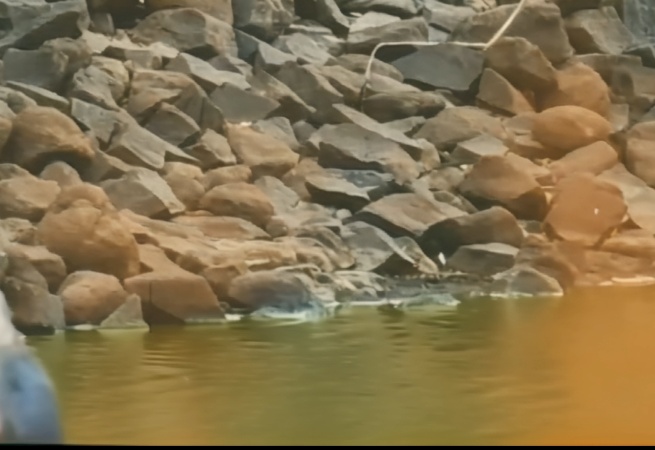}
    \end{subfigure}
    
    \vspace{2pt}
    \begin{subfigure}[b]{0.19\linewidth}
        \includegraphics[width=\linewidth]{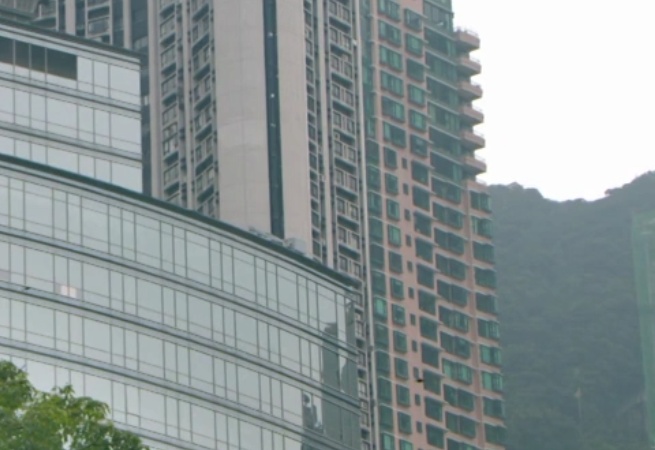}
    \end{subfigure}\hfill
    \begin{subfigure}[b]{0.19\linewidth}
        \includegraphics[width=\linewidth]{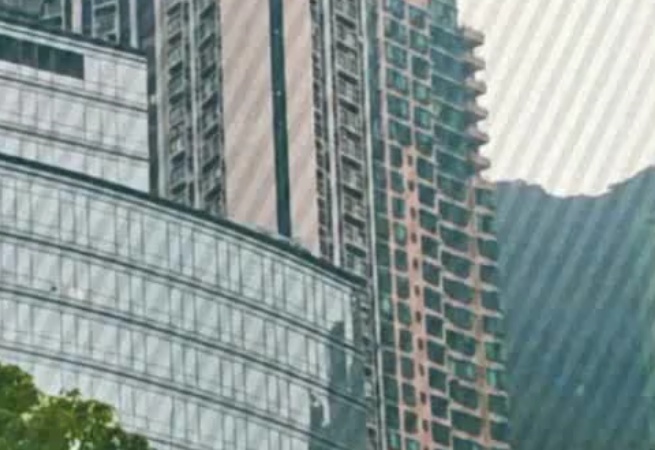}
    \end{subfigure}\hfill
    \begin{subfigure}[b]{0.19\linewidth}
        \includegraphics[width=\linewidth]{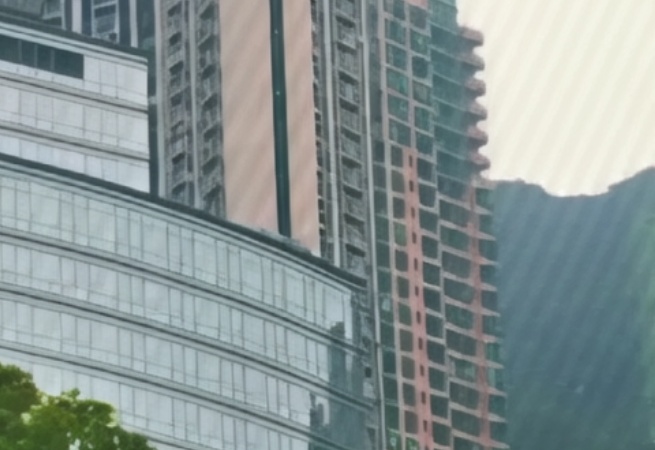}
    \end{subfigure}\hfill
    \begin{subfigure}[b]{0.19\linewidth}
        \includegraphics[width=\linewidth]{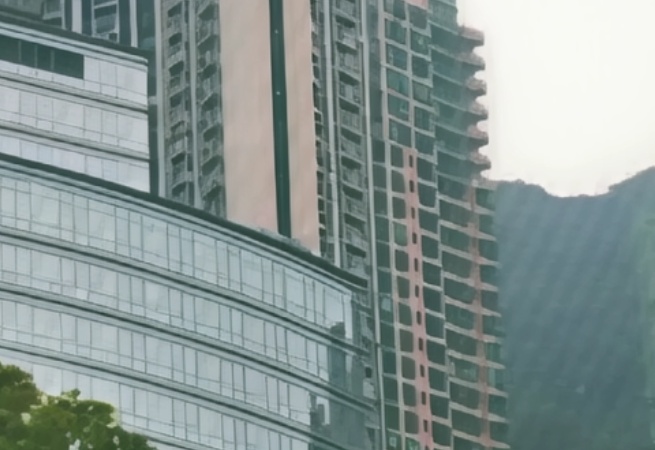}
    \end{subfigure}\hfill
    \begin{subfigure}[b]{0.19\linewidth}
        \includegraphics[width=\linewidth]{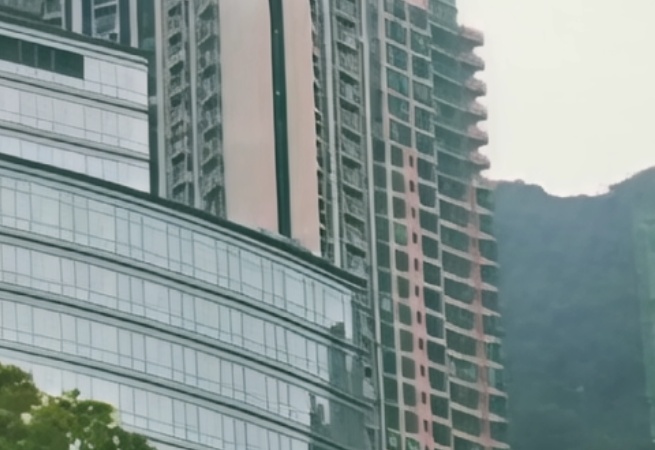}
    \end{subfigure}
    
    \vspace{2pt}

    \begin{subfigure}[b]{0.19\linewidth}
        \includegraphics[width=\linewidth]{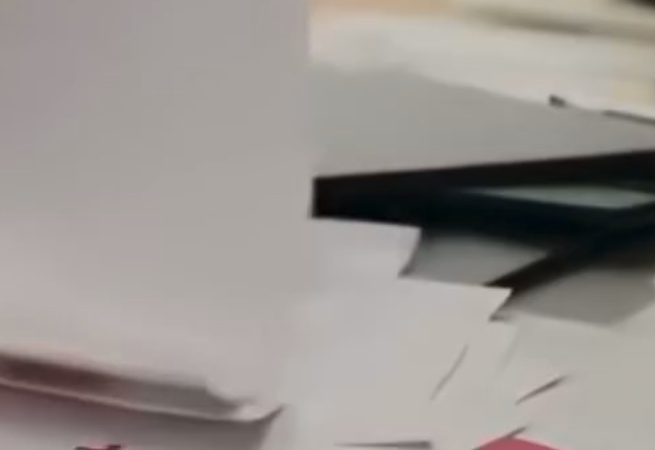}
    \end{subfigure}\hfill
    \begin{subfigure}[b]{0.19\linewidth}
        \includegraphics[width=\linewidth]{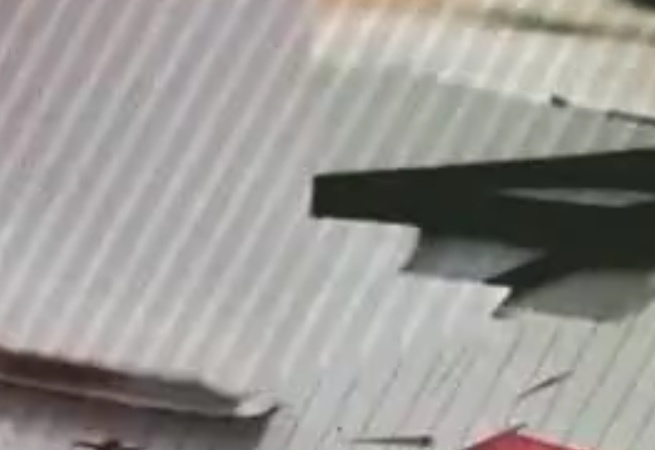}
    \end{subfigure}\hfill
    \begin{subfigure}[b]{0.19\linewidth}
        \includegraphics[width=\linewidth]{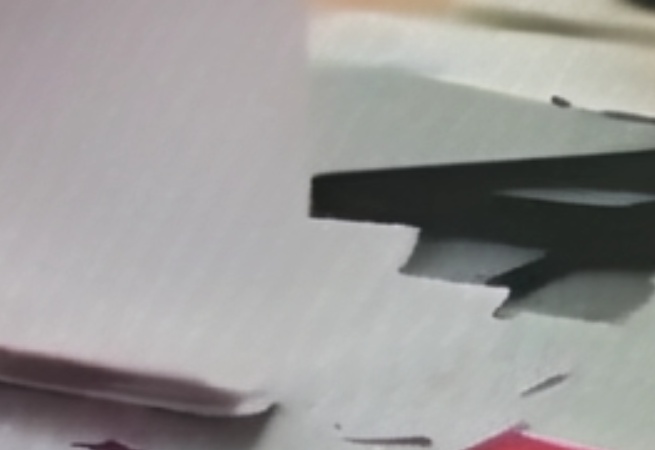}
    \end{subfigure}\hfill
    \begin{subfigure}[b]{0.19\linewidth}
        \includegraphics[width=\linewidth]{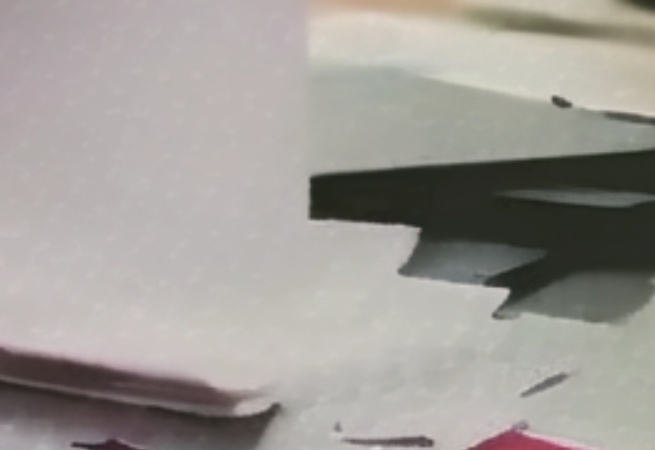}
    \end{subfigure}\hfill
    \begin{subfigure}[b]{0.19\linewidth}
        \includegraphics[width=\linewidth]{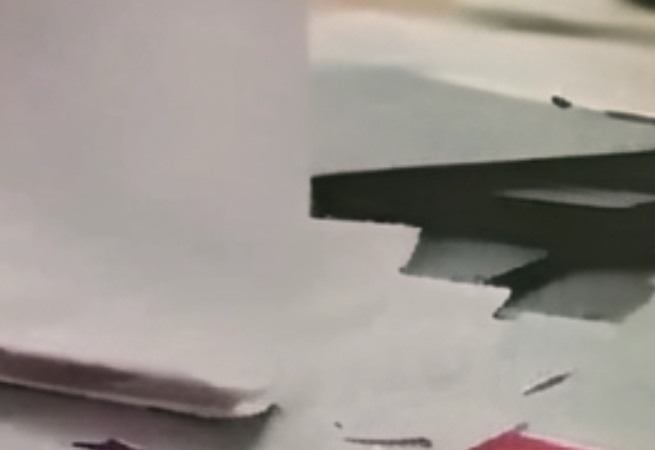}
    \end{subfigure}

    \vspace{2pt} 
    \begin{subfigure}[b]{0.19\linewidth}
        \includegraphics[width=\linewidth]{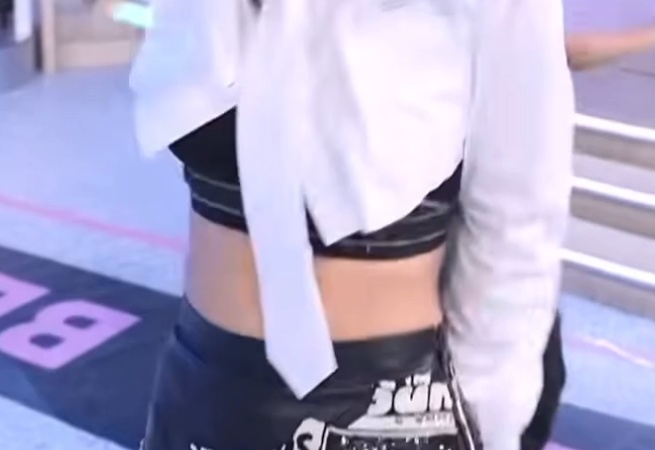}
    \end{subfigure}\hfill
    \begin{subfigure}[b]{0.19\linewidth}
        \includegraphics[width=\linewidth]{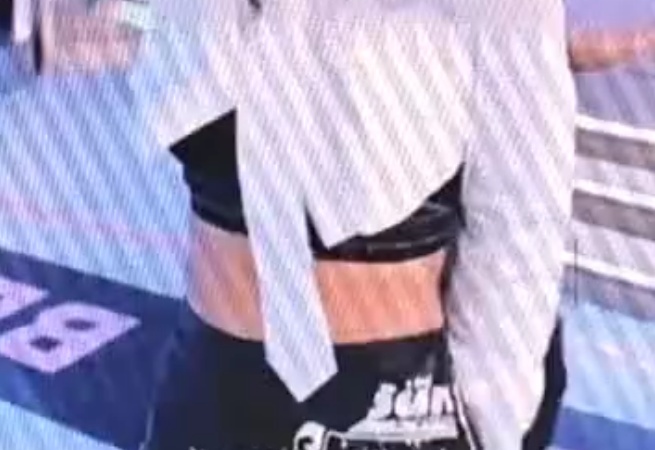}
    \end{subfigure}\hfill
    \begin{subfigure}[b]{0.19\linewidth}
        \includegraphics[width=\linewidth]{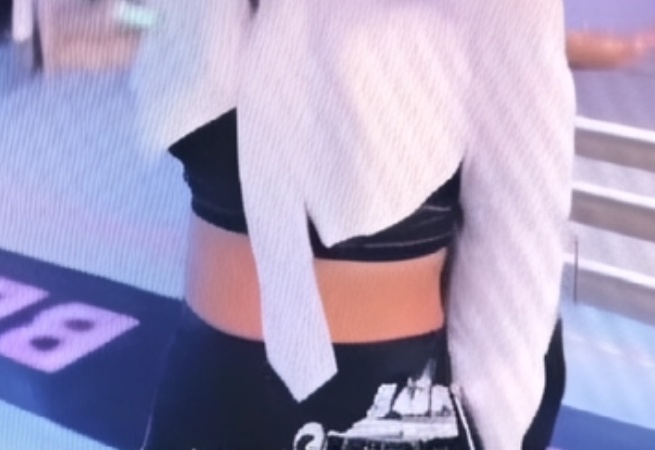}
    \end{subfigure}\hfill
    \begin{subfigure}[b]{0.19\linewidth}
        \includegraphics[width=\linewidth]{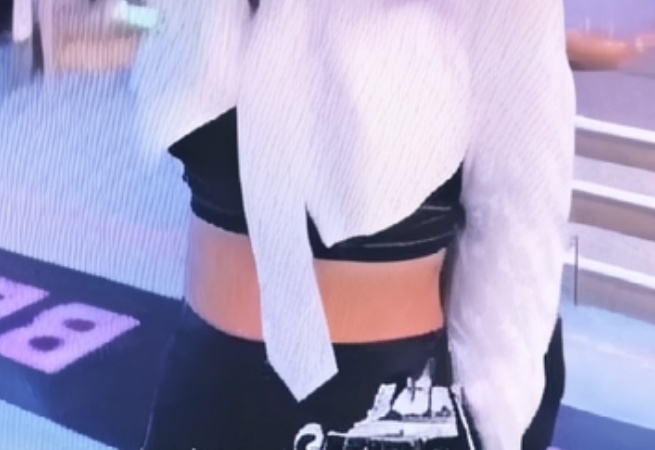}
    \end{subfigure}\hfill
    \begin{subfigure}[b]{0.19\linewidth}
        \includegraphics[width=\linewidth]{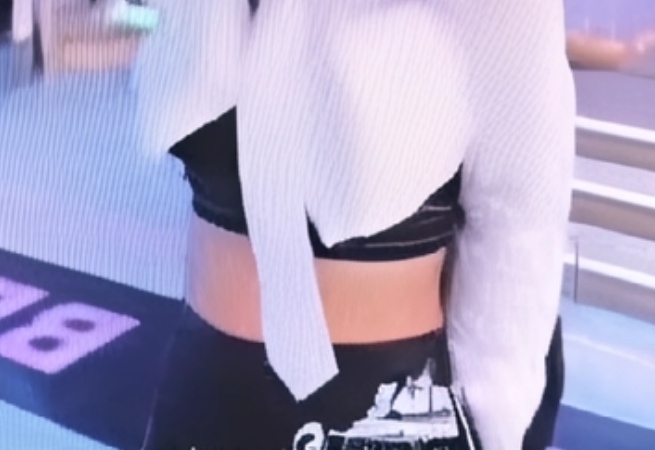}
    \end{subfigure}
    
    \vspace{2pt}
    \begin{subfigure}[b]{0.19\linewidth}
        \includegraphics[width=\linewidth]{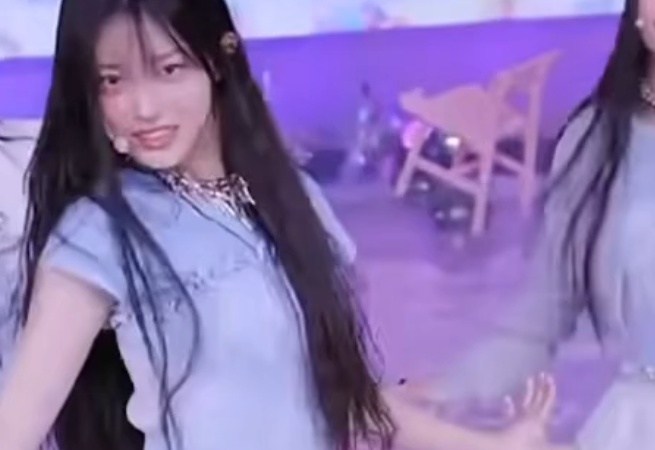}
    \end{subfigure}\hfill
    \begin{subfigure}[b]{0.19\linewidth}
        \includegraphics[width=\linewidth]{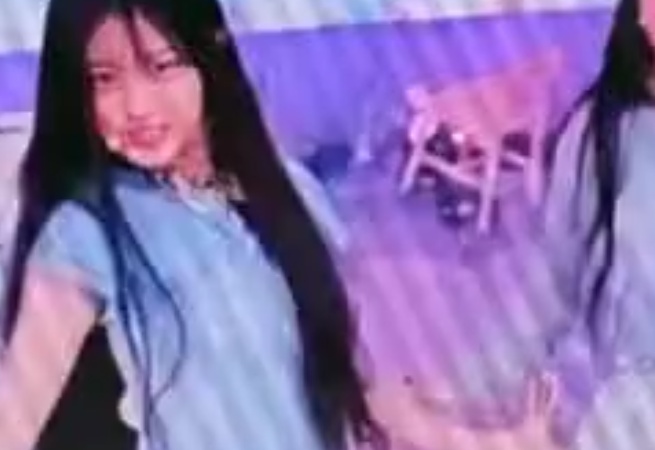}
    \end{subfigure}\hfill
    \begin{subfigure}[b]{0.19\linewidth}
        \includegraphics[width=\linewidth]{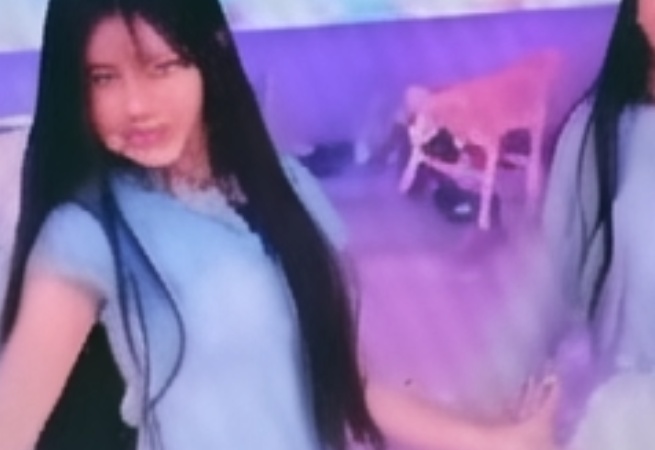}
    \end{subfigure}\hfill
    \begin{subfigure}[b]{0.19\linewidth}
        \includegraphics[width=\linewidth]{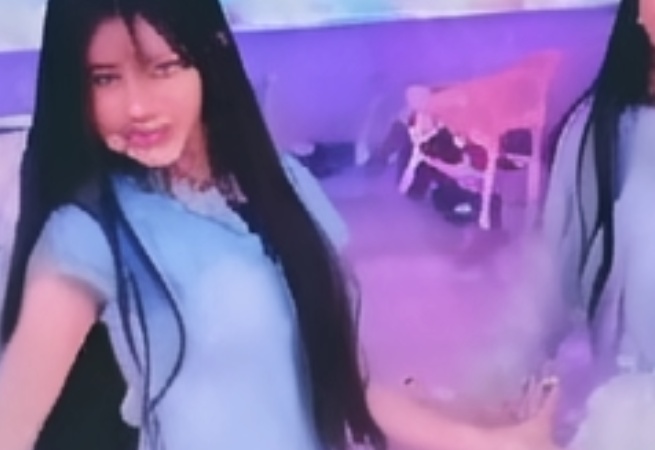}
    \end{subfigure}\hfill
    \begin{subfigure}[b]{0.19\linewidth}
        \includegraphics[width=\linewidth]{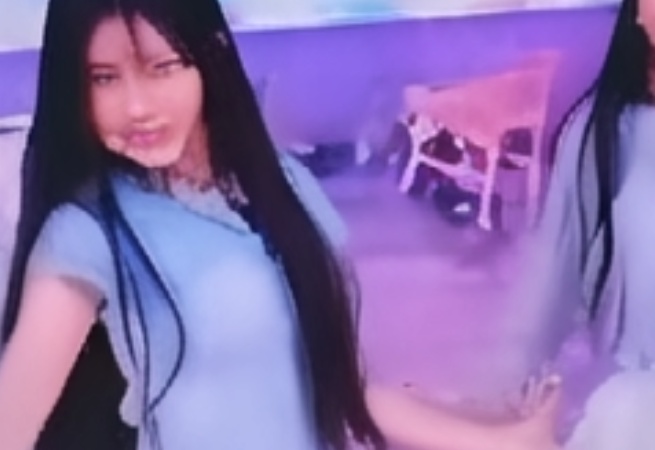}
    \end{subfigure}

      \vspace{2pt} 
    \begin{subfigure}[b]{0.19\linewidth}
        \includegraphics[width=\linewidth]{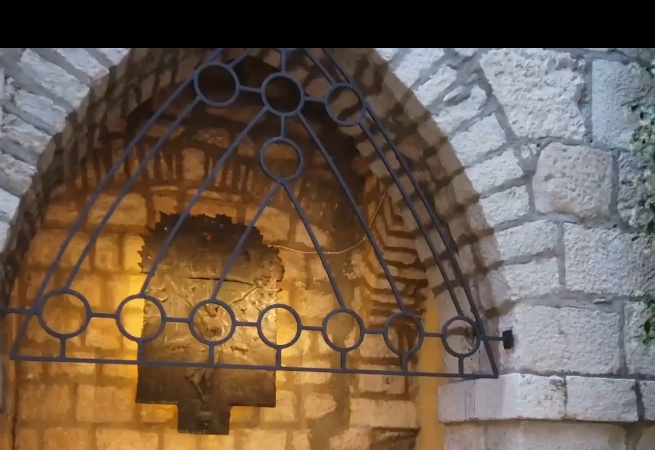}
    \end{subfigure}\hfill
    \begin{subfigure}[b]{0.19\linewidth}
        \includegraphics[width=\linewidth]{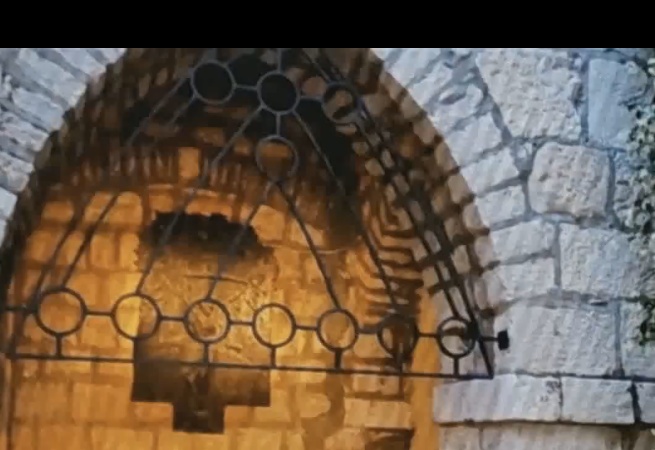}
    \end{subfigure}\hfill
    \begin{subfigure}[b]{0.19\linewidth}
        \includegraphics[width=\linewidth]{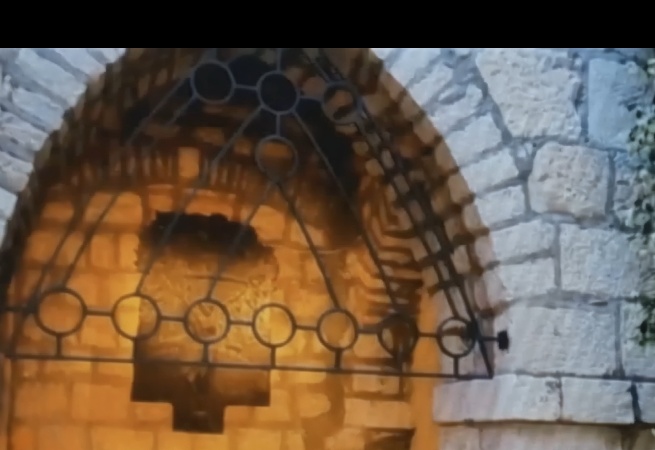}
    \end{subfigure}\hfill
    \begin{subfigure}[b]{0.19\linewidth}
        \includegraphics[width=\linewidth]{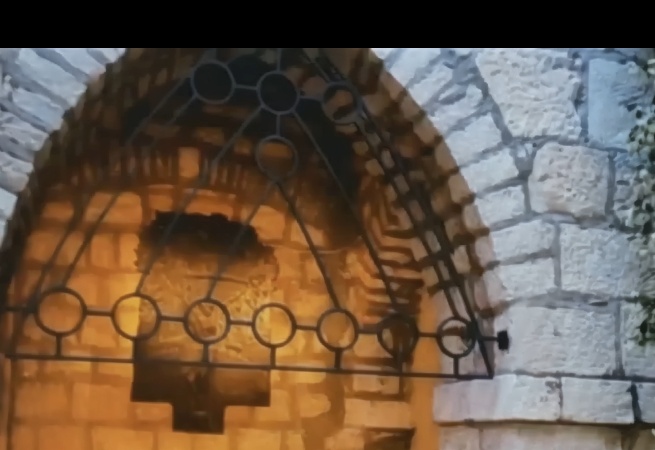}
    \end{subfigure}\hfill
    \begin{subfigure}[b]{0.19\linewidth}
        \includegraphics[width=\linewidth]{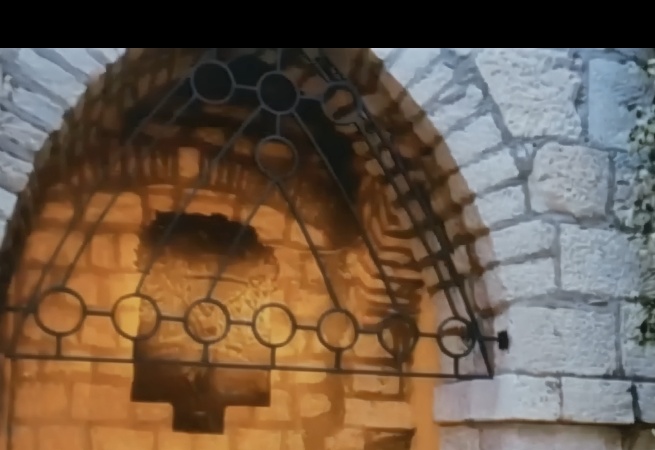}
    \end{subfigure}

      \vspace{2pt} 
    \begin{subfigure}[b]{0.19\linewidth}
        \includegraphics[width=\linewidth]{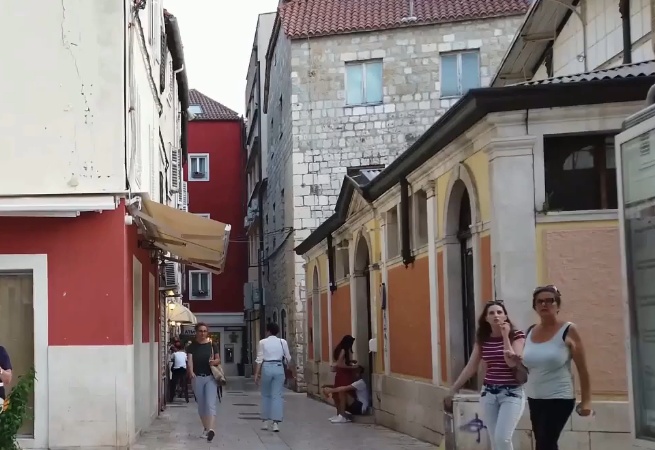}
    \end{subfigure}\hfill
    \begin{subfigure}[b]{0.19\linewidth}
        \includegraphics[width=\linewidth]{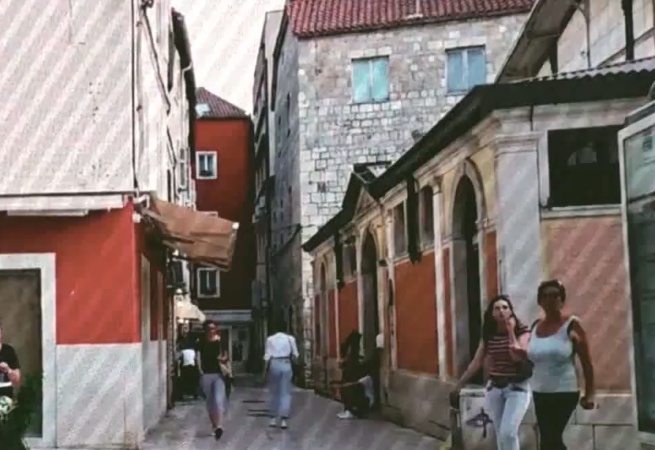}
    \end{subfigure}\hfill
    \begin{subfigure}[b]{0.19\linewidth}
        \includegraphics[width=\linewidth]{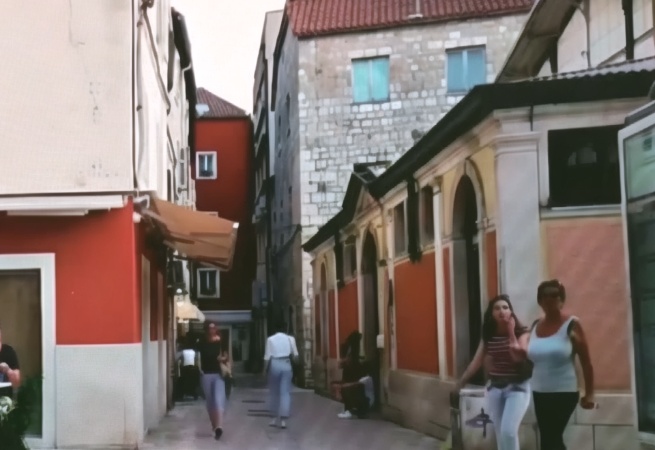}
    \end{subfigure}\hfill
    \begin{subfigure}[b]{0.19\linewidth}
        \includegraphics[width=\linewidth]{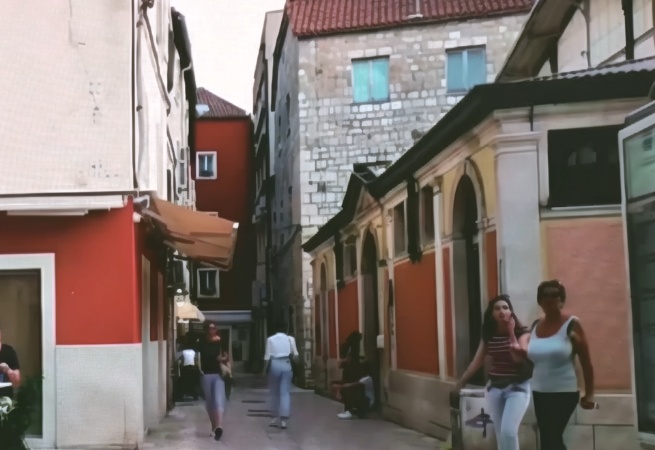}
    \end{subfigure}\hfill
    \begin{subfigure}[b]{0.19\linewidth}
        \includegraphics[width=\linewidth]{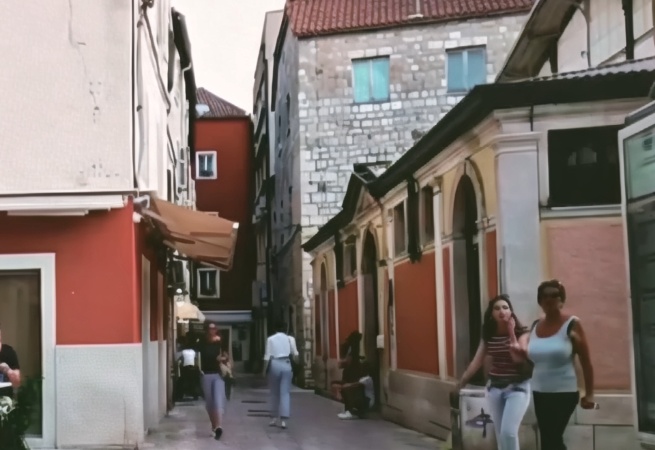}
    \end{subfigure}

      \vspace{2pt} 
    \begin{subfigure}[b]{0.19\linewidth}
        \includegraphics[width=\linewidth]{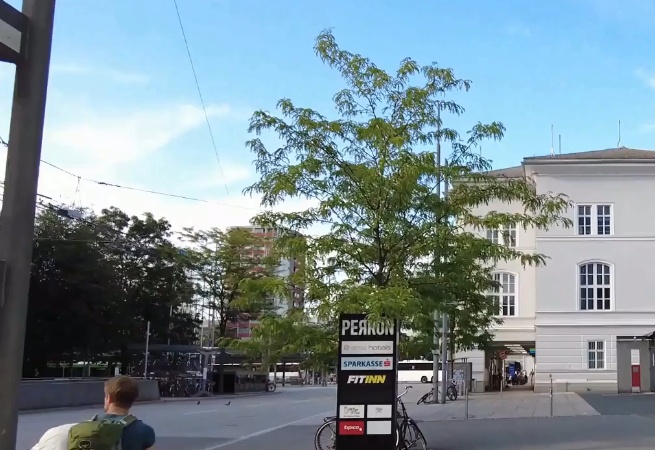}
    \end{subfigure}\hfill
    \begin{subfigure}[b]{0.19\linewidth}
        \includegraphics[width=\linewidth]{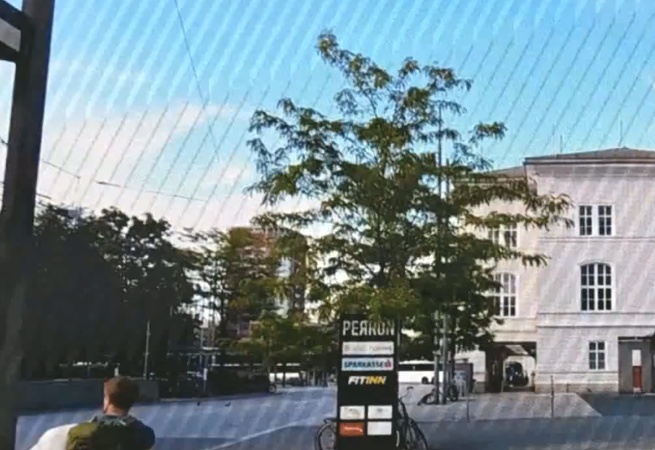}
    \end{subfigure}\hfill
    \begin{subfigure}[b]{0.19\linewidth}
        \includegraphics[width=\linewidth]{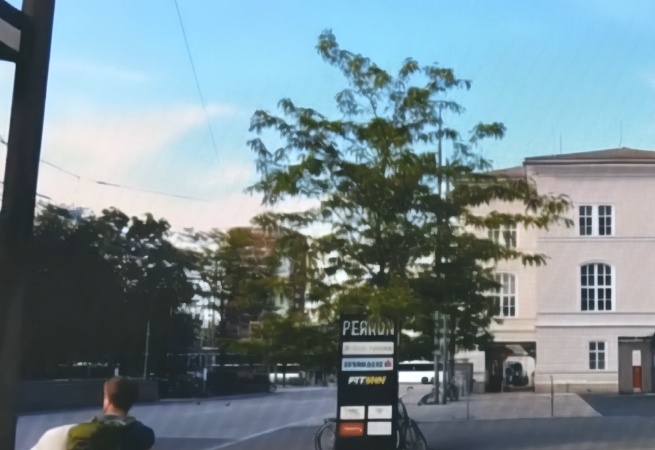}
    \end{subfigure}\hfill
    \begin{subfigure}[b]{0.19\linewidth}
        \includegraphics[width=\linewidth]{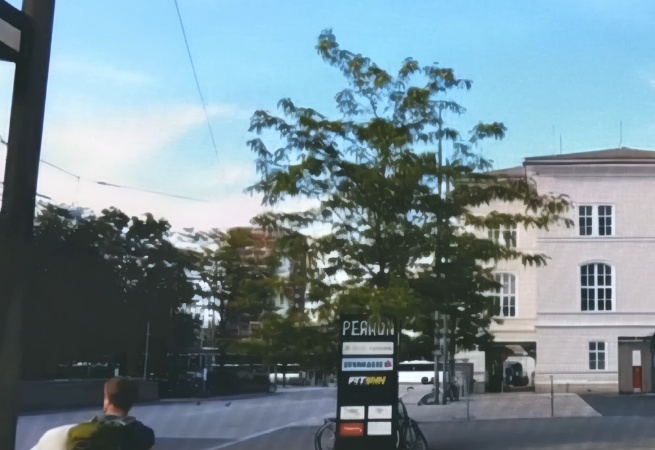}
    \end{subfigure}\hfill
    \begin{subfigure}[b]{0.19\linewidth}
        \includegraphics[width=\linewidth]{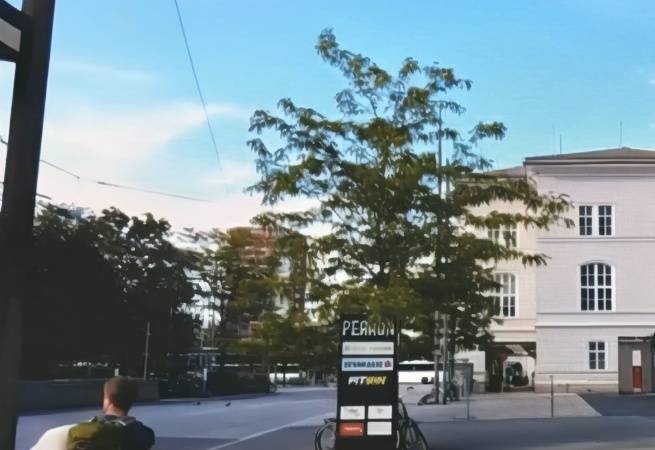}
    \end{subfigure}

      \vspace{2pt} 
    \begin{subfigure}[b]{0.19\linewidth}
        \includegraphics[width=\linewidth]{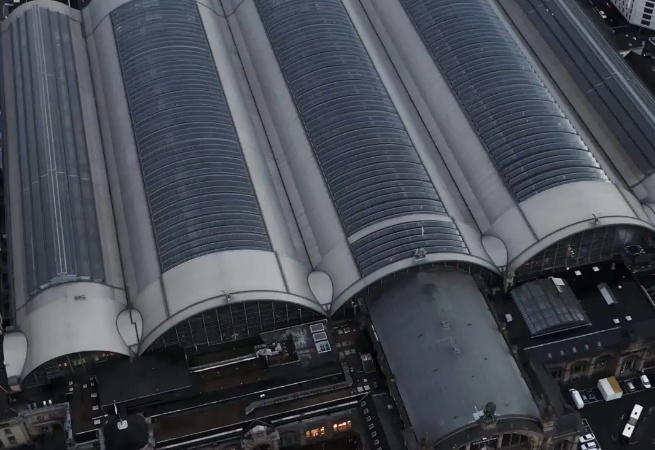}
    \end{subfigure}\hfill
    \begin{subfigure}[b]{0.19\linewidth}
        \includegraphics[width=\linewidth]{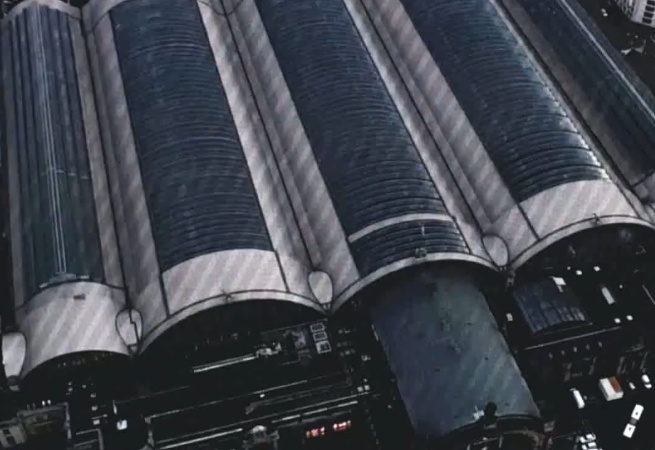}
    \end{subfigure}\hfill
    \begin{subfigure}[b]{0.19\linewidth}
        \includegraphics[width=\linewidth]{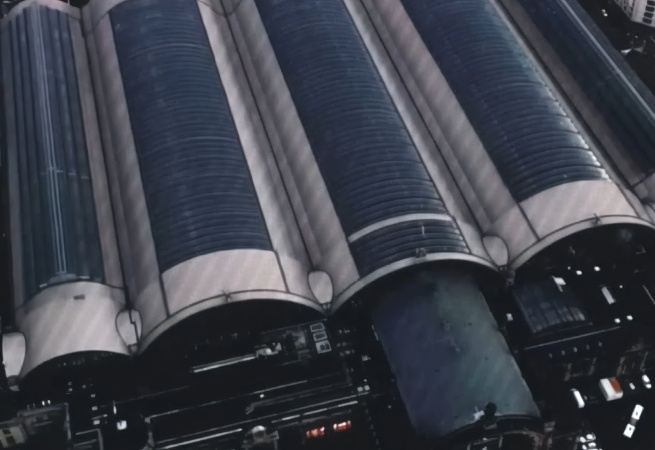}
    \end{subfigure}\hfill
    \begin{subfigure}[b]{0.19\linewidth}
        \includegraphics[width=\linewidth]{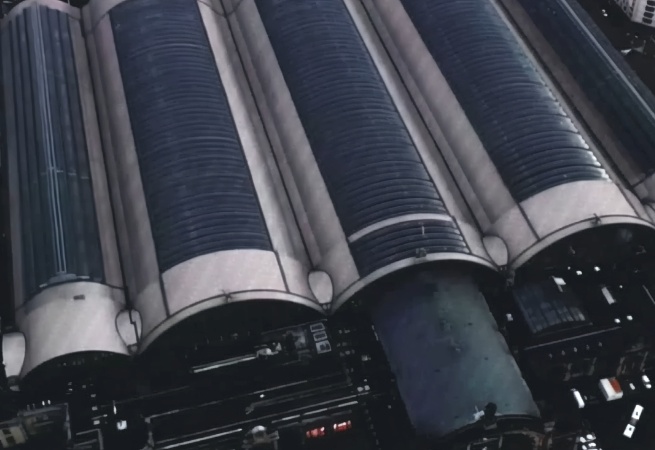}
    \end{subfigure}\hfill
    \begin{subfigure}[b]{0.19\linewidth}
        \includegraphics[width=\linewidth]{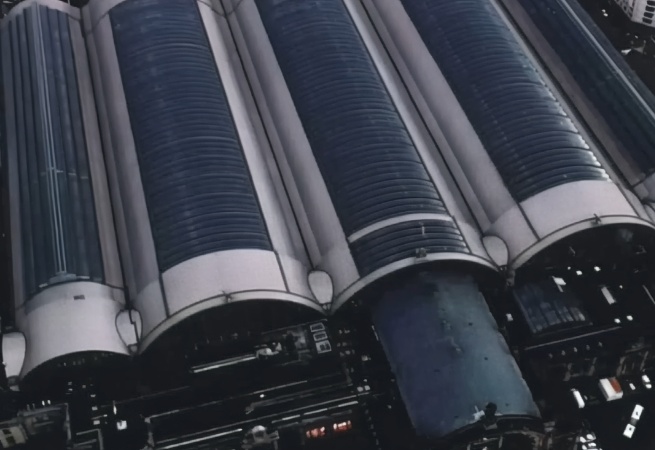}
    \end{subfigure}
    
    \makebox[0.19\linewidth][c]{GT}\hfill
    \makebox[0.19\linewidth][c]{LQ}\hfill
    \makebox[0.19\linewidth][c]{Baseline}\hfill
    \makebox[0.19\linewidth][c]{DFM}\hfill
    \makebox[0.19\linewidth][c]{DFM+CPP}
    \vspace{-6pt}
    \caption{More visual comparisons of ablation study. The figures show three sets of examples, and from left to right are: GT, LQ, Baseline, DFM, and DFM+CPP.}
    \label{fig:more_ablation_visual}
    \vspace{-5mm}
\end{figure*}

\subsection{More Visual Comparisons}
\label{section:visual_comparison}
This section will show more visual comparisons between our model (VDFP) and other methods, which further show the advantages of our model, as shown in Fig.~\ref{fig:more_visual_comparison}.
\subsection{Ablation Study}
\label{abl}
This section will show more examples of the ablation study of our model (VDFP), further indicating the efficiency of every step of our model, as shown in Fig.~\ref{fig:more_ablation_visual}.


\end{document}